\documentclass[12pt]{article}
\usepackage{amsmath}
\usepackage{amssymb}
\usepackage{enumerate}
\usepackage{graphicx}
\usepackage{pifont}
\usepackage{epsfig}
\usepackage{theorem}
\usepackage{xcolor}
\usepackage{algorithmic, algorithm}
\usepackage{cite}
\usepackage{rotating}
\usepackage{epstopdf}

\newcommand{\HH}{\ensuremath{{\mathcal H}}}

\newcommand{\RR}{\ensuremath{\mathbb R}}

\newcommand{\RX}{\ensuremath{\,\left]-\infty,+\infty\right]}}

\newcommand{\NN}{\ensuremath{\mathbb N}}

\newcommand{\prox}{\ensuremath{\mathrm{prox}}}

 \newcommand{\xx}{\underline{x}}
  \newcommand{\kx}{\underline{k}}
  \newcommand{\kk}{\underline{\xi}}

\theoremstyle{plain}{\theorembodyfont{\rmfamily}%
}
\theoremstyle{plain}{\theorembodyfont{\rmfamily}%

}

\definecolor{bsb}{rgb}{0.1,0.1,0.7}
\definecolor{captc}{rgb}{1,0.5,0.1}
\newcommand{\ccol}{\color{captc}}

\title{Combining local regularity estimation \\ and total variation optimization \\
for scale-free texture segmentation}

\makeatletter
\let\Title\@title
\makeatother

\author{
N. Pustelnik \thanks{Univ Lyon, Ens de Lyon, Univ Lyon 1, CNRS, Laboratoire de Physique, F-69342 Lyon, France (e-mail: firstname.lastname@ens-lyon.fr)}  \and H. Wendt \thanks{IRIT at INP-ENSEEIHT, University of Toulouse and CNRS} \and P. Abry$^*$ \and N. Dobigeon$^\dag$ }

\begin{document}
\maketitle

\begin{abstract}
Texture segmentation constitutes a standard image processing task, crucial for many applications.
\textcolor{black}
{The present contribution focuses on the particular subset of scale-free textures and its originality resides in the combination of three key ingredients:}
First, texture characterization relies on the concept of local regularity~;
Second, estimation of local regularity is based on new multiscale quantities referred to as wavelet leaders~;
Third, segmentation from local regularity faces a fundamental bias variance trade-off:
In nature, local regularity estimation shows high variability that impairs the detection of changes, while a posteriori smoothing of regularity estimates precludes from locating correctly changes.
{Instead, the present contribution proposes several variational problem formulations based on total variation and proximal resolutions that effectively circumvent this trade-off.}
\textcolor{black}{Estimation and segmentation performance for the proposed procedures are quantified and compared on synthetic as well as on real-world textures.}
\end{abstract}

\section{Introduction}

\noindent \textbf{Fractal based texture characterization.}
Texture characterization and segmentation are long-standing problems in Image Processing that received significant research efforts in the past and still attract considerable attention.
In essence, texture consists of a perceptual attribute.
It has thus no unique formal definition and has been envisaged using several different mathematical models, mostly relying on the definition and classification of either geometrical or statistical features, or primitives (cf. e.g., \cite{Haralick_R_1979_p-ieee_statistical_sat,Galloway_M_1975_j-cgip_texture_aug} and references therein for reviews).
Texture analysis can be performed using either parametric models (ARMA \cite{Deguchi_K_1986_p-fissf_two_dar}, Markov  \cite{Kashyap_R_1982_j-patt-rec-lett_texture_cuf}, Wold  \cite{Stitou_Y_2007_j-ieee-tsp_three_dti})
or non parametric approaches \textcolor{black}{(e.g., time-frequency and Gabor distributions \cite{Perona_P_1990_j-opt-soc-am_pre_tde,Yuan_J_2015_j-ieee-tip_fac_bts})}.
\textcolor{black}{Amongst this later class,} multiscale representations (wavelets, contourlet, \ldots) have repeatedly been reported as central in the last two decades (cf. e.g., \cite{Bouman_C_1994_j-ieee-tip_multiscale_rfm,Unser_M_1995_j-ieee-tip_texture_csu,Crouse_M_1998_j-ieee-tsp_wavelet_bss,Mallat_S_1999_ap_wavelet_awtosp}).
\textcolor{black}{They notably showed significant relevance for the large class of scale-free textures, often well accounted for by the celebrated fractional Brownian motion (fBm) model, on which the present contribution focuses.
Examples of such textures are illustrated in Fig.~\ref{fig:illprinciple}(a).}
\textcolor{black}{While scale free-like texture segmentation has been mainly conducted by exploiting the statistical distribution of the pixel amplitudes \cite{Guofang_X_2002_j-ieee-tmi_segmentation_umi,Pereyra_M_2012_j-ieee-tmi_segmentation_slu,Ni_K_2009_j-ijcv_local_hbs}, the present paper investigates the relevance of local regularity-based analysis.}

Deeply tied to multiscale analysis, the fractal and multifractal paradigms \cite{mandelbrot1983fractal,falconer2004fractal} have been intensively used for \textcolor{black}{scale-free} texture characterization (cf. e.g., \cite{Lopes_R_2011_j-patt-rec_local_fmf,Roux2013}).
\textcolor{black}{Scale-free} textures are essentially measured jointly at several scales, or resolutions. From the evolution across scales of such multiscale measures, semi-parametrically modeled as power-laws and  thus insensitive to texture resolution, scaling exponents are then extracted and used as characterizing features.
Such fractal features have been extensively used for texture characterization, notably for biomedical diagnosis (e.g., \cite{Benhamou_C_2001_j-bmr_fractal_art,Kestener_P_2004_j-ias_wavelet_bmf, Guo_Q_2009_j-ijcars_characterization_ctl,Lopes_R_2009_j-media_fractal_mar}), satellite imagery~\cite{Roux_S_2000_j-epjb_wavelet_bmm} or art investigations \cite{Abry_P_2013_j-sp_when_vgm,SPMagHPPCpaper}.

Often, in applications, it is assumed a priori that fractal properties are homogenous across the entire piece of texture to characterize, thus permitting reliable estimates of the fractal features (scaling exponents).
However, in numerous situations, images actually consist of several pieces, each with different textures, and hence different fractal properties.
Analysis becomes thus far more complicated as it requires to segment the texture into pieces (with unknown boundaries to be estimated) within which fractal properties can be considered homogeneous (i.e., fractal attributes are constant), yet unknown.\\

\noindent \textbf{Local regularity and H\"older exponent.}
To address such situations, the present contribution elaborates on the fractal paradigm, by a local formulation relying on the notion of  pointwise regularity.
The local regularity of a function (or sample path of a random field) at a given location $\xx\in \RR^2$ is most commonly quantified by the H\"older exponent $h(\xx)$ \cite{Jaffard_S_2004_p-spm_wavelet_tma}.
It is defined as the scaling exponent extracted from the a priori assumed power law dependence of the coefficients of a multiscale representation $X(a,\xx) $ (e.g., the modulus of wavelet coefficients \cite{Mallat_S_1999_ap_wavelet_awtosp}), across the scales $a$, in the limit of fine scales: $ X(a,\xx) \simeq  \eta(\xx) a^{h(\xx)}$ when $a\to 0$.
Theoretically, the collection of H\"older exponents $h(\xx)$ for all $\xx\in \RR^2$ yields access to the multifractal properties (and spectrum) of the texture, and thus consists of a multiscale higher-order statistics feature characterizing the texture \cite{Jaffard_S_2004_p-spm_wavelet_tma,Wendt_H_2007_j-ieee-spm_bootstrap_ema}. \\
Wavelet coefficients are the most popular multiscale representation used to perform fractal analysis.
Yet, it has recently been shown that, in the context of multifractal analysis and thus of local regularity estimation,
wavelet coefficients are significantly outperformed by \textit{wavelet leaders}, consisting of local suprema of wavelet coefficients~\cite{Jaffard_S_2004_p-spm_wavelet_tma,Wendt_H_2007_j-ieee-spm_bootstrap_ema,RouxSP2009,Pustelnik2013icassp}.
While the methods proposed in the present contribution could be used with any multiscale representation, reported and discussed results are thus explicitly obtained using wavelet leaders. \\
For each location $\xx$, $h(\xx)$ is classically estimated via a linear regression in $\log X(a,\xx)$ versus $\log a$ coordinates and is thus naturally framed into a classical bias-variance trade-off:
Pointwise estimates (relying only on the $X(a,\xx)$ at sole location $\xx$) show very large variances.
Conversely, averaging in space-windows either the $X(a,\xx)$ or directly the estimates of $h(\xx)$
results in significant bias.
In either case, the reliable and accurate detection and localization of actual changes in $h$ is precluded.
This explains why local regularity remains so far barely used for signal or texture characterization and segmentation (see, a contrario, \cite{PontTuriel2011,Pustelnik2013icassp, Nafornita_C_2014_p-icip_regularised_slh,Pustelnik_N_2014_p-icip_inverse_pfr, Nelson_J_2016_j-ieee-tip_sem_lse}).\\

\noindent \textbf{Goals, contributions and outline.}
In the present contribution, elaborating on preliminary attempts \cite{Pustelnik2013icassp,Pustelnik_N_2014_p-icip_inverse_pfr}, the overall goal is to enable the actual use of local regularity for performing the segmentation of \textcolor{black}{scale-free} textures into areas with piece-wise constant fractal characterization.  {This strategy has the great advantage of being fully nonparametric, since it does not require any explicit (statistical) modeling of the textures to be segmented}. To that end, it aims to marry wavelet-leader based estimation of local regularity (with no a priori smoothing) to segmentation procedures formulated as variational problems and constructed on \textit{total variation} (TV) optimization procedures and proximal based resolutions.
Wavelet-leader based estimation of H\"older exponents (as defined and detailed in Section \ref{sec-wl}) is first applied locally throughout the entire image.
Then, to go beyond a naive a posteriori smoothing and threshold-based segmentation procedure (as described in Section \ref{ss:hist}), three different TV-based segmentation procedures are theoretically devised:
 i) Local estimates of $h$ are subjected a posteriori to a TV based denoising procedure, followed by  a thresholding step for segmentation (cf. Section \ref{ss:tvden});
ii) Local estimation of $h$ is a priori embodied into the TV based procedure aiming to favor piecewise constant estimates, thresholding for segmentation is performed a posteriori (cf. Section \ref{ss:tvw});
iii) Local estimates of $h$ are subjected a posteriori to a TV based segmentation procedure, avoiding the denoising/thresholding steps (cf. Section \ref{sec-prox}).
This later procedure is inspired by the convex relaxation of the Potts model detailed in \cite{Chan_T_2006_siam_algorithms_fgm,Pock_Y_2009_p-cvpr_convex_rac}.

In Section \ref{sec-res}, the performance of the proposed \textcolor{black}{TV} segmentation procedures are compared and discussed for samples of synthetic textures with piece-wise constant $h$, obtained from a multifractional model, whose definition is customized by ourselves to achieve realistic textures (described in Section~\ref{sec:proc}). Several different geometries and different sets of values for the regularity $h$ of the different segments are considered. Furthermore, the impact of the TV-regularization parameter is investigated.
\textcolor{black}{These procedures are also compared against two texture segmentation procedures chosen because they are considered state-of-the art in the dedicated literature \cite{Arbelaez_P_2011_j-ieee-tpami_con_dhis,Yuan_J_2015_j-ieee-tip_fac_bts}.
The proposed approaches are further shown at work 
on samples chosen within well accepted references texture databases, such as the \textit{Berkeley Segmentation Dataset}.}
Synthesis and analysis procedures will be made publicly available at the time of publication. 
{Note that the present contribution significantly differs from the preliminary works \cite{Pustelnik2013icassp,Pustelnik_N_2014_p-icip_inverse_pfr} in that several methods involving TV are designed, and these methods are unified and compared on synthetic images and real-world textures.}

\section{H\"older exponent and wavelet leaders}
\label{sec-wl}

\subsection{H\"older exponent and Local regularity: Theory}

Let $f = \big(f(\underline{x})\big)$ with $\underline{x} = (x_1,x_2)\in \textcolor{black}{\Omega}$, 
denote the bounded 2D function (image) to analyze.
The local regularity around location $\xx_0$ is quantified using the so-called H\"older exponent $h(\xx_0)$,
formally defined as the largest $ \alpha >0$, such that there exist a constant $\chi>0$ and a polynomial $\mathcal{P}_{\xx_0}$ of degree lower than $\alpha$, such that $ \Vert f(\xx) - \mathcal{P}_{\xx_0}(\xx)\Vert \leq \chi \Vert\xx-\xx_0\Vert^\alpha$ in a neighborhood $\xx$ of $\xx_0$, where $\Vert \cdot\Vert$ denotes the Euclidean {norm} \cite{Jaffard_S_2004_p-spm_wavelet_tma}.
When $ h(\xx_0) $ is close to $0$, the image is locally highly irregular and close to discontinuous.
Conversely, large values of $h(\xx_0)$ correspond to locations where the field is locally smooth.
An example of a texture with piece-wise constant function $ h(\xx) $ is illustrated in Fig.~\ref{fig:illprinciple} (a).

Though the theoretical definition of the H\"older exponent above can be used for the mathematical study of the regularity properties of fields, it is also well-known that it turns extremely uneasy for practical purposes and for the actual computation of $h$ from real world data.
Instead, multiscale representations provide natural alternate ways for the practical quantification of $h$ \cite{Roux_S_2000_j-epjb_wavelet_bmm,PontTuriel2011,Wendt_H_2007_j-ieee-spm_bootstrap_ema,RouxSP2009}.
It is however nowadays well documented that the sole wavelet coefficients do not permit to accurately estimate local regularity \cite{Jaffard_S_2004_p-spm_wavelet_tma,Wendt_H_2007_j-ieee-spm_bootstrap_ema}.
Early contributions on the subject proposed to relate local regularity to the skeleton of the continuous wavelet transform \cite{Roux_S_2000_j-epjb_wavelet_bmm,Mallat_S_1999_ap_wavelet_awtosp}, which however does not permit to achieve estimation of $h$ at each location $\xx$ as targeted here.
Instead, it has been recently shown that an efficient estimation of $h$ can be conducted using \emph{wavelet leaders}\cite{Jaffard_S_2004_p-spm_wavelet_tma,Wendt_H_2007_j-ieee-spm_bootstrap_ema}. They are defined as local suprema of the coefficients of the discrete wavelet transform (DWT) and thus inherit their computational efficiency.

\subsection{H\"older exponent and wavelet leaders}

\noindent {\bf Wavelet coefficients.}
Let $\phi$ and $\psi$ denote respectively the scaling function and mother wavelet, defining a 1D multiresolution analysis \cite{Mallat_S_1999_ap_wavelet_awtosp}.
The mother wavelet $\psi$ is further characterized by an integer $N_\psi \geq 1$, referred to as the \emph{number of vanishing moments} and defined as: $\forall k = 0, 1, \, \ldots, N_\psi -1$, $\int |t|^k \psi(t) dt \equiv 0 $ and $\int |t|^{N_\psi} \psi(t) dt \neq 0$.
From these univariate functions, 2D wavelets are defined as
\begin{equation}
\begin{cases}
\psi^{(0)}(\xx) = \phi(x_1)\phi(x_2), & \psi^{(1)}(\xx)= \psi(x_1)\phi(x_2),\\
\psi^{(2)}(\xx)= \phi(x_1)\psi(x_2),  & \psi^{(3)}(\xx)= \psi(x_1)\psi(x_2).
\end{cases}
\end{equation}
{%
The 2D-DWT ($L^1$-normalized) coefficients of the image $f$ are defined as
\begin{equation}
Y^{(m)}_f(j,\kx) =  2^{-j}\langle f,\psi^{(m)}_{j,\kx} \rangle,
\end{equation}%
where $\{  \psi^{(m)}_{j,\kx}(\xx) = 2^{-j} \psi^{(m)}(2^{-j} \xx - \kx),  j \in \NN^*, \kx \in \NN^2, m = 0, 1, 2, 3 \}$ is the collection of dilated (to scales $a = 2^j$) and translated (to locations $\xx = 2^j \kx$, $\kx=(k_1,k_2)$) templates of $\psi^{(m)}(\xx)$.
}%
Interested readers may refer to \cite{Mallat_S_1999_ap_wavelet_awtosp} for further details. \\
\noindent {\bf Wavelet leaders.} The wavelet leader $L_f(j,\kx)$, at scale $2^j$ and location $\xx = 2^{j}\kx$, is defined as the local supremum of all wavelet coefficients $Y^{(m)}_f(j',\kx') $ taken across all finer scales $2^{j'} \leq 2^j$, within a spatial neighborhood  \cite{Jaffard_S_2004_p-spm_wavelet_tma,Wendt_H_2007_j-ieee-spm_bootstrap_ema,RouxSP2009}
\begin{equation}
\label{eq:defwl}
L_f^{(\gamma)}(j,\kx)=\sup_{\substack{m = \{1, 2, 3\} \\\,  \lambda_{j',\kx'}\subset 3\lambda_{j,\kx}}}  | 2^{j\gamma} Y_f^{(m)}(j',\kx')|
\vspace{-0.2cm}
\end{equation}
where
\begin{equation}
\label{equ-lamlx}
\begin{cases}
\lambda_{j,\kx}&=[\kx 2^j,(\kx+1)2^j),\\
3\lambda_{j,\kx}&=\bigcup_{p\in\{-1,0,1\}^2}\lambda_{j,\kx+p}.
\end{cases}
\end{equation}
The additional positive real parameter $\gamma$ can be tuned to ensure minimal regularity conditions for the case where the image $f$ to analyze can not be modeled as a strictly bounded 2D-function.
It is set to $\gamma = 1$ for the present contribution and not further discussed.
Interested readers are referred to \cite{Wendt_H_2007_j-ieee-spm_bootstrap_ema,RouxSP2009} for the details on the role and impact of parameter $\gamma$.

It has been proven in \cite{Jaffard_S_2004_p-spm_wavelet_tma} that the H\"older exponent $h(\xx) $ at location $\xx$ is measured by wavelet leaders as
\begin{equation}
\label{eq:scaleprop}
L_f^{(\gamma)}(j,\kx) \simeq \eta(\xx) 2^{j(h(\xx)+\gamma)} \quad  \mbox{when} \quad  2^j \rightarrow 0
\end{equation}
for $ \xx \in \lambda_{j,\kx}$, {provided that} $N_\psi $ is strictly larger than $h(\xx)$.

Note that \eqref{eq:scaleprop} implies that $h$ can be estimated by linear regressions as the slope of $\log L_f^{(\gamma)}$ versus $j$, cf., \eqref{eq1}-\eqref{eq:cons} below (similarly, $\log \eta$ could be obtained as the intercept, yet does not bear any information on local regularity and will hence not be further considered here).

\subsection{H\"older exponent estimation}
\label{ssec-hee}
In the present contribution, the estimation of the H\"older exponent is only performed using wavelet leaders $ L_f^{(\gamma)}(j,\kx) $, as preliminary contributions \cite{Wendt_H_2007_j-ieee-spm_bootstrap_ema,RouxSP2009,Pustelnik2013icassp,Pustelnik_N_2014_p-icip_inverse_pfr} report that wavelet leader based estimation outperforms those based on other multiresolution quantities.
To indicate that the estimation of $h$ and the segmentation procedures proposed in Section~\ref{sec-tv} below could be applied using any other multiresolution quantity, e.g., the modulus of the 2D-DWT coefficients, a generic notation $X(j,\kx)$
is used instead of the specific $ L_f^{(\gamma)}(j,\kx) $.
Yet, all numerical results in Section \ref{sec-res} are obtained with wavelet leaders, $X(j,\kx)=L_f^{(\gamma)}(j,\kx) $.

{%
In the discrete setting, H\"older exponents are estimated for the locations $\xx=2\kx$ associated with the finest scale $j=1$, and we make use of the notation $\underline{ \underline{ \;\cdot\;}}$ when we are dealing with matrices, rather than with matrix elements, e.g.,
$$
\underline{\underline{{h}}}=\Big({h}(\kx)\Big)_{\kx\in \mathcal{K}}, \, \mathcal{K} = \{1,\ldots,N_1\}\times \{1,\ldots,N_2\}.
$$
As a preparatory step, the multiresolution quantity $\underline{\underline{X}}$ is up-sampled in order to obtain as many coefficients at scales $j>1$ as at the finest scale $j=1$
\begin{equation}
\label{eq1}
\widetilde X(j,(k_1,k_2)) = X(j,(\lceil 2^{-j}k_1\rceil, \lceil 2^{-j}k_2\rceil))
\end{equation}
with $1\leq k_1\leq N_1$, $1\leq k_2\leq N_2$.
With a slight abuse of notation, we will continue writing $\underline{\underline{X}}$ for the upsampled multiresolution coefficients $\underline{\underline{\tilde X}}$.
}

The relation \eqref{eq:scaleprop} can obviously be rewritten as
\begin{equation}
\label{equ-lnLh}
\log_2 X(j,\kx)  \simeq  jh(\kx) + \log_2 \eta(\kx)
\end{equation}
which naturally leads to the use of (weighted) linear regressions across scales for the estimation of $h(\xx)$
\begin{align}
\label{equ-estimh}
\widehat h(\kx) &=   \sum_{j} w(j,\kx) \log_2 X(j,\kx).
 \end{align}
Combining \eqref{equ-lnLh} and \eqref{equ-estimh} above shows that, for each location $\kx$, the weights $w(j,\kx)$
must satisfy the following constraints to ensure unbiased estimation \cite{AFTV2000}
\begin{equation}
\label{eq:cons}
\sum_{j} w(j,\kx) = 0\quad \mbox{and}\quad
\sum_{j} j w(j,\kx) = 1.
\end{equation}
Though unusual, let us note that the weights $w(j,\kx)$ can in principle depend on location $\kx$.
This will be used in the segmentation procedure defined in Section~\ref{ss:tvw}.
An illustration of such unbiased estimates, with a priori chosen $w(j,\kx)$ that do not depend on location $\kx$, is presented in Fig.~\ref{fig:illprinciple}~(c).

\subsection{Piece-wise constant local regularity synthetic processes}
\label{sec:proc}
To illustrate the behavior of the segmentation procedures proposed in Section~\ref{sec-tv} as well as to quantify and compare segmentation performances from the different procedures, use is made of realizations of synthetic random fields with known and controlled piece-wise constant local regularity.
These are constructed as 2D multifractional Brownian fields \cite{Ayache_A_2000_p-icassp_covariance_smb}, which are among the most widely used models for mildly evolving local regularity.
Their definition has been slightly modified here to ensure the realistic requirement of homogeneous local variance across the entire image
\begin{equation}
\label{equ-2Dmbm}
f(\xx) = C(\xx)  \int_{\textcolor{black}{\Omega}} \frac{e^{\imath \xx \kk -1}}{|\kk|^{h(\xx) + \frac{1}{2}}} dW(\kk)
\end{equation}
where $dW(\kk)$ is 2D Gaussian white noise and $h(\xx)$ denotes the prescribed H\"older exponent function.
The normalizing factor $C(\xx)$ ensures that the local variance of $f$ does not depend on the location $\xx$.
The details of the models do not matter much for the present work, as we are only targeting the control of local regularity of the field $f(\xx)$.
In the present contribution, the function $h(\xx)$ is chosen as piece-wise constant.
Numerical procedures permitting the actual synthesis of such fields have been designed by ourselves.
A typical sample field of such a process is shown in Fig.~\ref{fig:illprinciple}~(a). \\

\section{Local regularity based image segmentation}
\label{sec-tv}
In this section, we will detail the proposed procedures for segmentation in homogeneous pieces of textures with constant local regularity.

\subsection{Local smoothing}
\label{ss:hist}
A straightforward attempt for obtaining labels consists in thresholding the histogram of pointwise estimates obtained with \eqref{equ-estimh} and \eqref{eq:cons}. However,  such a solution yields estimates with prohibitively large variance. This prevents the identification of modes in the histogram that would correspond to the different zones of constant local regularity, see
Fig. \ref{fig:illprinciple} (c)  for an illustration.
{The variance can be reduced a posteriori by  a \emph{local spatial smoothing} with a convolution filter  $\underline{\underline{g}}$:
$$
\widehat{\underline{\underline{h}}}^{\textrm{S}} = \underline{\underline{g}} \ast\underline{\underline{\widehat{h}}}
$$
where $\underline{\underline{\widehat{h}}}$ denotes the estimate described in Section~\ref{ssec-hee}.
For instance, $\underline{\underline{g}}$ can model a local spatial average.} {In this work, we will consider a Gaussian smoothing parametrized by its standard deviation $\sigma$. Note that particular cases of this smoothing can be expressed as a variational formulation
\begin{equation}
\label{eq:tv}
\underline{\underline{\widehat{h}}}^{\textrm S} \!\!\!= \!\!\!\underset{{\underline{\underline{h}}\in \RR^{N_1\times N_2}}}{\arg\min}\Bigg\{\sum_{\kx\in \mathcal{K}}\Big( \sum_{j=j_1}^{j_2} \!\!w(j,\kx) \log_2 X(j,\kx) - {{h}}(\kx)  \Big)^2 + \lambda   \Vert \Gamma\underline{\underline{h}}\Vert_{\text F}^2\Bigg\}
\end{equation}
where $\Vert \cdot \Vert_{\text{F}}$ denotes the Frobenius norm and the transform $\Gamma$ {and the regularization parameter $\lambda>0$ are} related to the convolution kernel $\underline{\underline{g}}$. In particular, when $\lambda=0$, the smoothed solution $\widehat{\underline{\underline{h}}}^{\textrm{S}}$ reduces to its non-smoothed counterpart $\underline{\underline{\widehat{h}}}$. An example of such an estimate with $\sigma=10$ is given in Fig.~\ref{fig:illprinciple}~(d). 
}
Clearly, the variance of $\widehat{\underline{\underline{h}}}^{\textrm{S}}$ is smaller than the variance of $\underline{\underline{\widehat{h}}}$ (see Fig. \ref{fig:illprinciple} (c) and (d)). Yet, it remains hard to identify separate modes in the histogram. What is more, the local averaging introduces bias at the edges of areas with constant $h$ values, hence prevents any accurate localization of the regularity changes in the image.

\subsection{TV denoising}
\label{ss:tvden}
To overcome these difficulties, we propose the use of \textit{total variation} (TV) based optimization procedures, naturally favoring sharp edges, instead of local spatial averages.
It is well known that the bounded total variation space, that is the space of functions with bounded $\ell_1$-norm of the gradient, allows to remove undesirable oscillations while it preserves sharp features. Rudin et al.~\cite{Rudin_L_1992_tv_atvmaopiip} formalize this recovery problem as a variational approach involving a non-smooth functional referred to as total variation. This has been widely used in image processing for image quality enhancement~\cite{Rudin_L_1992_tv_atvmaopiip,Chambolle_A_2004_jmiv_TV_aaftvmaa}
although it has been noted that it is not always well suited for restoration purposes due to the piece-wise constant nature of the restored images.
In the present context of detection of local regularity changes, precisely such a piece-wise constant behavior of the solution is desired.

{The corresponding minimization problem reads
\begin{equation}
\label{eq:tv}
\underline{\underline{\widehat{h}}}^{\textrm{TV}} \!\!\!= \!\!\!\underset{{\underline{\underline{h}}\in \RR^{N_1\times N_2}}}{\arg\min}\Bigg\{\frac{1}{2}\sum_{\kx\in \mathcal{K}}\Big( \sum_{j=j_1}^{j_2} \!\!w(j,\kx) \log_2 X(j,\kx) - {{h}}(\kx)  \Big)^2 + \lambda\text{TV}(\underline{\underline{h}})\Bigg\}
\end{equation}
where 
\begin{align*}
\text{TV}(\underline{\underline{h}}) 
&=  {\small{\sum_{k_1=1}^{N_1-1}\sum_{k_2=1}^{N_2-1}\sqrt{\big((D_1\underline{\underline{h}})(k_1,k_2)\big)^2 + \big((D_2 \underline{\underline{h}})(k_1,k_2)\big)^2}}},\\
&(D_1 \underline{\underline{h}})(k_1,k_2) = h(k_1+1,k_2+1) - h(k_1+1,k_2),\\
&(D_2 \underline{\underline{h}})(k_1,k_2) = h(k_1+1,k_2+1) - h(k_1,k_2+1).
\end{align*}
The weights are a priori fixed,  are independent of $\kx$, $w(j,\kx)=w(j)$, and satisfy the constraints \eqref{eq:cons}.} 
Here, $\lambda>0$ 
models the regularization parameter that tunes the piece-wise constant behavior of the solution.
\begin{figure*}[t]
\setlength{\tabcolsep}{0.2pt}
\centering
\hspace{-0.3cm}\begin{tabular}{cccccc}
\includegraphics[height =2.8cm]{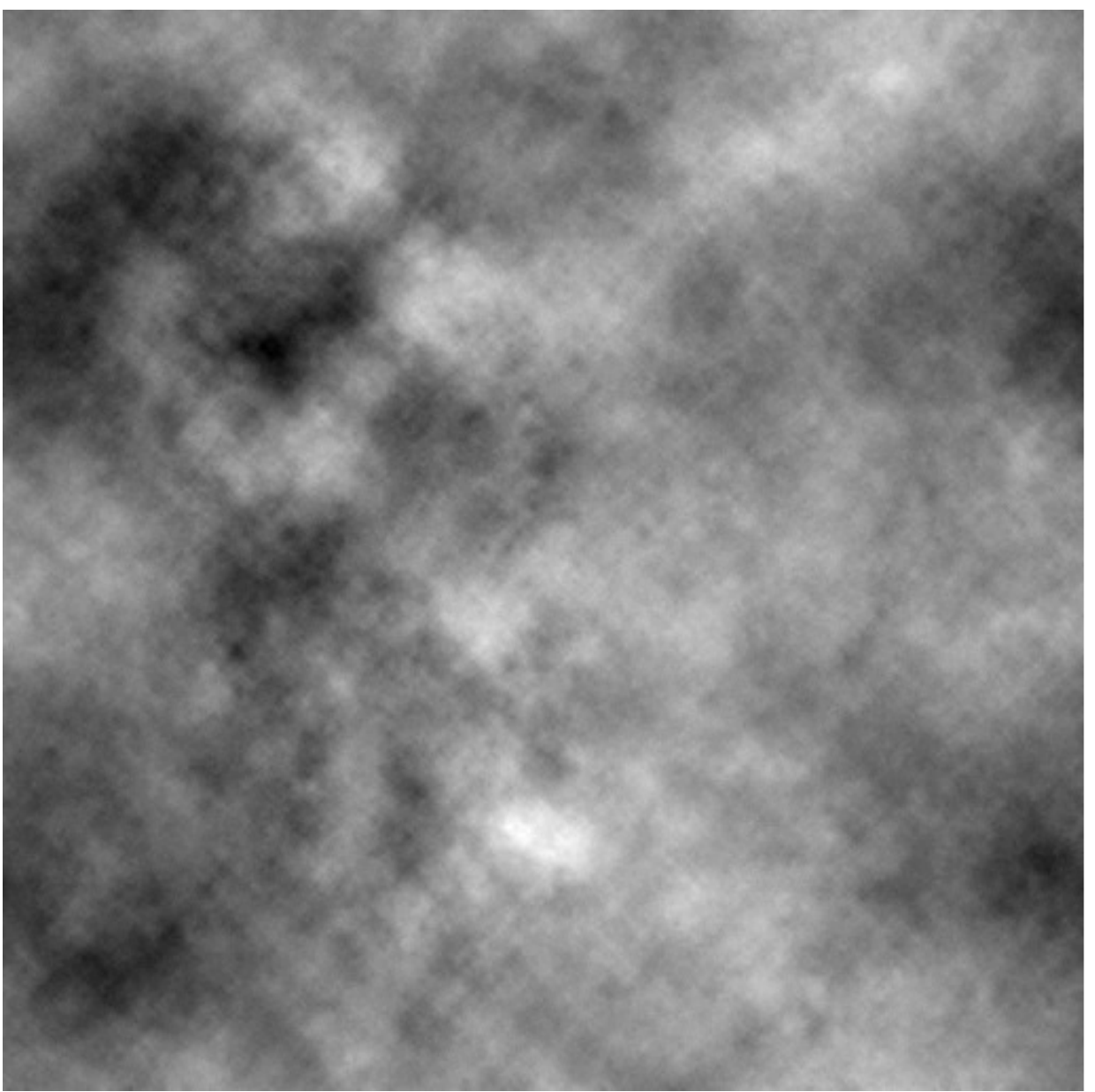} &
\includegraphics[height =2.8cm]{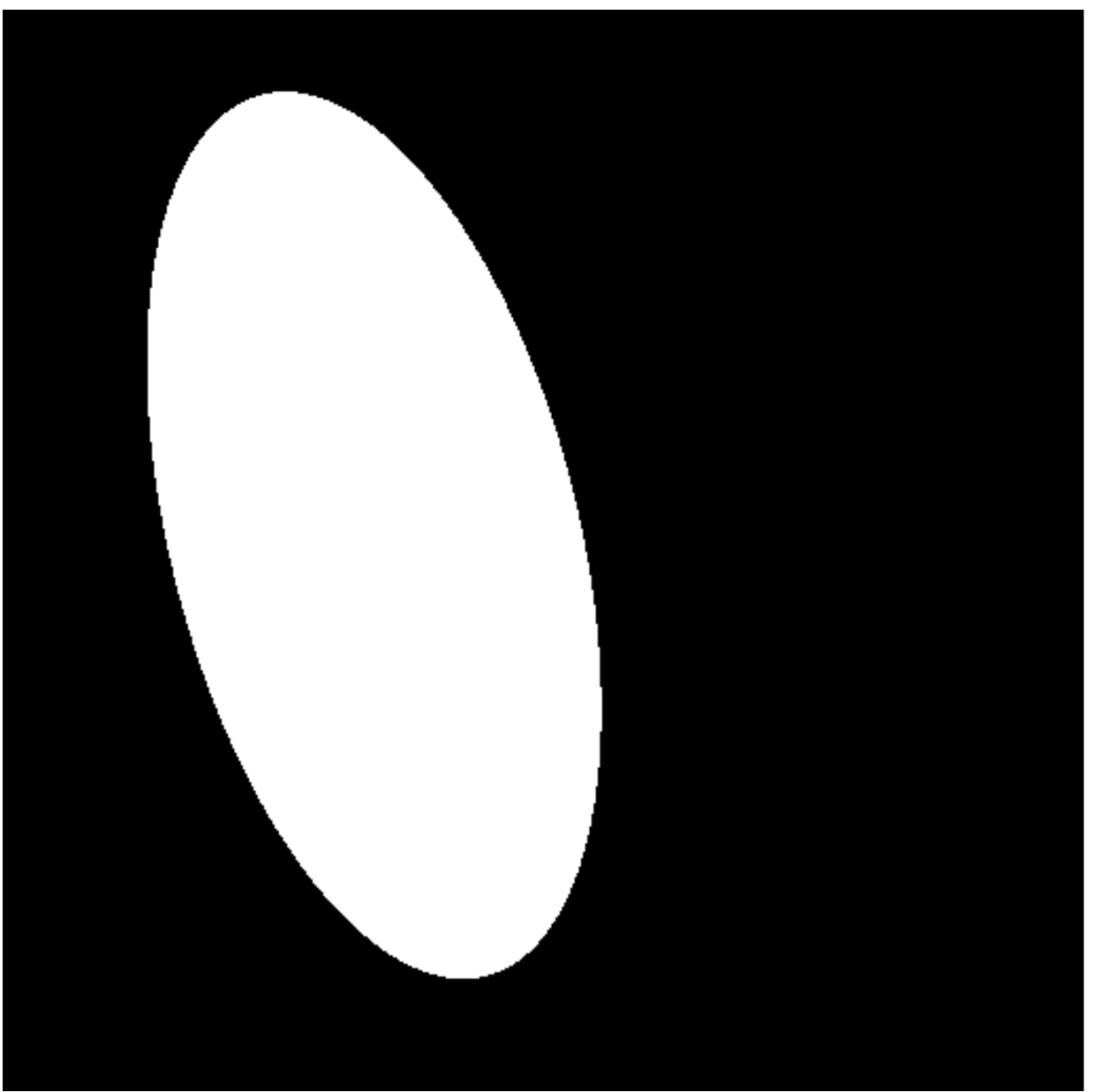}&
\includegraphics[height =2.8cm]{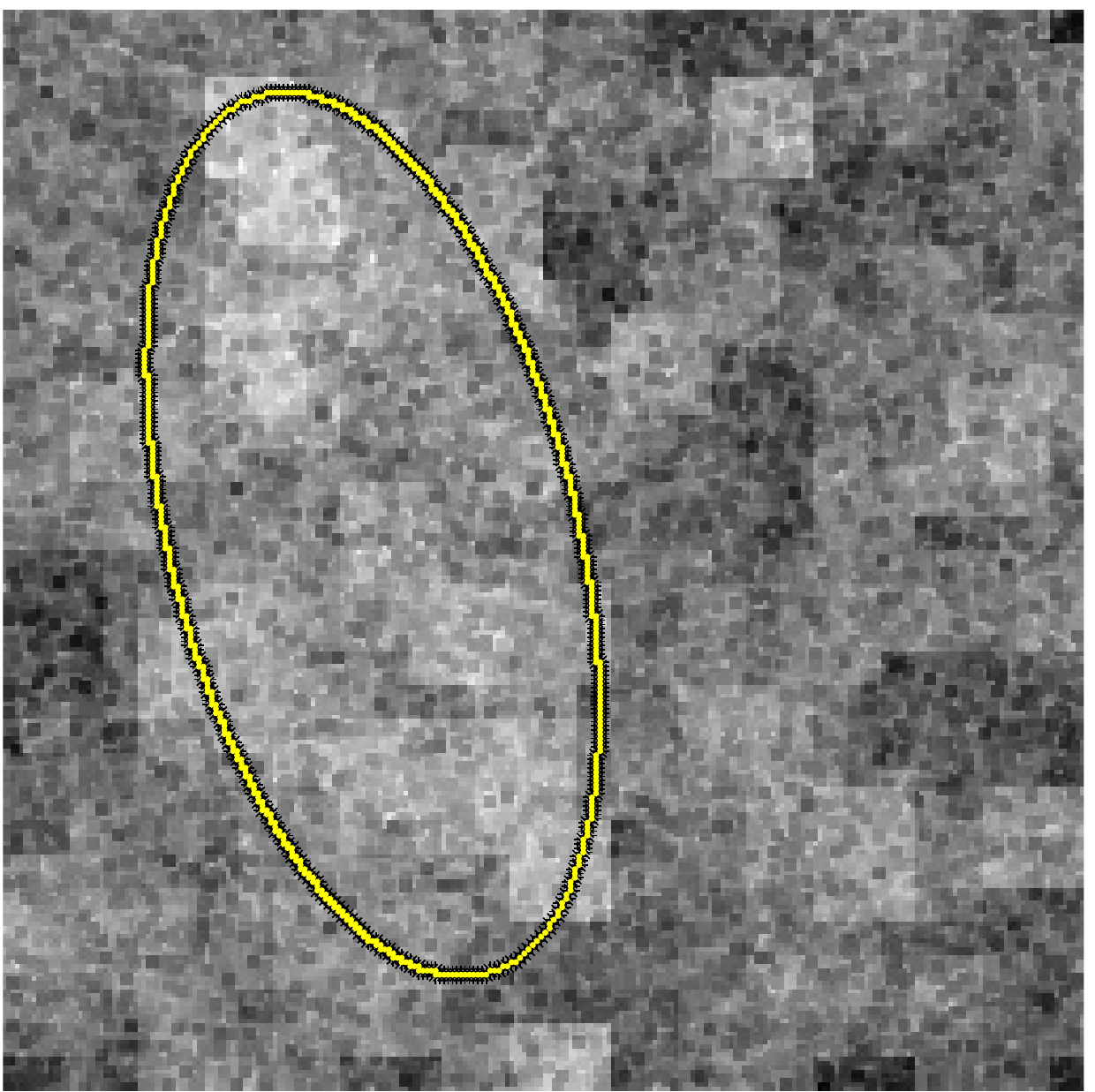} &
\includegraphics[height =2.8cm]{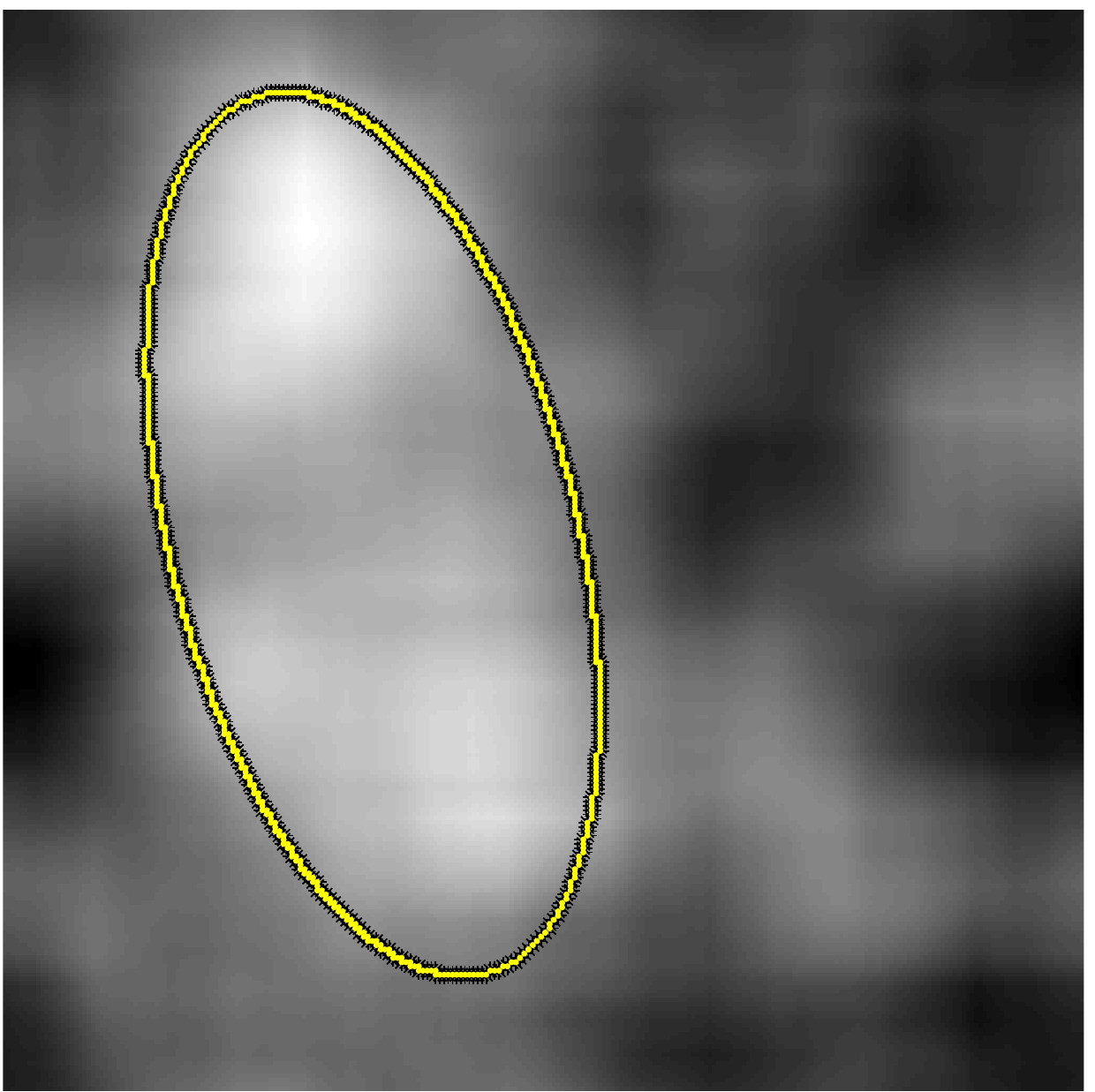} &
\includegraphics[height =2.8cm]{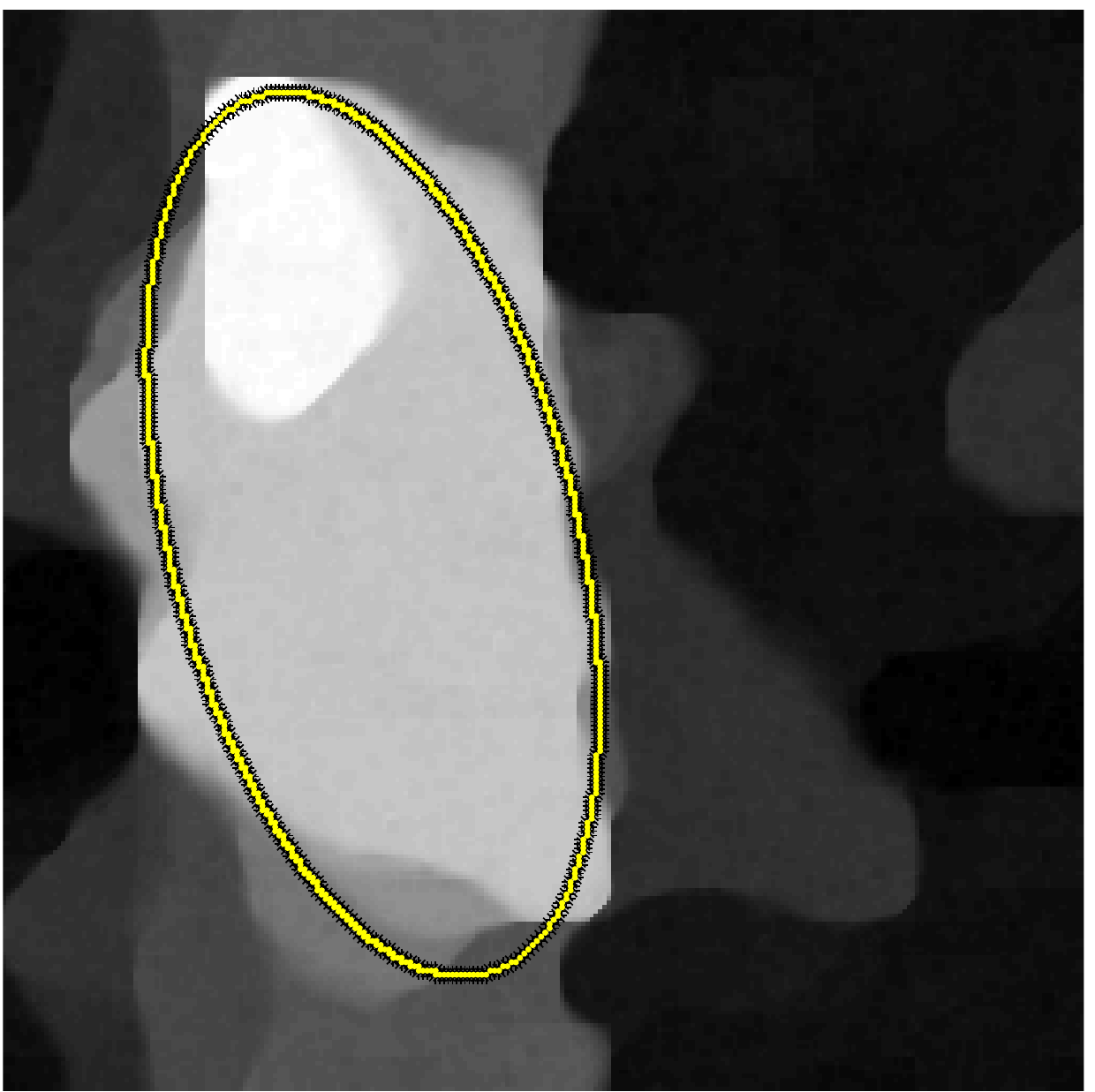} &
\includegraphics[height =2.8cm]{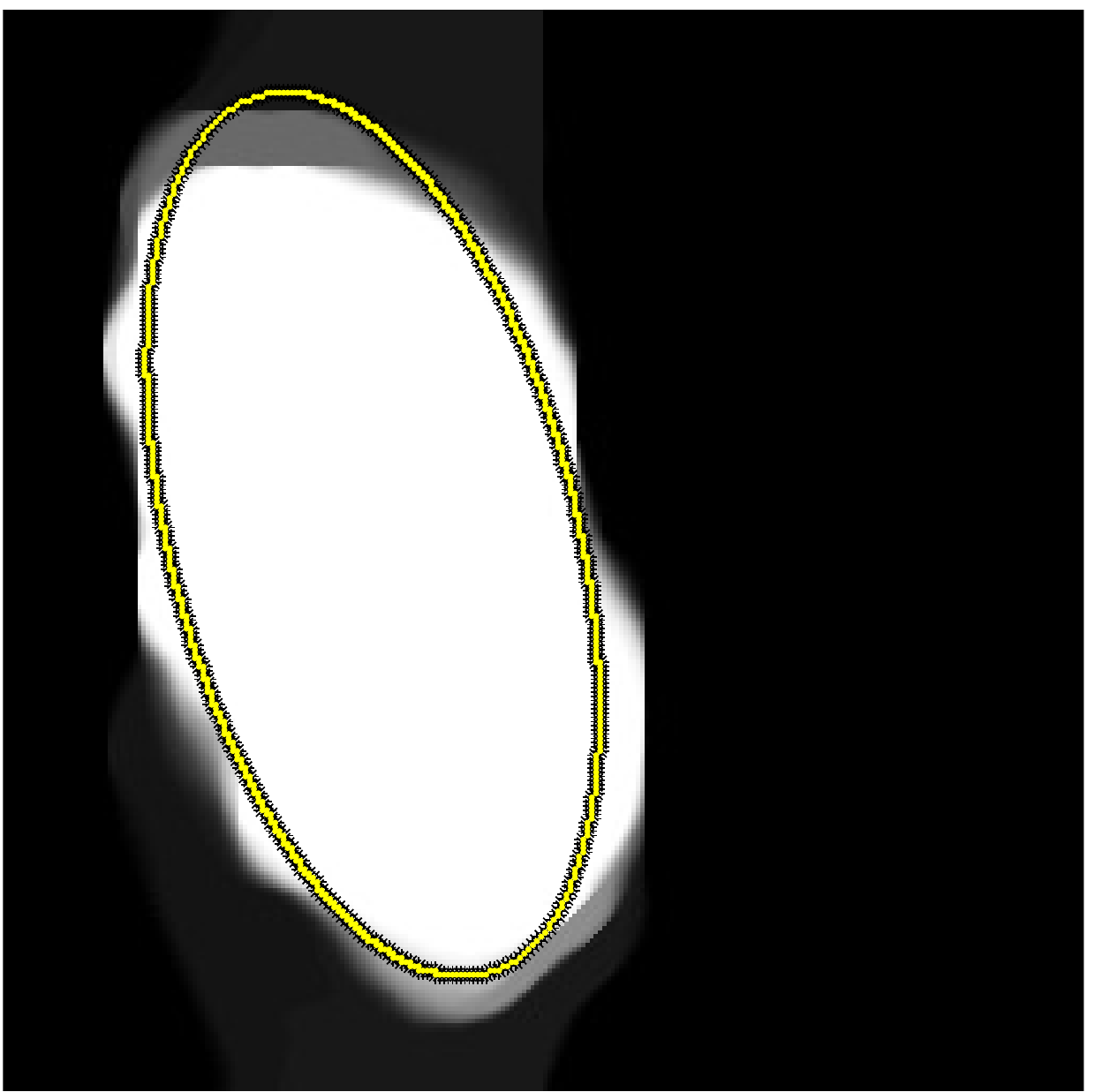}\vspace{-28mm}\\
&&\ccol\hfill\small  estimate&\ccol\hfill\small estimate&\ccol\hfill\small estimate&\ccol\hfill\small estimate
\vspace{22mm}\\
&&&&&\\
\includegraphics[height =2.4cm,width =2.8cm]{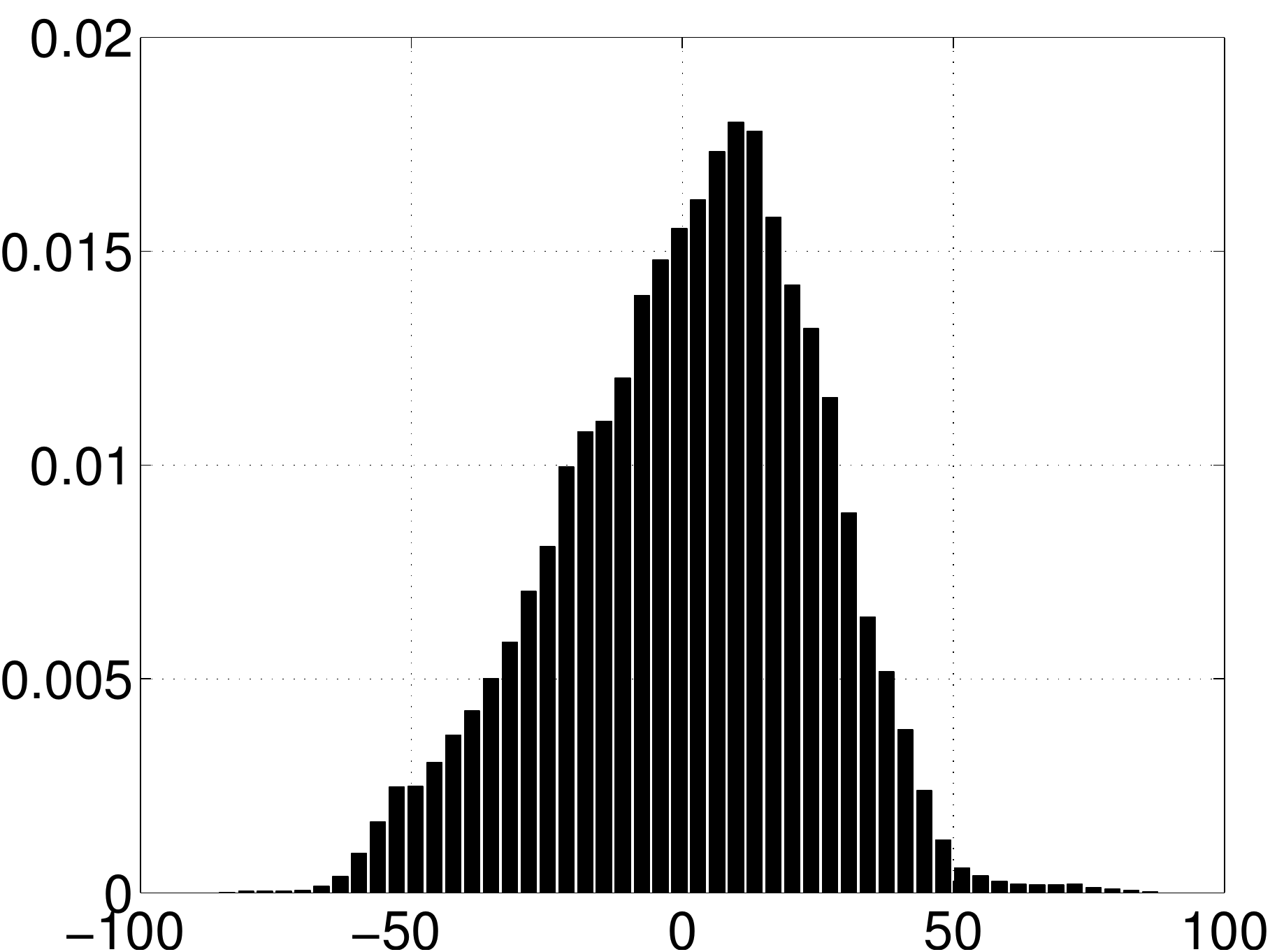} &
\includegraphics[height =2.4cm,width =2.8cm]{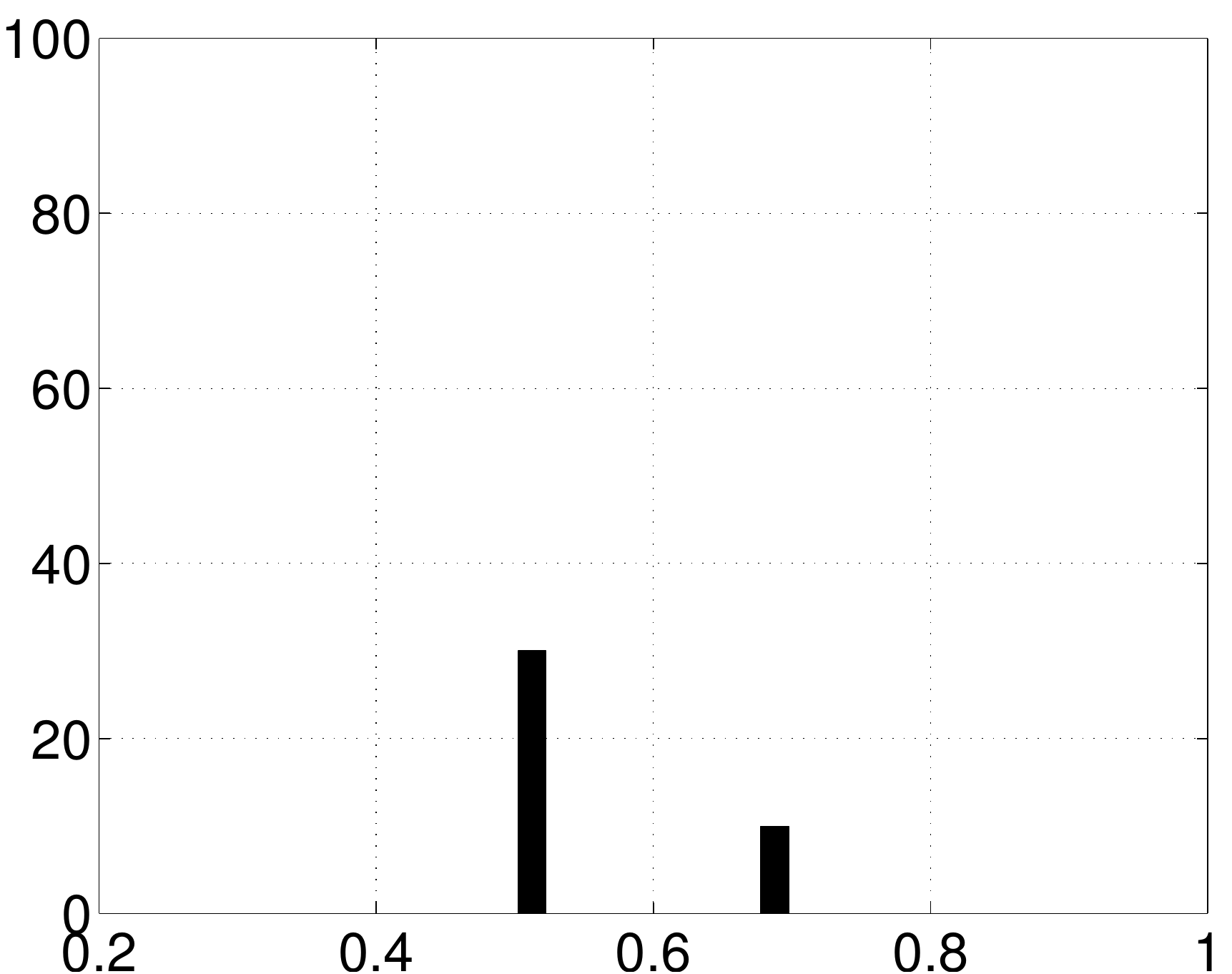}&
\includegraphics[height =2.4cm,width =2.8cm]{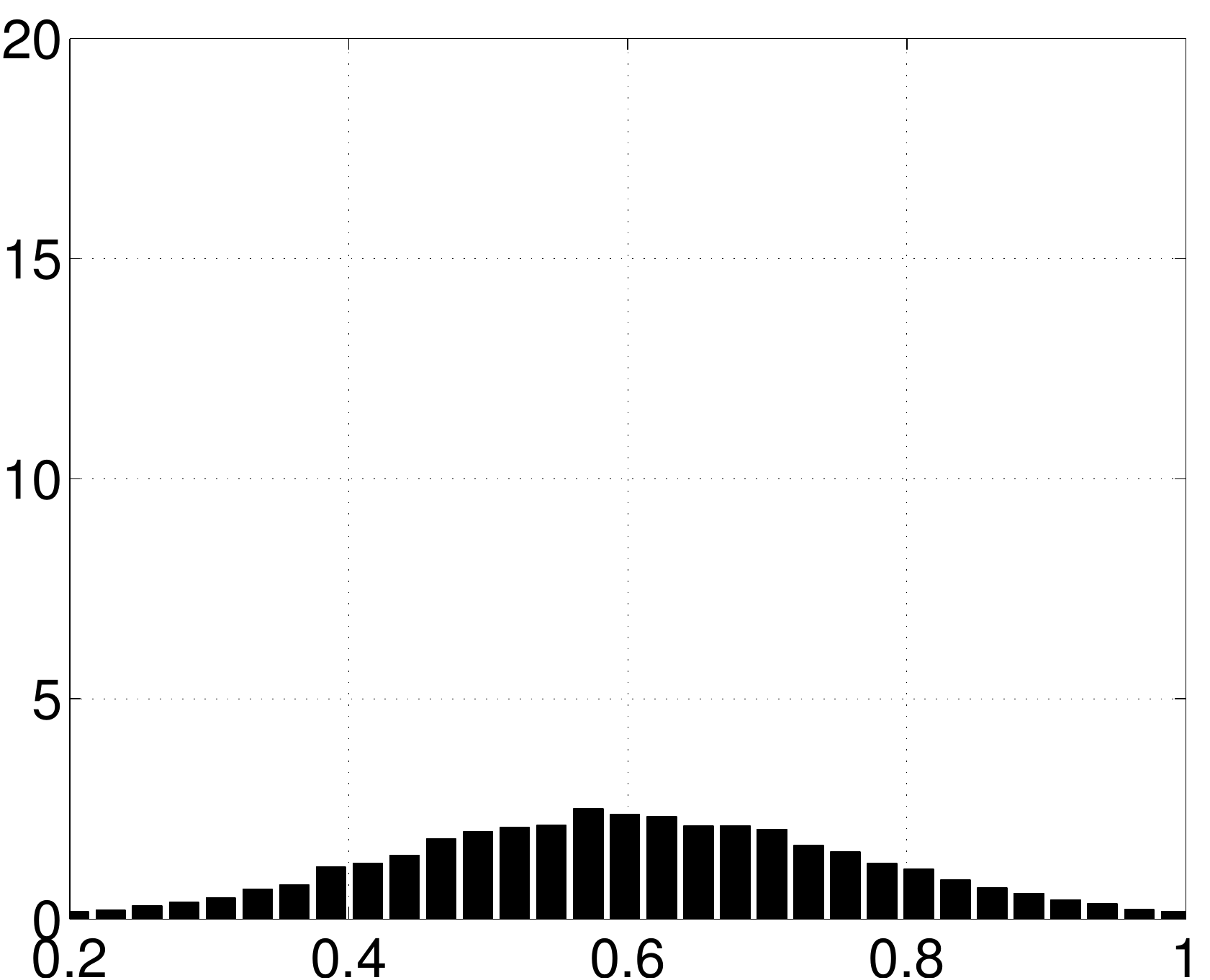} &
\includegraphics[height =2.4cm,width =2.8cm]{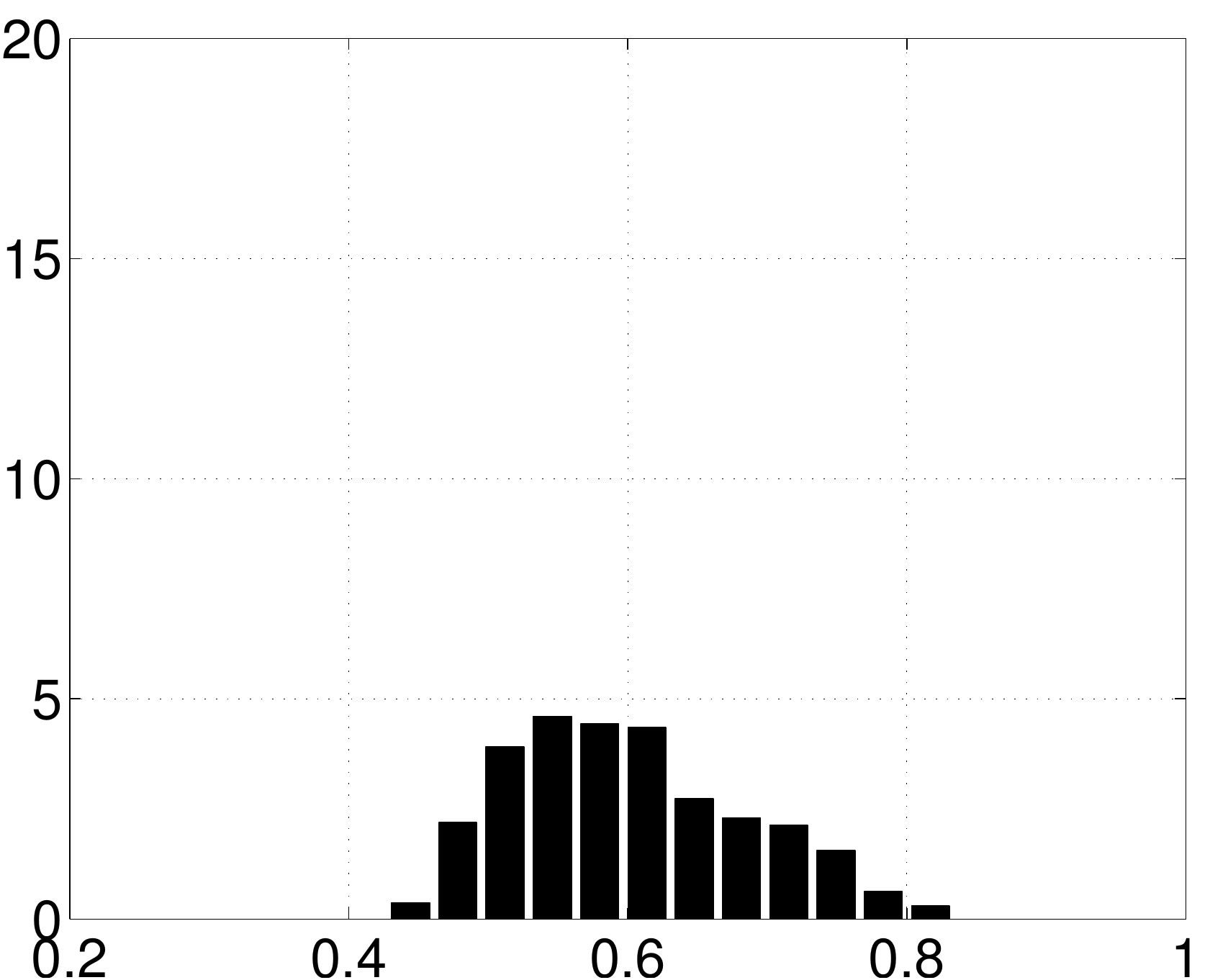} &
\includegraphics[height =2.4cm,width =2.8cm]{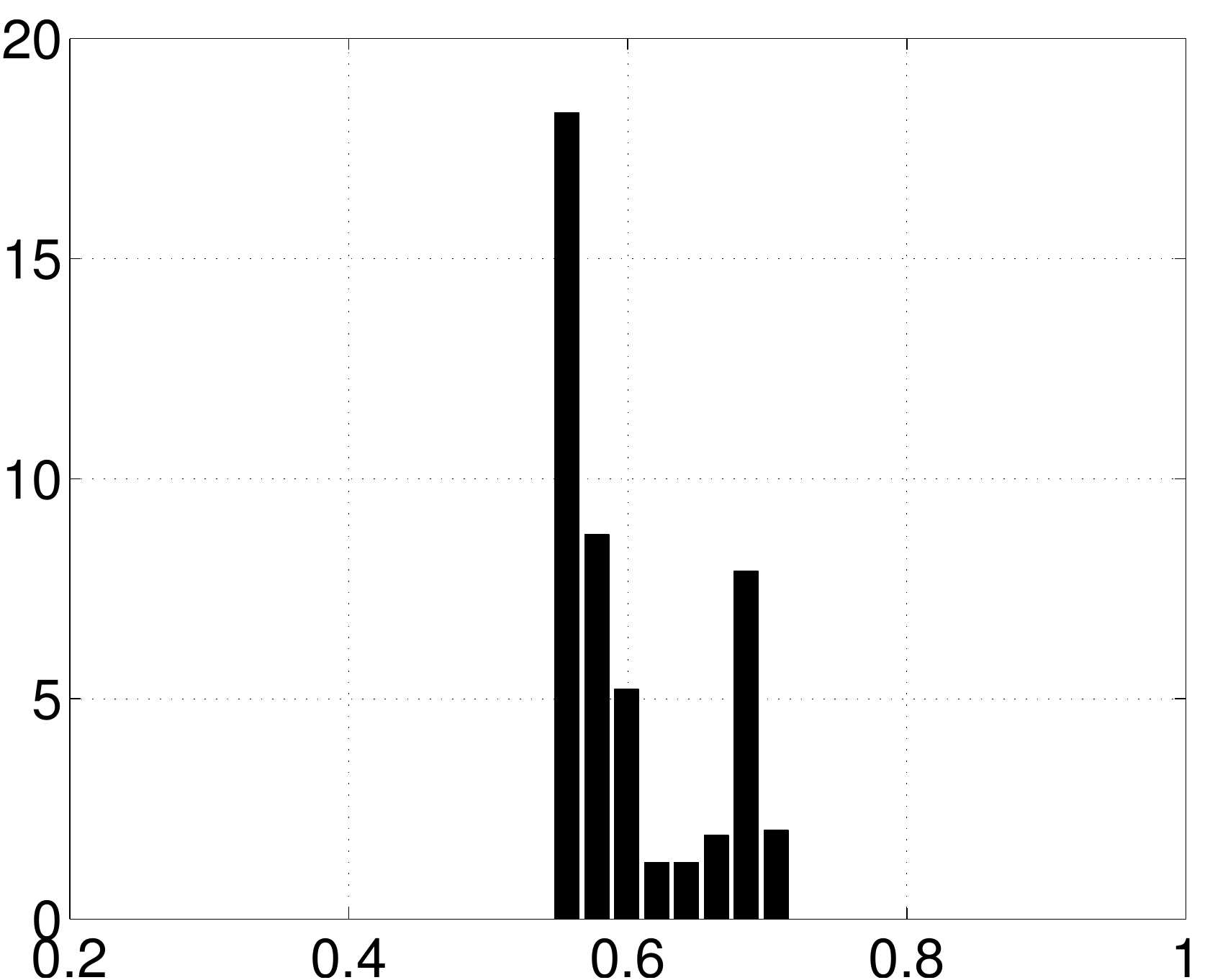} &
\includegraphics[height =2.4cm,width =2.8cm]{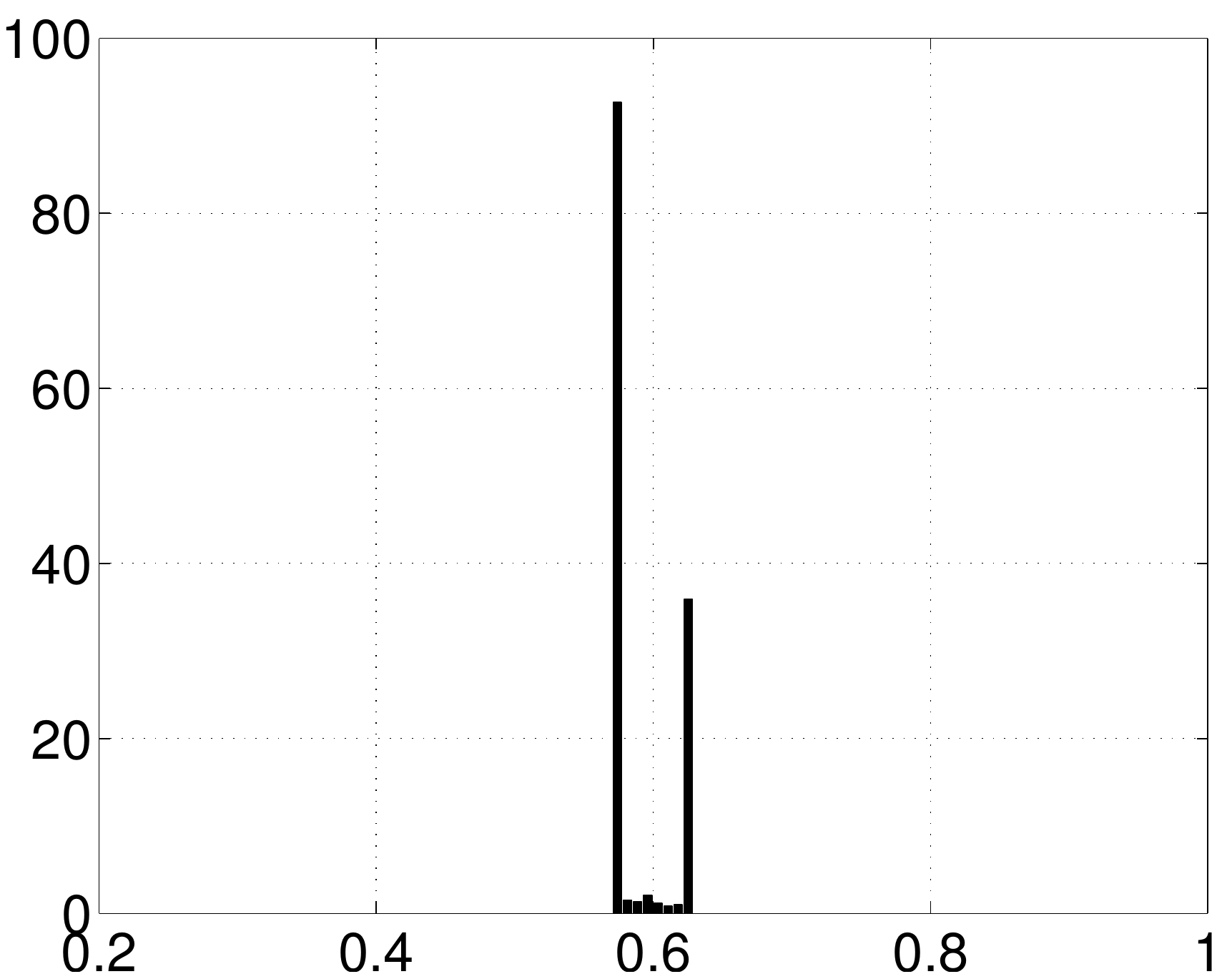}\vspace{-24mm}\\
\ccol\hfill\small  histogram&\ccol\hfill\small  histogram&\ccol\hfill\small  histogram&\ccol\hfill\small  histogram&\ccol\hfill\small  histogram&\ccol\hfill\small  histogram\vspace{19mm}\\
(a) Data $\underline{\underline{f}}$ & (b) Original $\underline{\underline{h}}$ &(c) $\widehat{\underline{\underline{h}}}$ & (d) $\widehat{\underline{\underline{h}}}^{\textrm{S}}$ & (e)  $\widehat{\underline{\underline{h}}}^{\textrm{TV}}$ & (f) $\widehat{\underline{\underline{h}}}^{\textrm{TVW}}$\vspace{-1mm}\\
\end{tabular}%
\caption{Illustration of the local regularity estimation from a multifractional data. Top row: (a) image $\underline{\underline{f}}$; (b) the original local regularity $\underline{\underline{h}}$ from which $\underline{\underline{f}}$ is generated, the area in black (resp. white) corresponds to a local regularity of $0.5$ (resp. $0.7$); (c) an estimation based on wavelet leaders; (d) smooth estimate of $\underline{\underline{\widehat{h}}}$ described  in Section~\ref{ss:hist}; (e) estimation using the method described in Section~\ref{ss:tvden}; (f) estimation based on the method described in Section~\ref{ss:tvw}. Bottom row: histograms corresponding  to top row.\label{fig:illprinciple}}
\end{figure*}
Several techniques have been proposed in the literature to solve \eqref{eq:tv}, see for instance~\cite{Chambolle_A_2004_jmiv_TV_aaftvmaa,Condat_L_2012}.
A forward-backward algorithm \cite{Combettes_PL_2005_j-siam-mms_Signal_rbpfbs} applied on the dual formulation of~\eqref{eq:tv} is used here.

An example of $\widehat{\underline{\underline{h}}}^{\textrm{TV}}$ is displayed in Fig.~\ref{fig:illprinciple}~(e) for $\lambda = 6$.
It is piecewise constant and provides a satisfactory estimate of the true regularity displayed in Fig.~\ref{fig:illprinciple}~(b).
Furthermore, the histogram of the estimates $\widehat{\underline{\underline{h}}}^{\textrm{TV}}$ is pronouncedly peaked. Thresholding hence enables labeling for a preset number $Q$ of classes.

\subsection{Joint estimation of regression weights and local regularity}
\label{ss:tvw}
The solution $\widehat{\underline{\underline{h}}}^{\textrm{TV}}$ described in the previous section is a two-step procedure that addresses the bias-variance trade-off difficulty by:
(i)~ computing unbiased estimates of the H\"older exponents $\widehat{\underline{\underline{h}}}$ from $\underline{\underline{X}}$ using \eqref{equ-estimh} with fixed pre-defined weights $\underline{\underline{w}}$, and (ii)~ extracting areas with constant H\"older exponent based on these estimates $\widehat{\underline{\underline{h}}}$ using a variational procedure relying on TV.
In this section, we propose a one-step procedure that directly yields piece-wise constant local regularity estimates $\widehat{\underline{\underline{{h}}}}$ from the multiresolution quantities $\underline{\underline{X}}$. The originality of this approach resides, on one hand, in the use of a criterion that involves directly $\underline{\underline{X}}$, instead of the intermediary local estimates $\widehat{\underline{\underline{h}}}$, and a TV-regularization; on other hand, in the fact that the weights $\underline{\underline{w}}$ are jointly and simultaneously estimated instead of being fixed a priori.
To this end, the weights $\underline{\underline{w}}$ are subjected to a penalty for deviation from  the hard constraints \eqref{eq:cons}.

\noindent \textbf{Problem formulation.} \quad
The estimation \eqref{equ-estimh} underlies a linear inverse problem in which the estimate $\underline{\underline{\widehat{h}}}$ of $\underline{\underline{h}}$ needs to be recovered from the logarithm of the multiresolution quantities $\underline{\underline{X}}$ of the image $\underline{\underline{f}}$.
This inverse problem resembles a denoising problem, yet including the additional challenge that a part of the observations (the regression weights $\underline{\underline{w}}$) are unknown and governed by the constraints~\eqref{eq:cons}, yielding the following convex minimization problem:\vspace{-0.6cm}

{\begin{multline}
\label{eq:problem}
\!\!\!\big(\widehat{\underline{\underline{h}}}^{\textrm{TVW}} \!\!\!\!,\widehat {\underline{\underline{w}}}\big) = \!\!\!\!\!\!\underset{\substack{\underline{\underline{h}}\in\RR^{N_1\times N_2},\\\underline{w}\in\RR^{J\times N_1\times N_2}}}{\arg\min} \, \!\!\!\!\Bigg\{\!\!\sum_{\kx\in \mathcal{K}}\Big( \sum_{j=j_1}^{j_2} \!\!w(j,\kx) \log_2 X(j,\kx) - {{h}}(\kx) \Big)^2\\
\!\!\!+ \lambda \mbox{TV}(\underline{\underline{h}})
 + \eta_1\sum_{\kx\in \mathcal{K}} d_{C_1}(w(\cdot,\kx)) + \eta_2\sum_{\kx\in \mathcal{K}} d_{C_2}(w(\cdot,\kx))\Bigg\}\!\!\!\!
\end{multline}}

\noindent where $J=j_2-j_1+1$. The functions $d_{C_1}$ and $d_{C_2}$ are distances
to the convex sets
{\begin{align}
\label{eq:c1c2}
C_1 &= \{(\omega({j_1}),\ldots,\omega({j_2})) \in \RR^J \,\vert \, \sum_{j=j_1}^{j_2} \omega(j) = 0\},\\
C_2 &= \{(\omega({j_1}),\ldots,\omega({j_2})) \in \RR^J \,\vert \, \sum_{j=j_1}^{j_2} j\omega(j) = 1\}
\end{align}
and are defined as
$$
(\forall \underline{v}\in \RR^J)(\forall i=1,2)\quad d_{C_i}(\underline{v}) = \Vert \underline{v} - P_{C_i}(\underline{v})\Vert_2
$$
where $P_{C_i}(\underline{v}) =\arg\min_{\underline{\omega}\in C_i} \Vert \underline{v} - \underline{\omega} \Vert_2^2$ denotes the projection onto the convex set $C_i$.
Note that the distances $d_{C_1}$ and $d_{C_2}$ provide the possibility to relax the hyperplane constraints $C_1$ and $C_2$:
The choice $\eta_1=\eta_2= +\infty$ imposes \eqref{eq:cons} as hard constraints (i.e., the intermediary quantities $\sum_{j=j_1}^{j_2} w(j,\kx) \log_2 X(j,\kx)$ are unbiased estimates of ${h}(\kx)$), while for  $\eta_1,\eta_2 < +\infty$, a violation of the constraints  \eqref{eq:cons} is possible but penalized. This adds a degree of freedom as compared to the standard estimation procedure \eqref{equ-estimh} and the solution $\widehat{\underline{\underline{h}}}^{\textrm{TV}}$ in \eqref{eq:tv}. Furthermore, note that the joint estimation of $\underline{\underline{h}}$ and $\underline{\underline{w}}$ enables the use of spatially varying weights $\underline{\underline{w}}$, which is otherwise impractical for a priori fixed weights (here, the weights are tuned automatically).

\noindent\textbf{Proposed algorithm.} \quad The minimization problem \eqref{eq:problem} is convex but non-smooth.
In the recent literature dedicated to non-smooth convex optimization, several efficient algorithms have been proposed, see, e.g., \cite{Bauschke_H_2011_book_con_amo, Combettes_P_2010_inbook_proximal_smsp,Parikh_N_2014_j-fto_proximal_a}.
Due to the gradient Lipschitz data fidelity term and the presence of several regularization terms (TV regularization, distances to convex sets), one suited algorithm is given in \cite{Vu_B_2011_j-acm_spl_adm, Condat_L_2012, Combettes_P_2014_p-icip_forward-bpd, Komodakis_N_2015}, referred to as FBPD (for Forward-Backward Primal-Dual).
This algorithm is tailored here to the problem~\eqref{eq:problem}, ensuring
convergence of a sequence $({\underline{\underline{h}}}^{[\ell]}, {\underline{\underline{w}}}^{[\ell]} )_{\ell \in \NN}$ to a solution of \eqref{eq:problem}.
The corresponding iterations are given in Algorithm~\ref{algo:cpcv}.
{{\begin{algorithm}
\small\begin{algorithmic}
\STATE  Initialization\\
$\left \lfloor \begin{array}{l}
\mbox{Set} \;\;\sigma >0\, \mbox{and}\, \tau \in \Big]0,\frac{1}{1+\Vert \sum_{j=j_1}^{j_2} \underline{X}(j,\cdot)\Vert^2+ 3\sigma}\Big[. \\
\mbox{Set} \;\;  \underline{\underline{h}}^{[0]}\in \RR^{N_1\times N_2}, \underline{\underline{w}}^{[0]}\in \RR^{J\times N_1\times N_2}, \, \\
\,\underline{\underline{u}}^{[0]} \in \RR^{J\times N_1\times N_2},\, \underline{\underline{y}}_1^{[0]}\in \RR^{ N_1 \times N_2} \,\mbox{and} \,\underline{\underline{y}}_2^{[0]}\in \RR^{ N_1 \times N_2}.
                \end{array} \right.$
\STATE
\STATE For $\ell = 0,1,\ldots$\\
$\left \lfloor \begin{array}{l}
\mbox{ \underline{\textit{Gradient descents steps}}  }\\
\underline{\underline{h}}^{[\ell+1]} =  \underline{\underline{h}}^{[\ell]}  - \tau\big( D_1^* \underline{\underline{y}}_1^{[\ell]} + D_2^* \underline{\underline{y}}_2^{[\ell]}\big) \\
\quad\quad\quad\quad\quad\quad\quad -2\tau\big({{h}}^{[\ell]}(\kx)-  \sum_j{{w}}^{[\ell]}(j,\kx)  \log_2{{X}}(j,\kx)\big)_{\kx \in \mathcal{K}}\\
\mbox{For}\, j=j_1,\ldots,j_2\\
\left \lfloor \begin{array}{l}
  \widetilde{\underline{\underline{w}}}^{[\ell]}(j,\cdot) = {\underline{\underline{w}}}^{[\ell]}(j,\cdot) - \tau \underline{\underline{u}}^{[\ell]}(j,\cdot) \\
  \hfill- 2\tau \Big(\big(\sum_j{{{w}}}^{[\ell]}(j,\kx)  \log_2{{{X}}}(j,\kx) \!-  \!{h}^{[\ell]}(\kx)\big)  \! \log_2\!\!{{{X}}}(j,\kx)  \Big)_{\kx\in\mathcal{K}} \\
\end{array} \right.\\\\
\mbox{\underline{\textit{Proximity operator step based on $d_{C_1}$}} }\\
\underline{{\underline{w}}}^{[\ell+1]}= \big(\prox_{\tau\eta_1 d_{C_1}} (\widetilde{\underline{{w}}}(\cdot,\kx))\big)_{\kx\in \mathcal{K}}\\\\
\mbox{\underline{\textit{Proximity operator step for TV on $h$} } }\\
\widetilde{\underline{\underline{y}}}_1^{[\ell]} = \underline{\underline{y}}_1^{[\ell]} + \sigma D_1 (2\underline{\underline{h}}^{[\ell+1]} - \underline{\underline{h}}^{[\ell]} )\\
\widetilde{\underline{\underline{y}}}_2^{[\ell]} = \underline{\underline{y}}_2^{[\ell]} + \sigma D_2 (2\underline{\underline{h}}^{[\ell+1]} - \underline{\underline{h}}^{[\ell]} )\\
(\underline{\underline{y}}_1^{[\ell+1]}\!\!\!,\;\underline{y}_2^{[\ell+1]}) = (\widetilde{\underline{\underline{y}}}_1^{[\ell]},\widetilde{\underline{\underline{y}}}_2^{[\ell]}) \\\quad \quad \quad \quad \quad \quad  - \sigma \Big(\prox_{(\lambda/\sigma)\Vert \cdot \Vert_{2,1}} \big(\sigma^{-1} (\widetilde{{{y}}}_1^{[\ell]}(\kx),\widetilde{{{y}}}_2^{[\ell]}(\kx))\big)\Big)_{\kx\in \mathcal{K}}\\\\
\mbox{\underline{\textit{Proximity operator step for $d_{C_2}$} }}\\
\underline{\underline{q}}^{[\ell]} = \underline{\underline{u}}^{[\ell]} + \sigma (2\underline{\underline{w}}^{[\ell+1]} - \underline{\underline{w}}^{[\ell]} )\\
\underline{\underline{u}}^{[\ell+1]}=\underline{\underline{q}}^{[\ell]} - \Big(\prox_{\frac{\eta_2d_{C_2}}{\sigma}} (\sigma^{-1}{\underline{q}}^{[\ell]}(\cdot,\kx))\Big)_{\kx\in \mathcal{K}}\\
\end{array} \right.$
\end{algorithmic}
\caption{
Algorithm for solving~\eqref{eq:problem}\label{algo:cpcv}
}
\end{algorithm}}}

The notation $D_1^*$ and $D_2^*$  used in Algo.~\ref{algo:cpcv} stands for the adjoint of $D_1$ and $D_2$, respectively, while the notation $\prox$ stands for the proximity operator.
The proximity operator associated to a convex, lower semi-continuous function $\varphi$ from $\HH$ (where $\HH$ denotes a real Hilbert space) to $\RX$ is defined at the point $u \in \HH$ as $\prox_\varphi(u) = \arg\min_{v\in \HH} \frac{1}{2} \Vert u - v \Vert^2 +\varphi(v)$.
The proximity operators involved in Algo.~\ref{algo:cpcv} have a closed-form expression (cf. \cite{Peyre_G_2011_p-eusipco_gro_sop}):
\begin{equation}
(\forall\;\underline{u}  \in \RR^2) \quad \prox_{\frac{\lambda}{\sigma}\Vert\cdot \Vert_{2,1}} (\underline{u}) =  \max(0,1 - \frac{\lambda}{\sigma \Vert \underline{u}\Vert}_2) \underline{u}.
\end{equation}
Moreover, according to \cite[Proposition~2.8]{Combettes_PL_2008_j-ip_proximal_apdmfscvip}, if $C$ denotes a non-empty closed convex subset of $\RR^{J}$ and if $\eta>0$, 
\begin{equation}
\prox_{\eta d_C}( \underline{u}) =
\begin{cases}
\underline{u} + \frac{\eta (P_c (\underline{u}) - \underline{u})}{d_C(\underline{u})} \quad &\mbox{if} \quad d_C(\underline{u})>\eta\\
P_C({u}) &\mbox{if} \quad d_C(\underline{u})\leq\eta
\end{cases}
\end{equation}
for every $\underline{u}\in \RR^{J}$.
For our purpose, $C$ models the hyperplane constraints $C_1$ and $C_2$ and $P_{C_1}$ and $P_{C_2}$ have a closed-form expression given in \cite{Theodoridis_S_2011_j-ieee-spm_adaptive_lwp}, consequently $\prox_{\eta d_{C_1}}$ and  $\prox_{\eta d_{C_2}}$ also have a closed-form expression. Note that the projection onto the convex set $P_C$ is the proximity operator of the indicator function $\iota_C$ of a non-empty closed convex set $C\subset \HH$ (i.e., $\iota_C(x)= 0$ if $x\in C$ and $+\infty$ otherwise).

An example of the result provided by the proposed procedure is displayed in Fig.~\ref{fig:illprinciple}~(f) for $\lambda = 16$ and $\eta_1 = \eta_2=1000$.
{It yields a piece-wise constant estimate that very well reflects the true regularity displayed in Fig.~\ref{fig:illprinciple}~(b). Notably, the obtained histogram is pronouncedly spiked and can easily be thresholded in order to determine a labeling for the $Q=2$ regions.}
For this example, the strategy clearly outperforms the more classical TV procedure of Section \ref{ss:tvden}.

\subsection{Direct estimation of local regularity labels}
\label{sec-prox}
The common feature of the approaches of Sections \ref{ss:tvden} and \ref{ss:tvw} is that they aim at first providing denoised (piece-wise constant) estimates of $\underline{\underline{h}}$. The labeling of regions with constant pointwise regularity is then performed a posteriori by thresholding of the global histogram.
In this section, we propose a TV-based algorithm that addresses the partitioning problem directly from the estimates $\widehat{\underline{\underline{h}}}$ obtained  using \eqref{equ-estimh} with a priori fixed weights $\underline{\underline{w}}$ {and yields estimates of the areas with constant regularity without recourse to intermediary denoising and histogram thresholding steps}.

\noindent\textbf{Partitioning problem.\quad}
Formally, the problem consists in identifying the areas $(\Omega_q)_{1\leq q \leq Q}$ of a domain $\Omega$ that are associated with different values $(\mu_q)_{1\leq q \leq Q}$ of $\underline{\underline{h}}$,
\[(\forall q\in \{1,\ldots,Q\})(\forall \xx_0\in \Omega_q)  \qquad h(\xx_0) \equiv \mu_q\]
where $\bigcup_{q=1}^Q \Omega_q = \Omega$, and  $(\forall q\neq p),\;\Omega_q\cap \Omega_p= \emptyset$ (by convention, $\mu_q\leq \mu_{q+1}$).
Most methods for solving the partitioning problem are either based on the resolution of a non-convex criterion or require specific initialization \cite{Kass_M_1988_j-ijcv_snakes_acm,Mumford_D_1989_j-comm-pure-appl-math_optimal_aps, caselles1997geodesic, Boykov_Y_2001_p-iccv_interactive_gco}.
Here, we adopt the minimal partitions technique proposed in \cite{Chambolle_A_j-siam_is_con_amp}, which is based on a convex relaxation of the Potts model %
and consequently enables convergence to a global minimizer.
According to \cite{Cremers_D_2011_inbook_convex_rts}, our partitioning problem can be written as
\begin{equation}
\label{eq:potts}
\underset{\Omega_1,\ldots,\Omega_Q}{\text{minimize}} \sum_{q=1}^Q\int_{\Omega_q} \ell_q(\widehat{h}(\xx)) d\underline{x} + \chi \sum_{q=1}^Q \mbox{Per}(\Omega_q)   \quad  \mbox{subj. to}\qquad \begin{cases}\bigcup_{q=1}^Q \Omega_q = \Omega,\\ (\forall q\neq p),\;\Omega_q\cap \Omega_p= \emptyset  \end{cases}
\end{equation}
where $\mbox{Per}(\Omega_q)$ measures the perimeter of region $\Omega_q$, $\ell_q(\widehat{h}(\xx))$ denotes the negative log-likelihood of the estimated local regularity  associated with region $\Omega_q$, and the constraints ensure a non-overlapping partition of $(\Omega_q)_{1\leq q \leq Q}$. The parameter $\chi>0$ models the roughness of the solution.

\noindent\bf Problem formulation.\quad}
Let  $\underline{\underline{\Omega}}_q\in\{0,1\}^{N_1\times N_2},\;1\leq q \leq Q$, denote a set of $Q$ partition matrices, i.e., $\sum_{q=1}^{Q+1}\underline{\underline{\Omega}}_q$ is an $N_1\times N_2$ matrix with all entries equal to one. The discrete analogue of \eqref{eq:potts} is the Potts model, which is known to be NP-hard to solve.
A convex relaxation, involving the total variation, is given by \cite{Cremers_D_2011_inbook_convex_rts,Pock_Y_2009_p-cvpr_convex_rac}
{\begin{multline}
\label{eq:crit2}
\underset{\theta_0, \ldots, \theta_{Q}}{\text{min}}\!\!\Bigg\{\sum_{q=1}^{Q}\sum_{\kx\in\mathcal{K}} (\theta_{q-1}(\kx) -\theta_{q}(\kx) )\ell_q\Big(\sum_{j=j_1}^{j_2} \!\!w(j,\kx) \log_2 X(j,\kx)\Big) + \lambda \sum_{q=1}^{Q-1} \text{TV}(\underline{\underline{\theta}}_q)\Bigg\} \\
 \mbox{subj. to}\quad (\forall \kx \in \mathcal{K}) \quad  1\equiv \theta_{0}(\kx)\geq \ldots \geq  \theta_{Q}(\kx) \equiv 0
\end{multline}
where the weights are a priori fixed,  are independent of $\kx$, $w(j,\kx)=w(j)$, and satisfy the constraints \eqref{eq:cons}.}
The regularization parameter $\lambda>0$ impacts the number of areas created for each single label. When $\lambda$ is small,  several unconnected areas can occur for a single label while the solution favors dense regions when $\lambda$ is large.
A bound on the error of the solution of the convex relaxation \eqref{eq:crit2} is provided in \cite{Cremers_D_2011_inbook_convex_rts} (for the special case of two classes the solution coincides with the global minimizer of the Ising problem \cite{Chan_T_2006_siam_algorithms_fgm}). It results that the solutions of the minimization problem~\eqref{eq:crit2}, denoted $\widehat{\underline{\underline{\theta}}}_q\in\{0,1\}^{N_1\times N_2}$, are binary matrices that encode the partition matrices $\underline{\underline{\Omega}}_q$ such that
\begin{equation}
\widehat{\theta}_{q-1}(\kx) - \widehat{\theta}_{q}(\kx) =
\begin{cases}
1  \quad &\mbox{if $\Omega_q(\kx)=1$},\\
0  &\mbox{otherwise}.
\end{cases}
\end{equation}

\noindent \textbf{Algorithmic solution.} \quad
The functions involved in \eqref{eq:crit2} are convex, lower semi-continuous and proper, but the total variation penalty and the hard constraints are not smooth.
The algorithm proposed in \cite{Cremers_D_2011_inbook_convex_rts} considers the use of PDEs. In \cite{Pock_Y_2009_p-cvpr_convex_rac}, it is based on a Arrow-Hurwicz type primal-dual algorithm but requires inner iterations and upper boundedness of the primal energy in order to improve convergence speed. Here we employ a proximal algorithm in order to  avoid  inner iterations.
Note that in~\cite{Hiltunen_S_2012_p-eusipco_comparison_tps}, a proximal solution was proposed for a related minimization problem in the context of disparity estimation.
For the same reasons as those discussed in Section \ref{ss:tvw}, we propose a solution based on the FBPD algorithm \cite{Combettes_P_2014_p-icip_forward-bpd}, specifically tailored to the problem \eqref{eq:crit2}. The iterations are detailed in Algo.~\ref{algo:seg} and involve the hyperplane constraints
\begin{equation*}
\widetilde{C}_q = \{(\underline{\underline{\theta}}_{q-1}, \underline{\underline{\theta}}_{q}) \,\vert\,  \theta_{q-1}(\kx) \geq\theta_q(\kx)  ,\;\kx\in \mathcal{K}\}
\end{equation*}
for every $q = 1,\ldots,Q$ with ${{\theta}}_{0}(\kx) \equiv 1$ and ${{\theta}}_{Q}(\kx) \equiv 0$, $\kx\in \mathcal{K}$.
The projections onto $\widetilde{C}_q$, denoted $P_{\widetilde{C}_q}(\,\cdot\,)$, have closed-form expressions given in \cite{Theodoridis_S_2011_j-ieee-spm_adaptive_lwp}.
\vspace{0.1cm}

{{\begin{algorithm}
\small\begin{algorithmic}
\STATE Initialization\\
$\left \lfloor \begin{array}{l}
\mbox{Set} \;\;\sigma >0\, \mbox{and}\, \tau \in \Big]0,\frac{1}{3\sigma}\Big[. \\
\mbox{For} \;q=1,\ldots,Q-1,\;  {\underline{\underline{\theta}}}_q^{[0]} \in \RR^{N_1\times N_2}. \\
\mbox{For} \;q=1,\ldots,Q-1,\;   \underline{\underline{y}}_q^{1,[0]},\,\underline{\underline{y}}_q^{2,[0]}, \, \underline{\underline{z}}_q^{[0]} \in \RR^{N_1\times N_2}.
                \end{array} \right.$
\STATE
\STATE For $\ell = 0,1,\ldots$\\
$\left \lfloor \begin{array}{l}
\mbox{\textit{\underline{Gradient descents steps}}}\\
\mbox{For}\, q=1,\ldots,Q-1\\
\left \lfloor \begin{array}{l}
\widetilde{\underline{\underline{\theta}}}_q^{[\ell]} =  P_{[0,1]^{N_1\times N_2}} \Big(\underline{\underline{\theta}}_q^{[\ell]} - \tau (\underline{\underline{\ell}}_{q+1} - \underline{\underline{\ell}}_q) - \tau \underline{\underline{z}}_q^{[\ell]}\\\quad\quad \quad \quad  \quad \quad \quad\quad\quad\quad \quad \quad \quad\quad - \tau\big(D_1^* \underline{\underline{y}}_q^{1,[\ell]} +D_2^* \underline{\underline{y}}_q^{2,[\ell]}\big) \Big)\\
\end{array} \right.\\
\mbox{For}\, q = 1,\ldots,Q \, \mbox{ odd}\, \\
\left \lfloor \begin{array}{l}
\big(\underline{\underline{\theta}}_{q-1}^{[\ell+1]},\underline{\underline{\theta}}_{q}^{[\ell+1]}\big)  = P_{\widetilde C_q} \big(\widetilde{\underline{\underline{\theta}}}_{q-1}^{[\ell]},\widetilde{\underline{\underline{\theta}}}_{q}^{[\ell]}\big)\\
\end{array} \right.\\
\\
\mbox{\textit{\underline{Proximity operator step based on TV and on even $\widetilde{C}_q$}} }\\
\mbox{For}\, q=1,\ldots,Q-1\\
\left \lfloor \begin{array}{l}
\widetilde{\underline{\underline{y}}}_q^{1,[\ell]} = \underline{\underline{y}}_q^{1,[\ell]} + \sigma D_1 (2\underline{\underline{\theta}}_q^{[\ell+1]} - \underline{\underline{\theta}}_q^{[\ell]} )\\
\widetilde{\underline{\underline{y}}}_q^{2,[\ell]} = \underline{\underline{y}}_q^{2,[\ell]} + \sigma D_2 (2\underline{\underline{\theta}}_q^{[\ell+1]} - \underline{\underline{\theta}}_q^{[\ell]} )\\
\widetilde{\underline{\underline{z}}}_q^{[\ell]} = \underline{\underline{z}}_q^{[\ell]} + \sigma (2\underline{\underline{\theta}}_q^{[\ell+1]} - \underline{\underline{\theta}}_q^{[\ell]} )\\
(\underline{\underline{y}}_q^{1,[\ell+1]},\underline{\underline{y}}_q^{2,[\ell+1]}) = (\widetilde{\underline{\underline{y}}}_q^{1,[\ell]},\widetilde{\underline{\underline{y}}}_q^{2,[\ell]})\\
\qquad\qquad\qquad\qquad - \sigma \prox_{\sigma^{-1}\lambda\Vert \cdot \Vert_{2,1}} \big(\sigma^{-1} (\widetilde{\underline{\underline{y}}}_q^{1,[\ell]},\widetilde{\underline{\underline{y}}}_q^{2,[\ell]})\big)\\
\end{array} \right.\\
\mbox{For}\, q = 1,\ldots,Q \, \mbox{ even}\\
\left \lfloor \begin{array}{l}
\big(\underline{\underline{z}}_{q-1}^{[\ell+1]},\underline{\underline{z}}_{q}^{[\ell+1]}\big) = \big(\widetilde{\underline{\underline{z}}}_{q-1}^{[\ell+1]},\widetilde{\underline{\underline{z}}}_{q}^{[\ell+1]}\big) - \sigma  P_{\widetilde C_q} \big(\frac{\widetilde{\underline{\underline{z}}}_{q-1}^{[\ell]}}{\sigma},\frac{\widetilde{\underline{\underline{z}}}_{q}^{[\ell]}}{\sigma}\big)\\
\end{array} \right.\\
\end{array} \right.$
\end{algorithmic}
\caption{Algorithm for solving~\eqref{eq:crit2}\label{algo:seg}}
\end{algorithm}}}

\newpage
\noindent{\bf Negative log-likelihood.\quad}
In the present work we focus on the Gaussian negative log-likelihood, i.e.,
{{\small{\[
\ell_q\Big(\!\sum_{j=j_1}^{j_2} \!\!w(j,\kx) \log_2 X(j,\kx)\!\Big) \!=\! \frac{\Big(\!\sum_{j=j_1}^{j_2} \!\!w(j,\kx) \log_2 X(j,\kx) - \mu_q\!\Big)^{\!2}}{2\sigma_q^2}
\]}}}
\!\!where $\mu_q$ and $\sigma^2_q$ denote the mean value and the variance {in the region $\underline{\underline{\Omega}}_q$}, respectively.

The a priori choice of $(\mu_q)_{1\leq q \leq Q}$ is likely to strongly impact the estimates $(\widehat{\underline{\underline{\theta}}}_q)_{1\leq q \leq Q}$. We therefore propose to alternate the estimation of $(\underline{\underline{\theta}}_q)_{1\leq q \leq Q-1}$ and $(\mu_q)_{1\leq q \leq Q}$. The values $(\mu_q)_{1\leq q \leq Q}$ are first initialized equi-distantly
 between the minimum and the maximum values of $\widehat{\underline{\underline{h}}}$. Then Algo.~\ref{algo:seg}  is run until convergence and  the values $(\mu_q)_{1\leq q \leq Q}$ are re-estimated on the estimated areas $(\widehat{\underline{\underline{\Omega}}}_q)_{1\leq q \leq Q}$. We iterate until stabilization of the estimates.
 The variances are fixed to $\sigma_q^2=\frac{1}{2}$ here.

\section{Segmentation performance assessment}
\label{sec-res}

Performance of the proposed procedures are qualitatively and quantitatively assessed using both synthetic scale-free textures  (using realizations of 2D multifractional Brownian fields, described in Section \ref{sec:proc}, with prescribed piecewise constant local regularity values) and real-world textures, chosen in reference databases.
Sample size is set to $N\times N = 512\times512$ (hence, $N_1=N_2=256$ due to the decimation operation in the DWT).
Analysis is conducted using a standard 2D--DWT with orthonormal tensor product Daubechies mother wavelets with $N_\psi = 2 $ vanishing moments and 4 decomposition levels, $(j_1, j_2) = (1,4)$.
The labeling solutions obtained with the algorithms based on local smoothing, TV denoising, joint estimation of regularity and weights and direct local regularity labels are referred to as $\widehat{\underline{\underline{\Omega}}}^{\textrm{S}}$, $\widehat{\underline{\underline{\Omega}}}^{\textrm{TV}}$,  $\widehat{\underline{\underline{\Omega}}}^{\textrm{TVW}}$ and  $\widehat{\underline{\underline{\Omega}}}^{\textrm{RMS}}$, respectively.
For the algorithms involving a histogram thresholding step,  
thresholds are automatically set at the local minima between peaks of (smoothed) histograms (and, when less local minima than desired labels are detected, at the position of the largest peak).

\begin{figure*}
\centering
\includegraphics[width=4.5cm, height = 3.5cm]{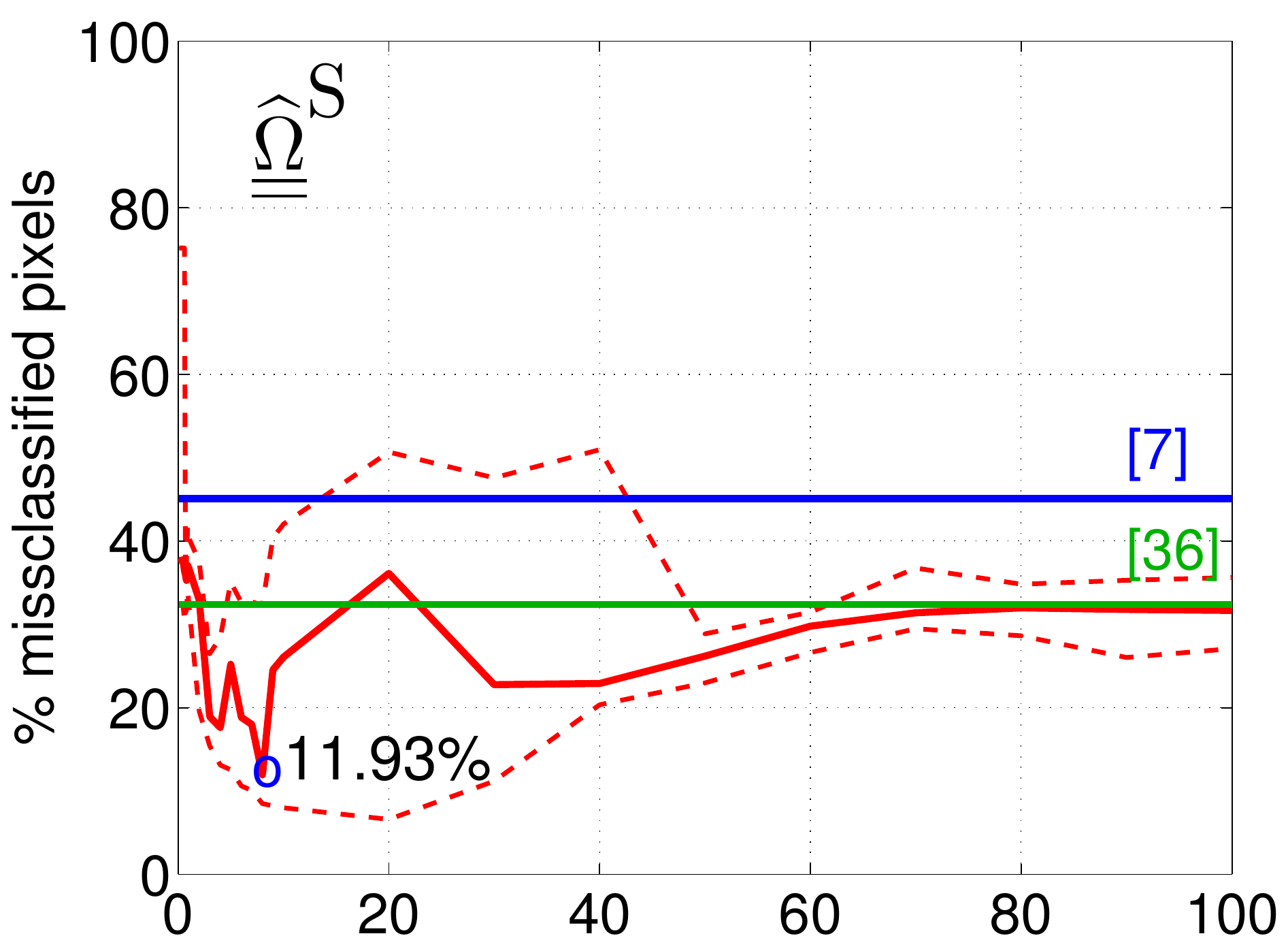}%
\includegraphics[width=4.5cm, height = 3.5cm]{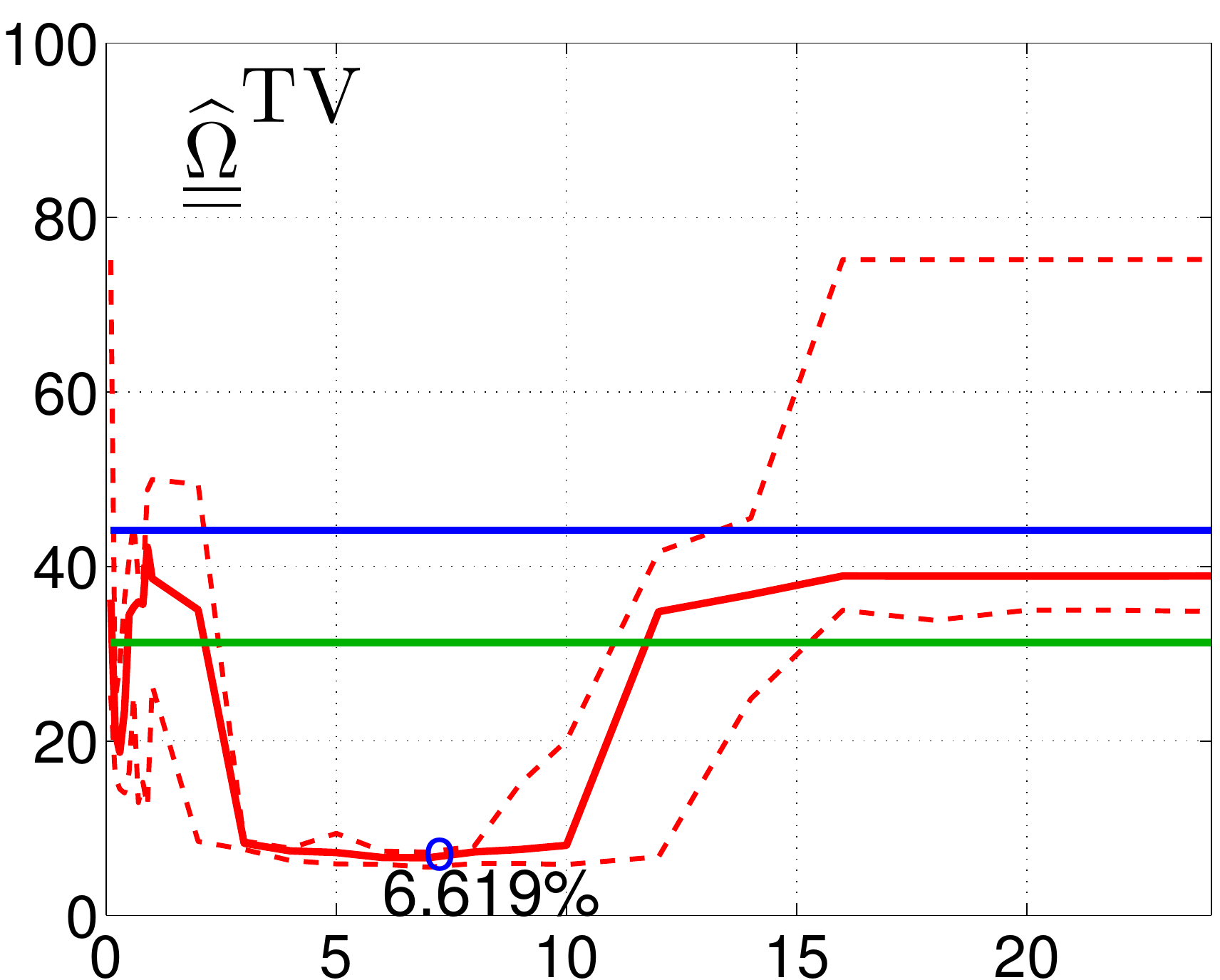}%
\includegraphics[width=4.5cm, height = 3.5cm]{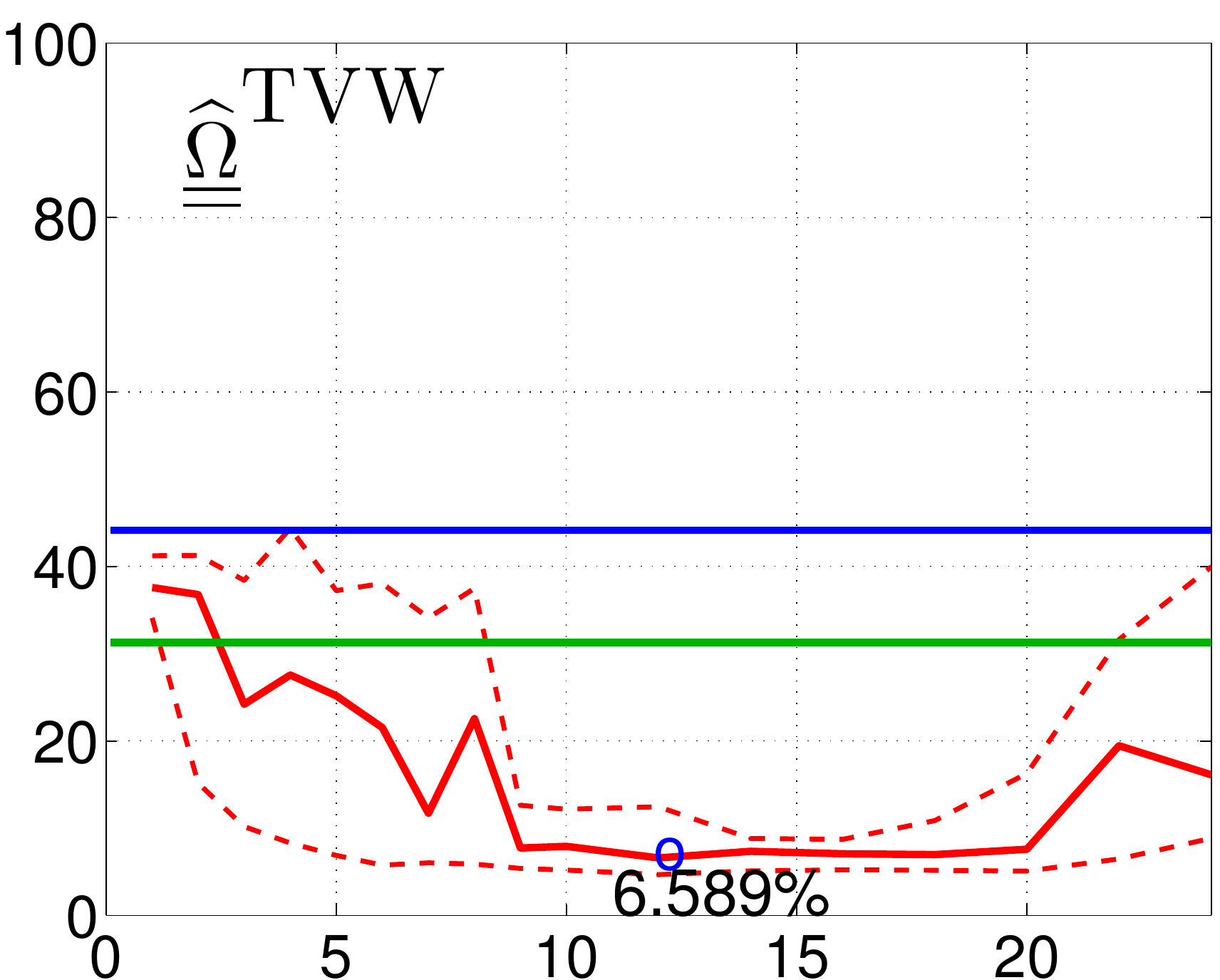}%
\includegraphics[width=4.5cm, height = 3.5cm]{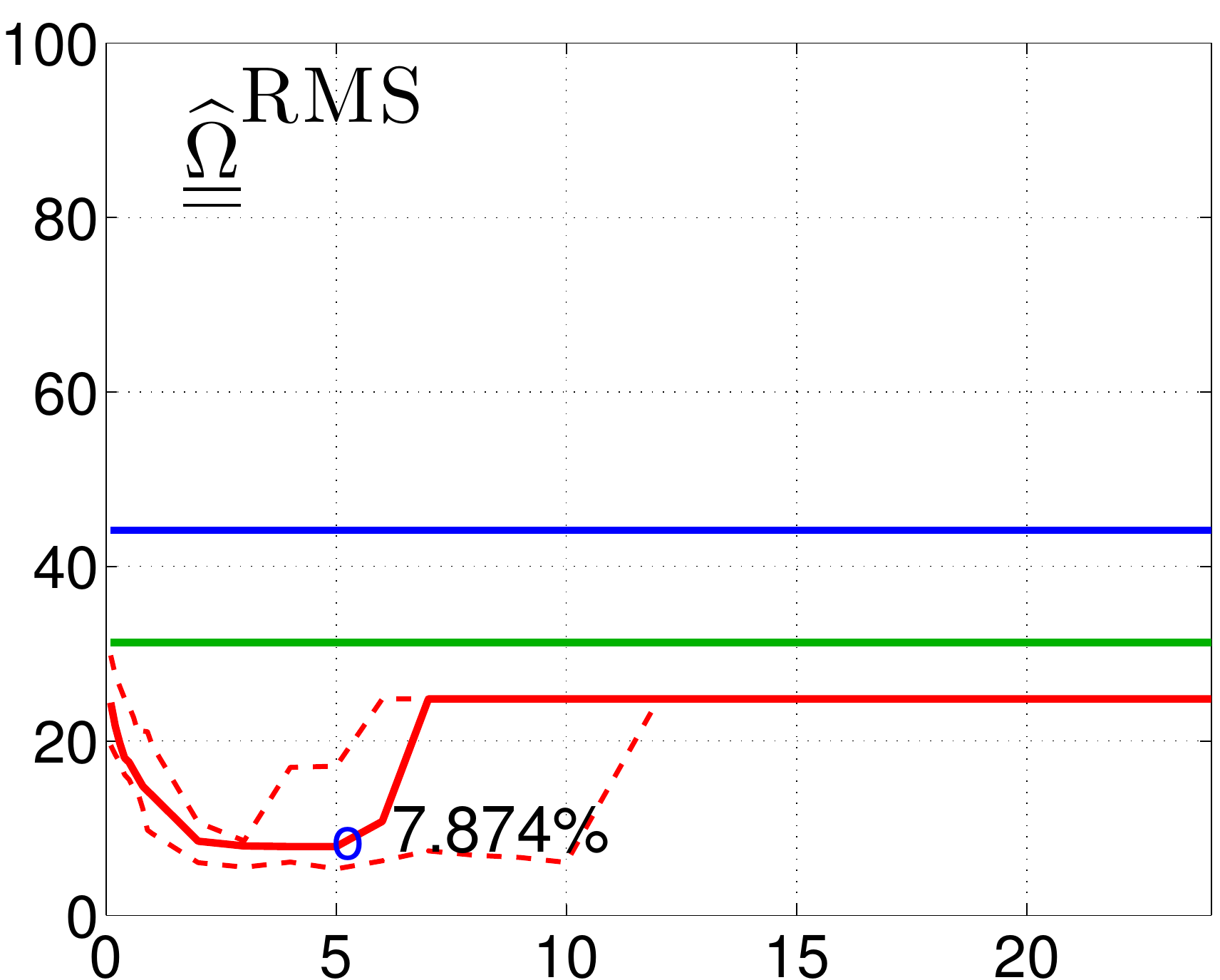} \vskip-1mm
(a) \small$h_1=0.5$, \;$h_2=0.7$  \vspace{-0.5mm}\vskip1mm
\includegraphics[width=4.5cm, height = 3.5cm]{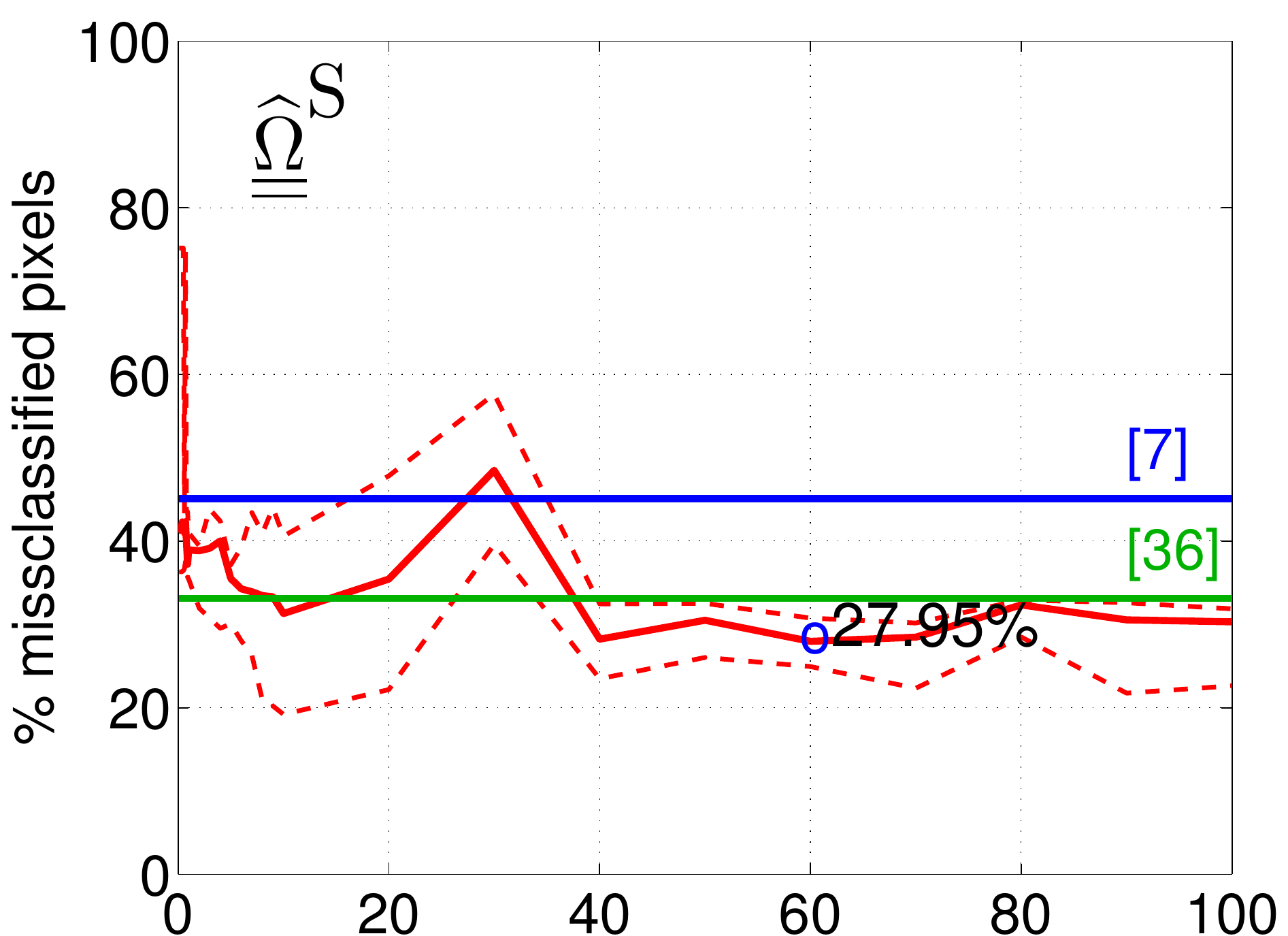}%
\includegraphics[width=4.5cm, height = 3.5cm]{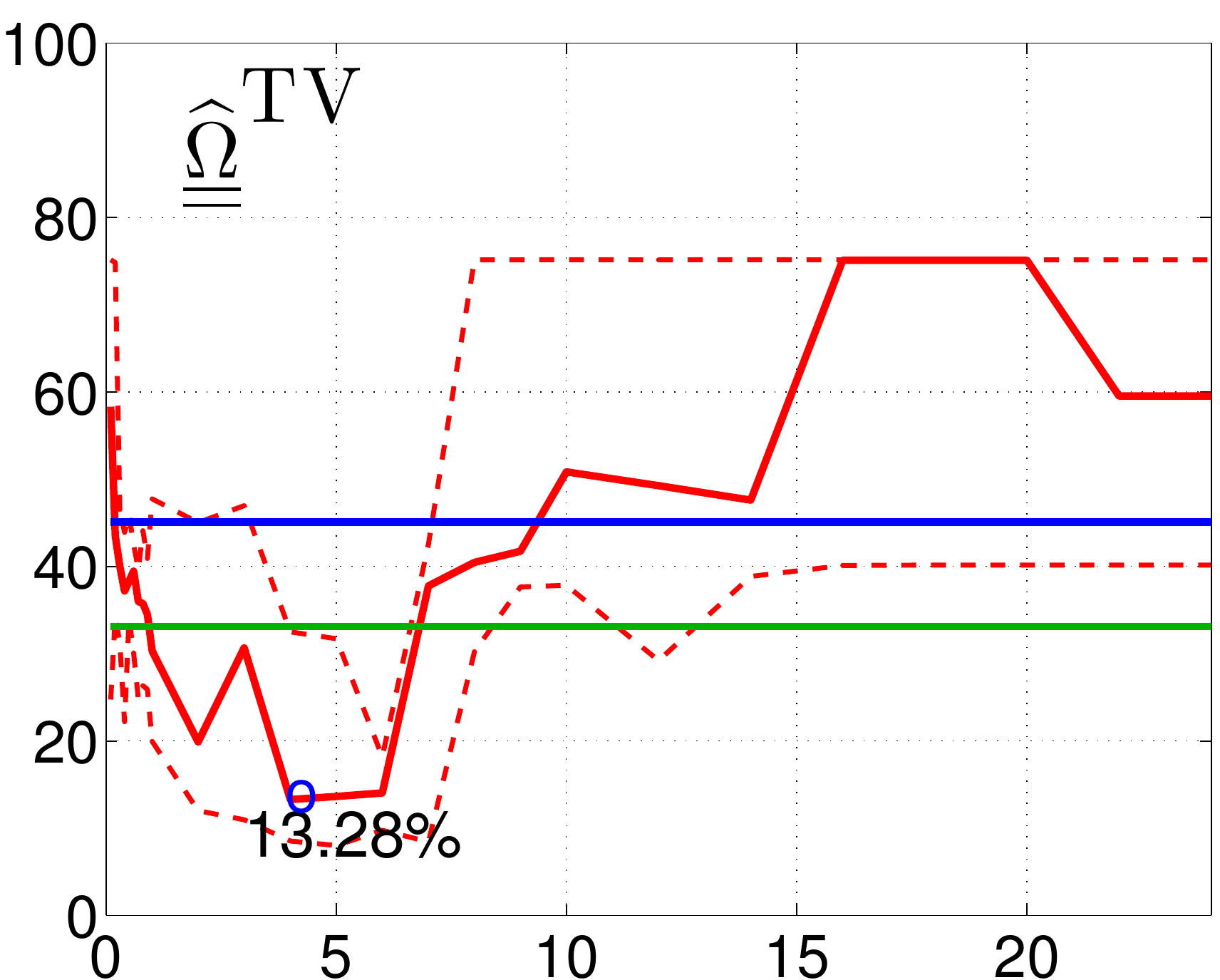}%
\includegraphics[width=4.5cm, height = 3.5cm]{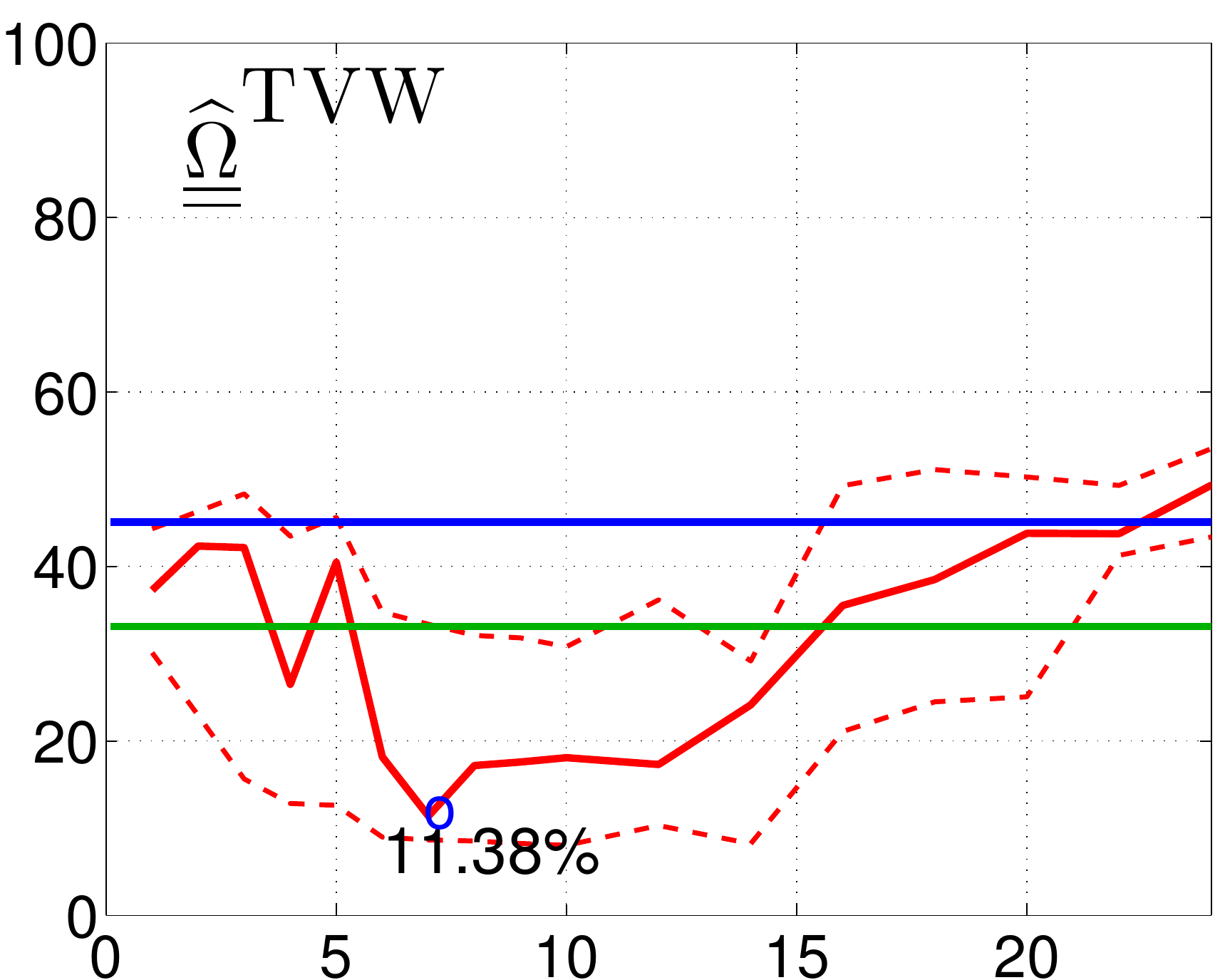}%
\includegraphics[width=4.5cm, height = 3.5cm]{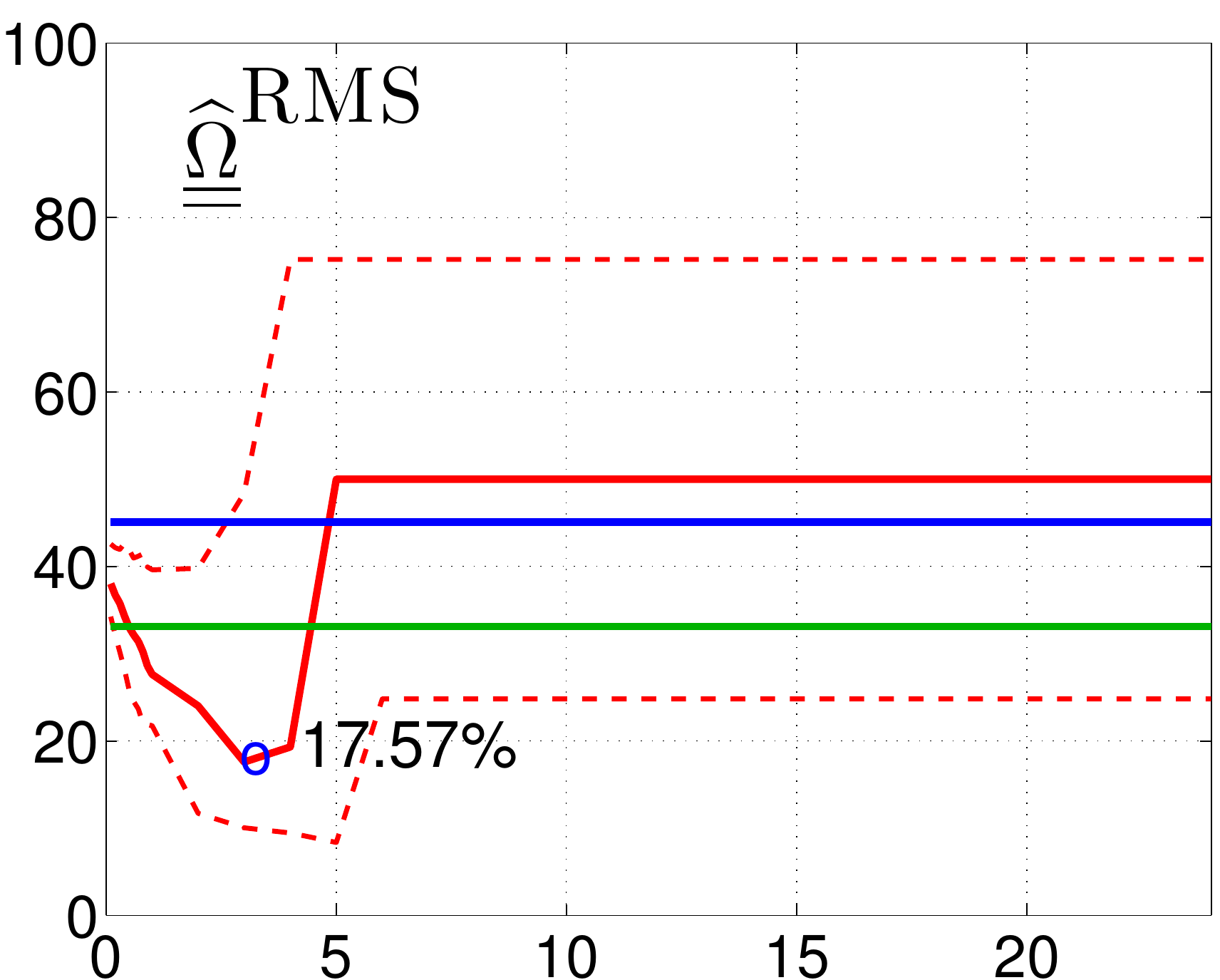}\vskip-1mm
(b) \small$h_1=0.6$, \;$h_2=0.7$
\caption{
Percentage of misclassified pixels as a function of penalty parameter $\lambda$ (respectively, {standard deviation $\sigma$} for $\widehat{\underline{\underline{\Omega}}}^{\textrm{S}}$): median (solid red) and upper and lower quartile (dashed red) for 10 realizations for  $\widehat{\underline{\underline{\Omega}}}^{\textrm{S}}$,  $\widehat{\underline{\underline{\Omega}}}^{\textrm{TV}}$,  $\widehat{\underline{\underline{\Omega}}}^{\textrm{TVW}}$ and  $\widehat{\underline{\underline{\Omega}}}^{\textrm{RMS}}$ (from left to right, respectively).
Subfigure (a): $(h_1,h_2) = (0.5,0.7)$; Subfigure (b):  $(h_1,h_2) = (0.6,0.7)$.
\label{fig:results_lambda}
}
\end{figure*}

\setlength{\tabcolsep}{0.2pt}
\begin{figure*}
\centering
\begin{tabular}{ccccccccc}
\begin{sideways}\parbox{28mm}{\!\!\!\!\footnotesize$\;\;h_1\!=\!0.5$,\; $h_2\!=\!0.7$}\end{sideways} &
\includegraphics[height =2.2cm]{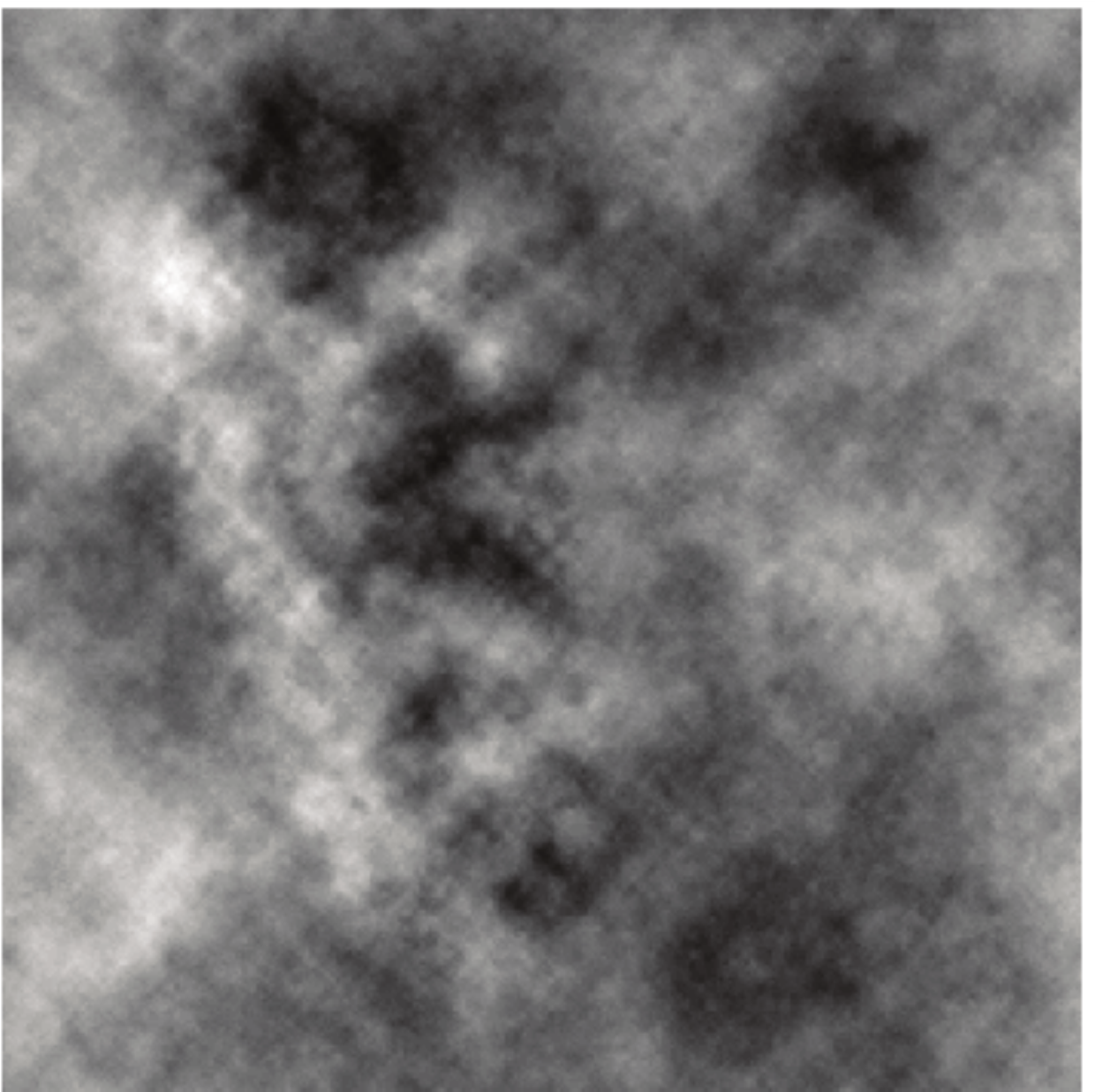}&
\includegraphics[height =2.2cm]{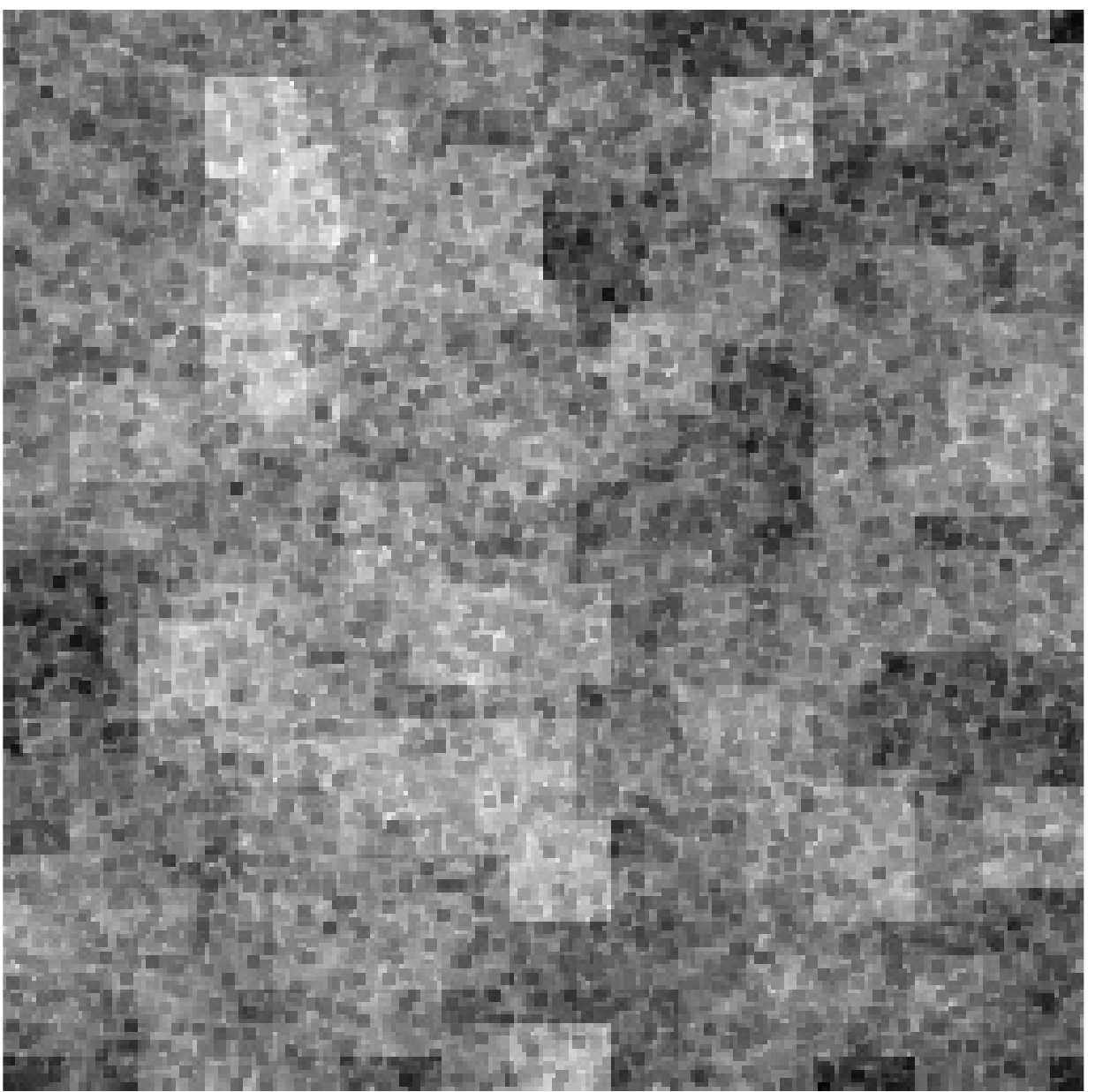}&
\includegraphics[height =2.2cm]{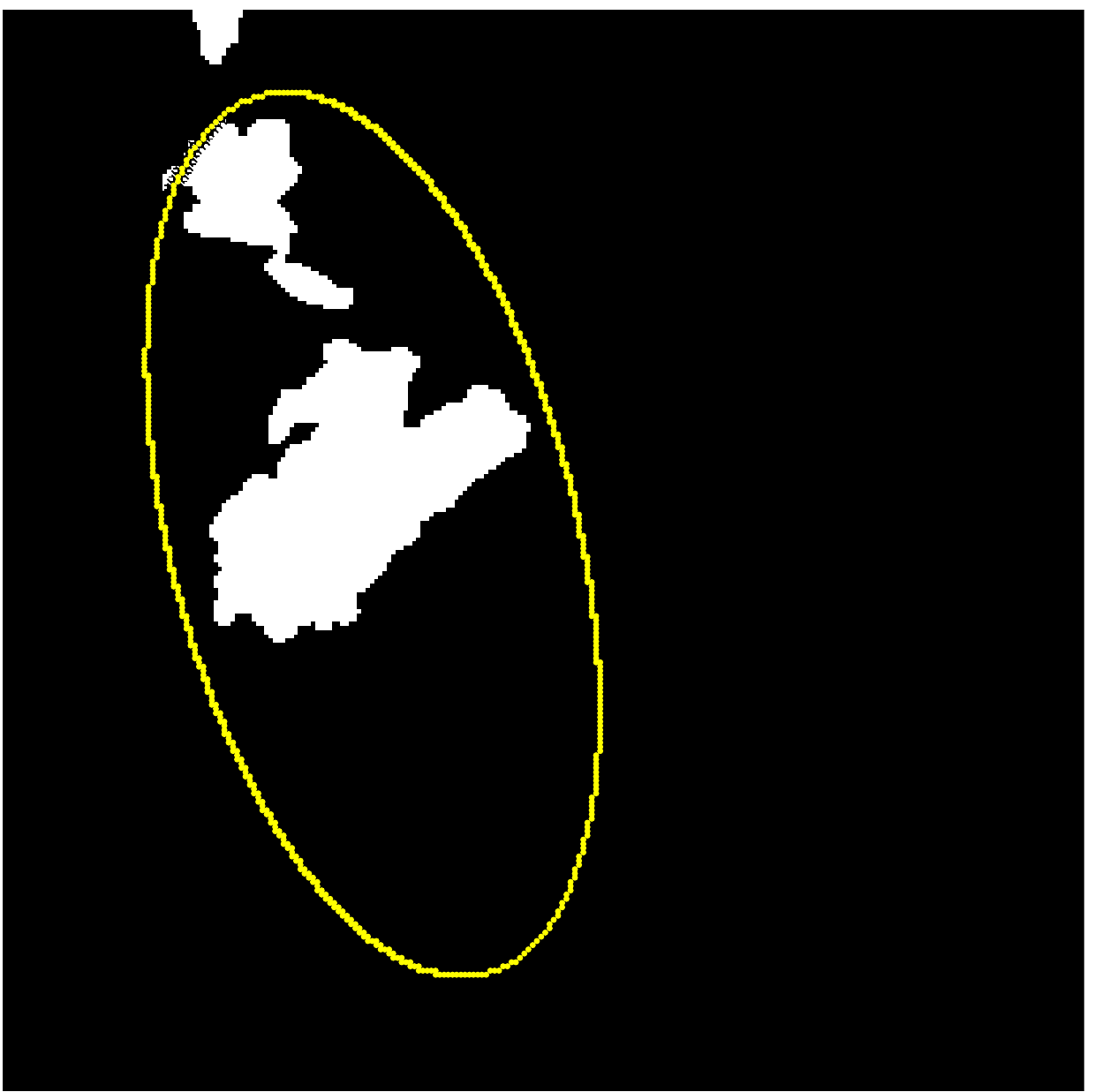}&
\includegraphics[height =2.2cm]{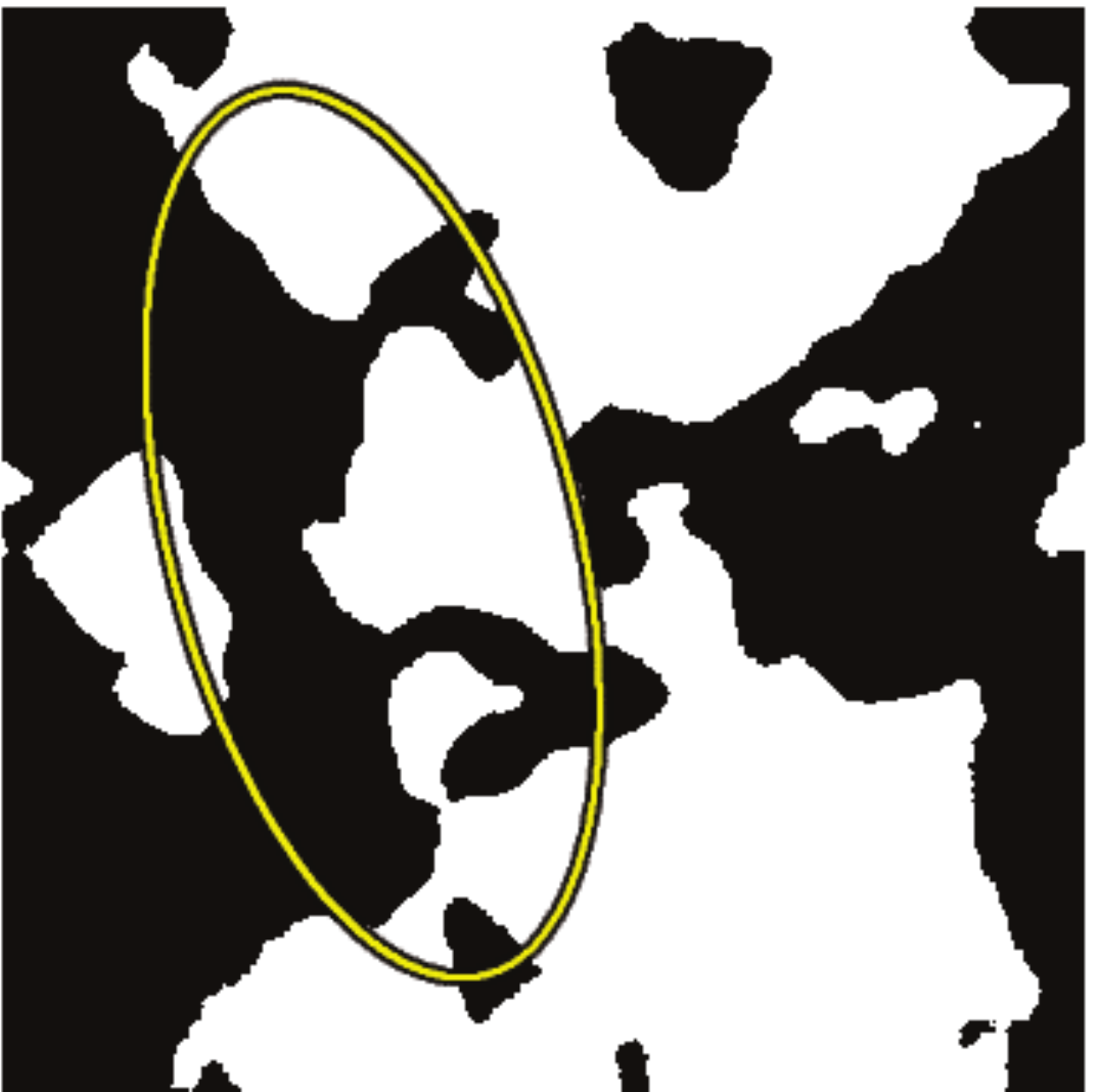}&
\includegraphics[height =2.2cm]{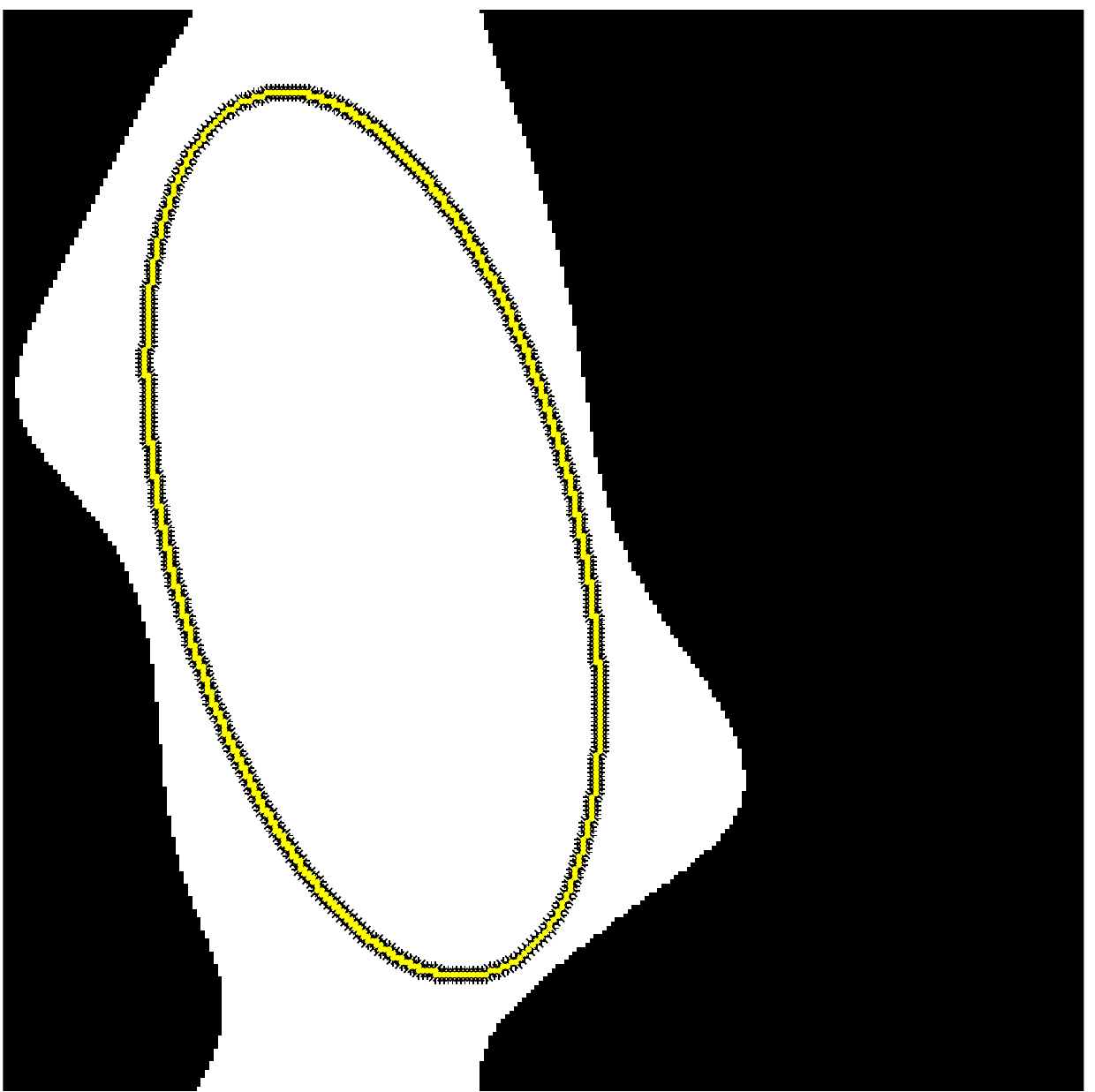} &
\includegraphics[height =2.2cm]{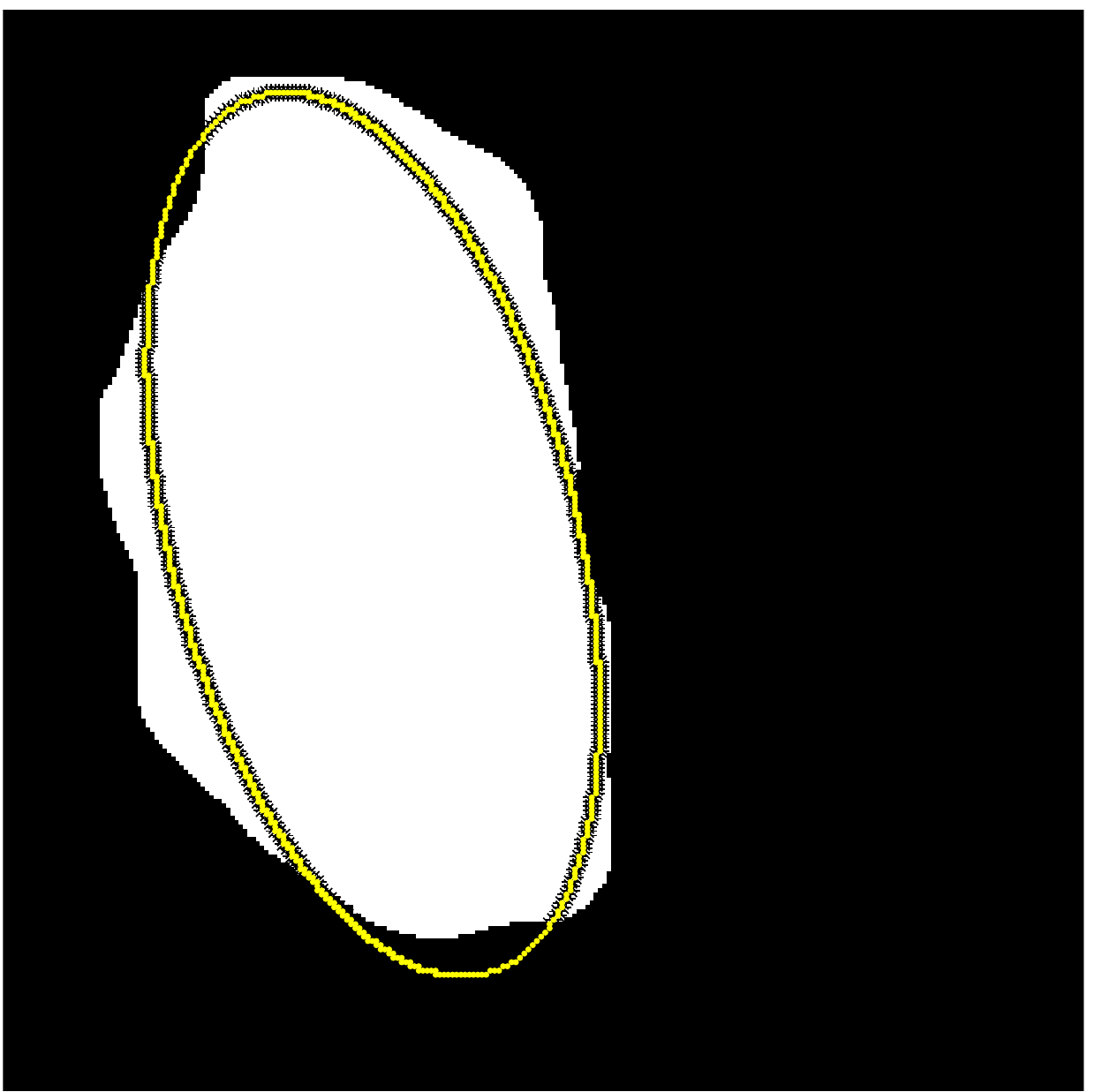}&
\includegraphics[height =2.2cm]{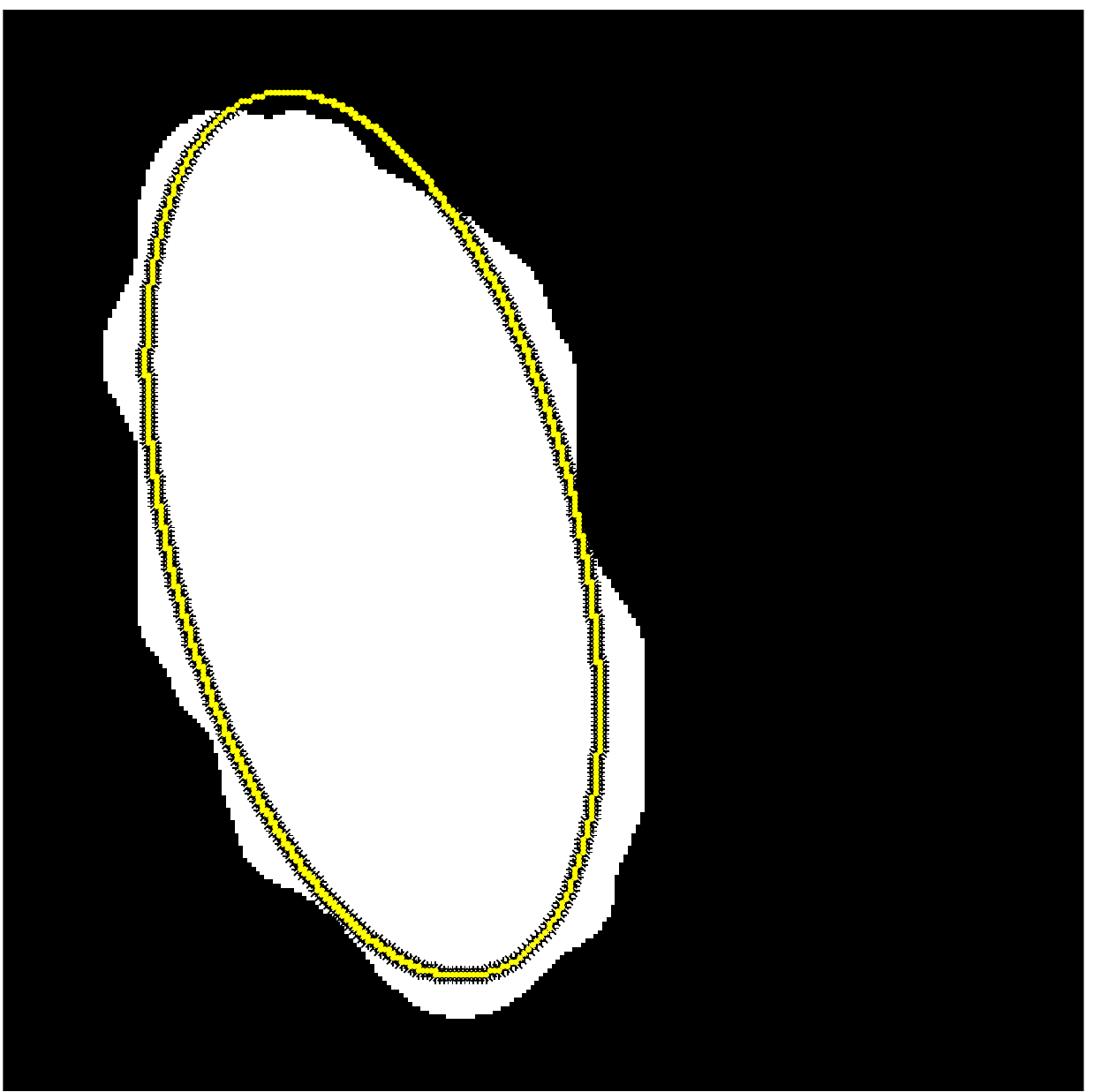} &
\includegraphics[height =2.2cm]{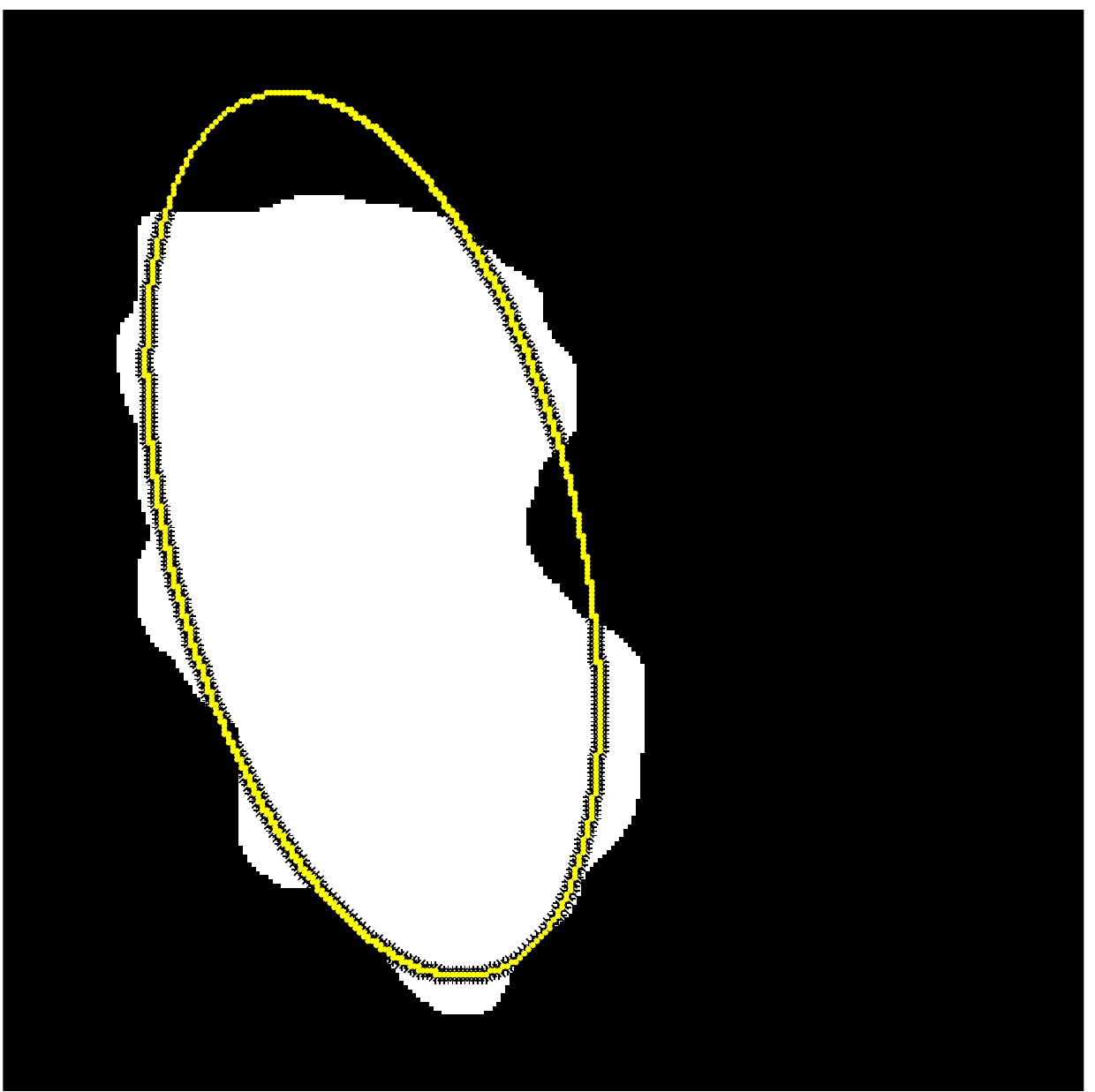}\vspace{-6mm}\\
&&&\ccol\hfill\small error 19.4\%\;&\ccol\hfill\small error 47.6\%\;&\ccol\hfill\small error 18.5\%\;&\ccol\hfill\small error 5.51\%\;&\ccol\hfill \small error 5.33\%\;&\ccol\hfill\small error 5.52\%\;\vspace{1mm}\\
\begin{sideways}\parbox{28mm}{\!\!\!\!\footnotesize$\;\;h_1\!=\!0.6$,\; $h_2\!=\!0.7$}\end{sideways} &
\includegraphics[height =2.2cm]{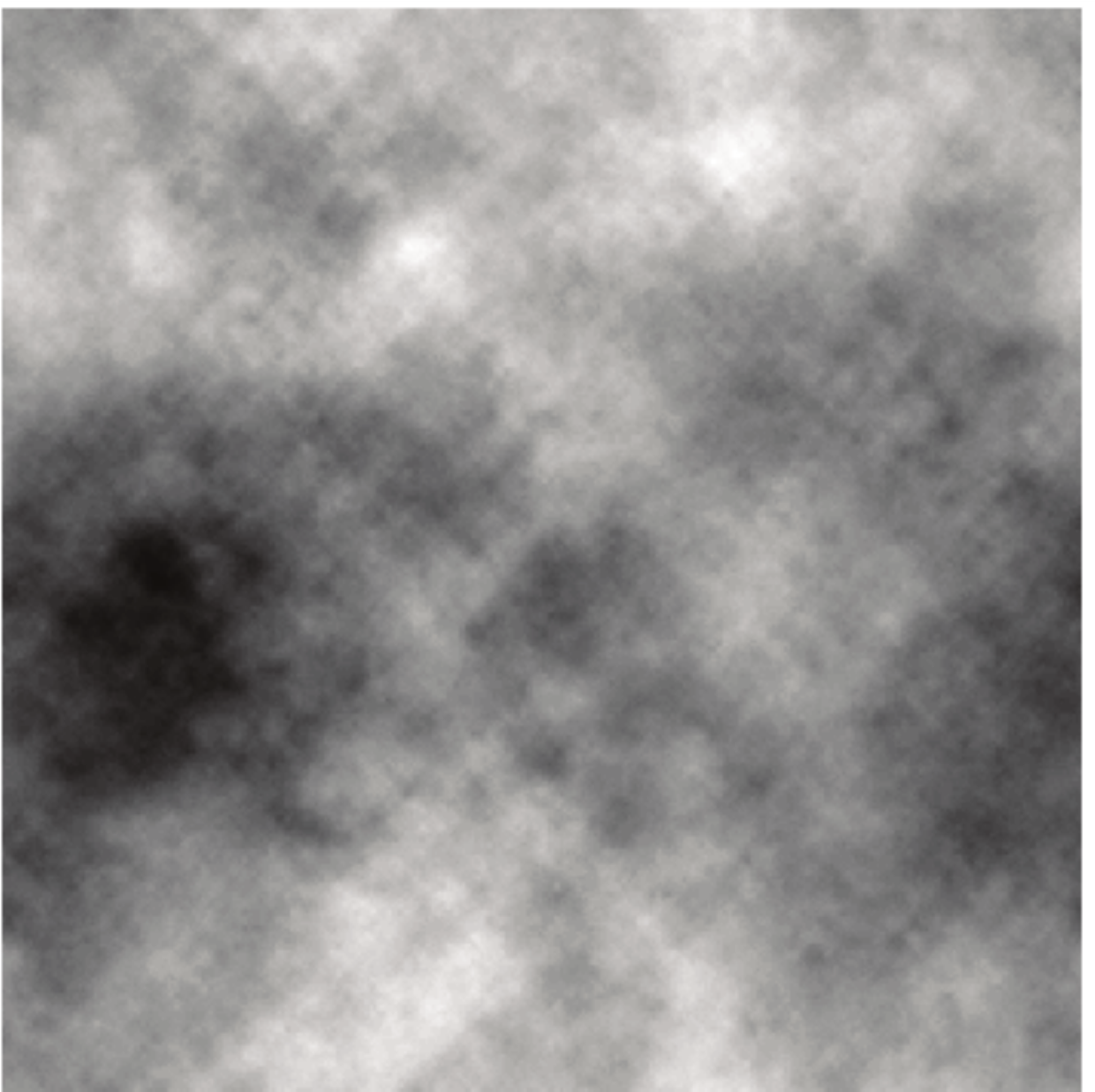}&
\includegraphics[height =2.2cm]{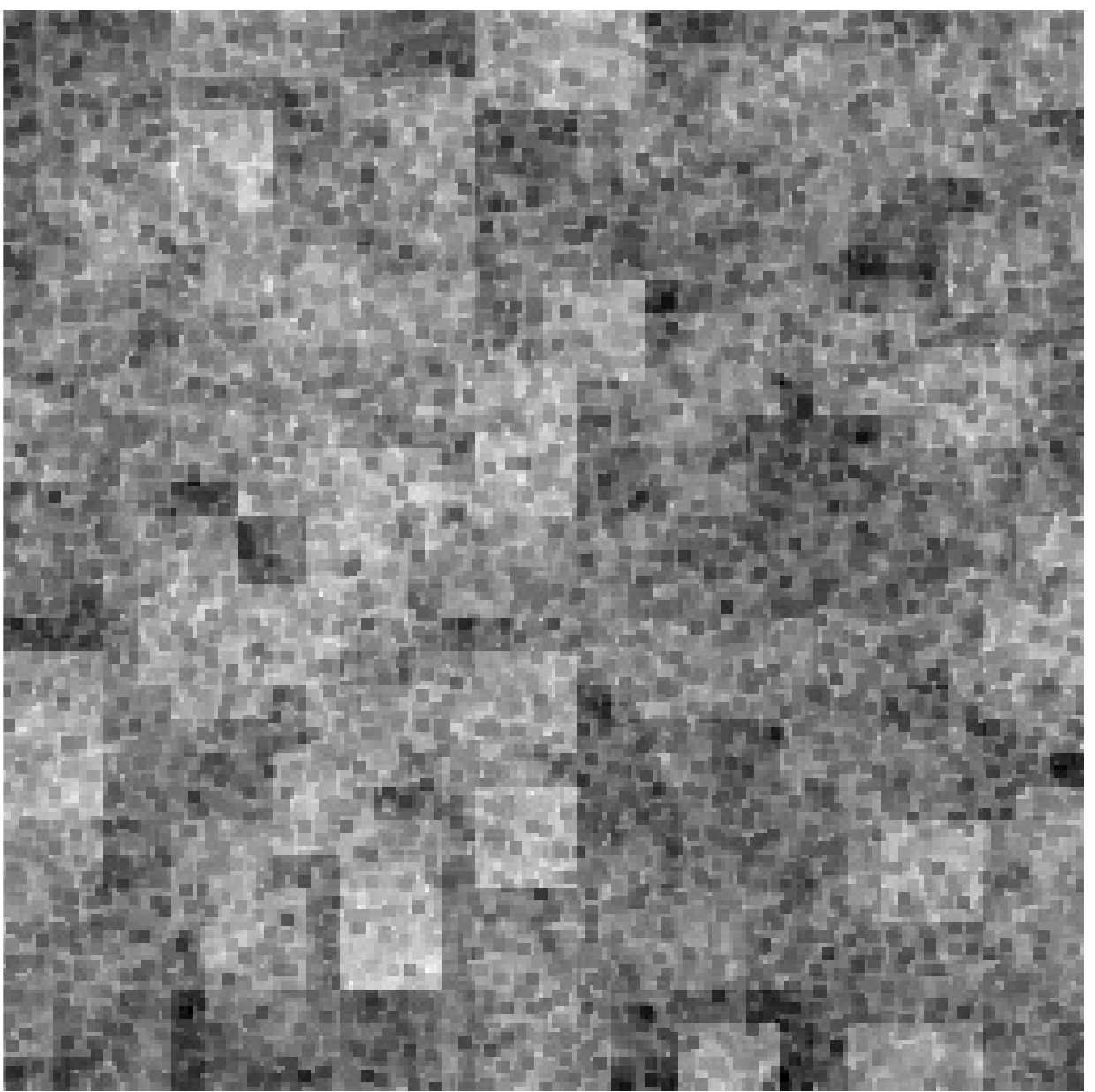}&
\includegraphics[height =2.2cm]{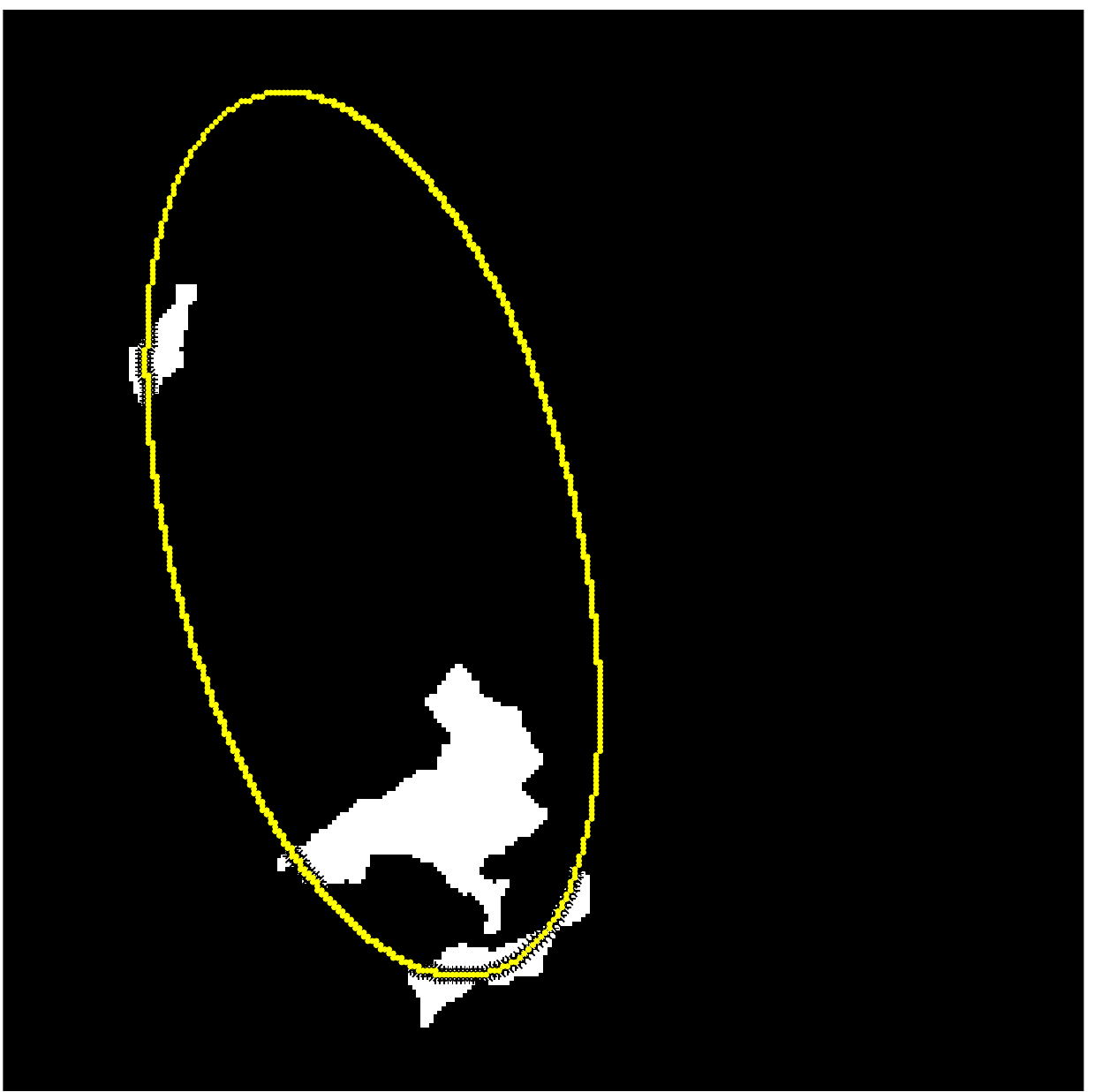}&
\includegraphics[height =2.2cm]{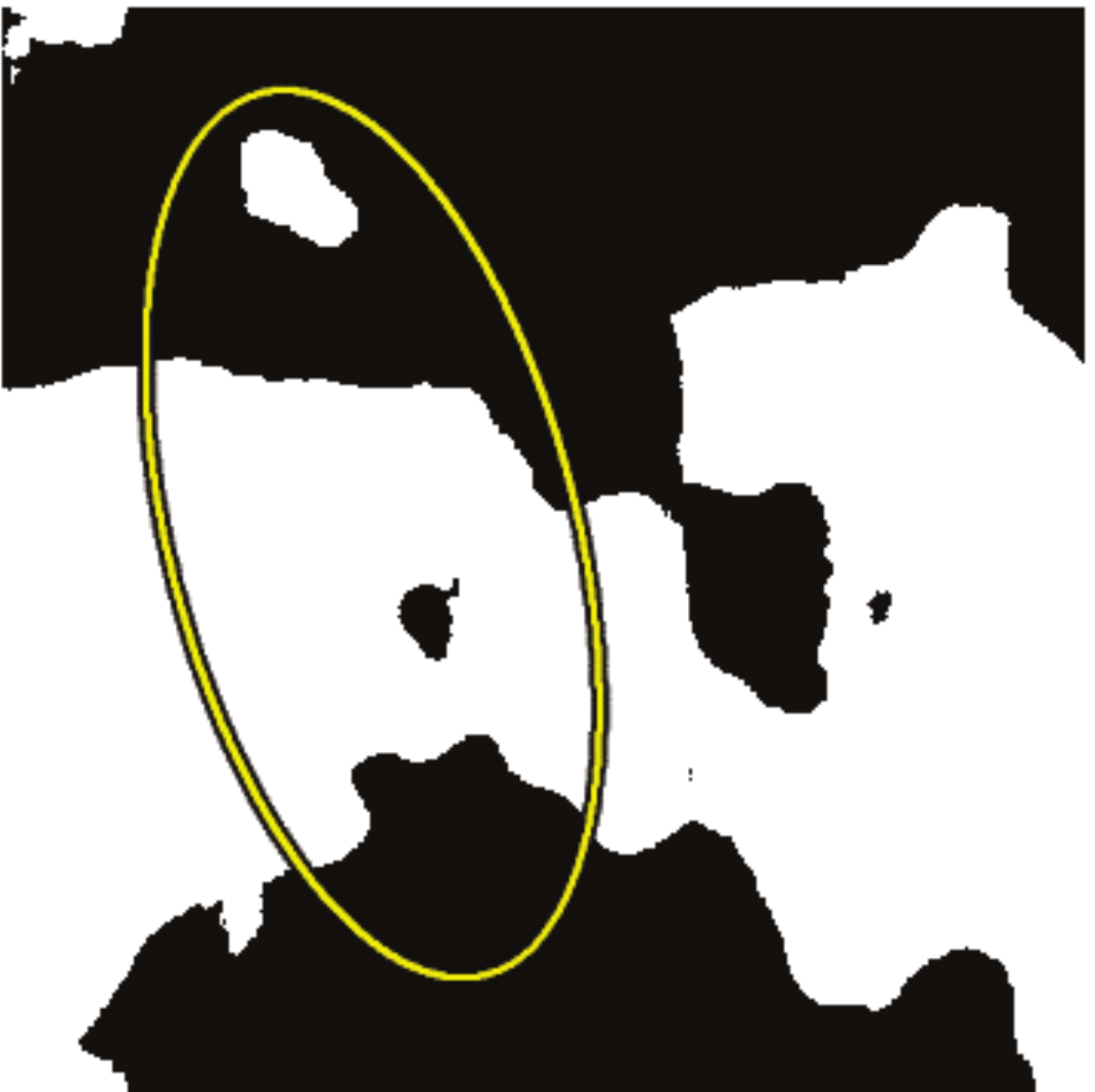}&
\includegraphics[height =2.2cm]{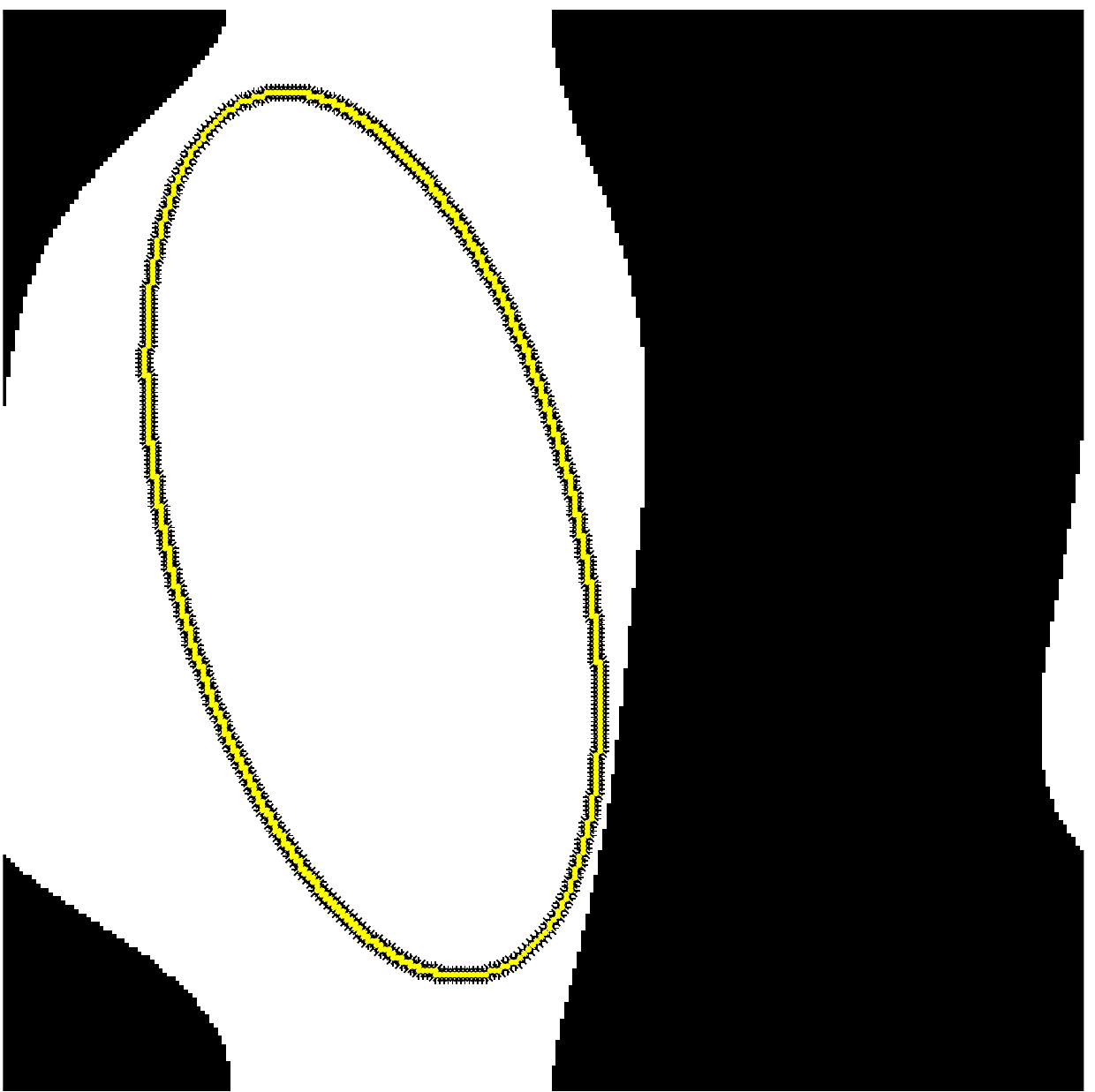} &
\includegraphics[height =2.2cm]{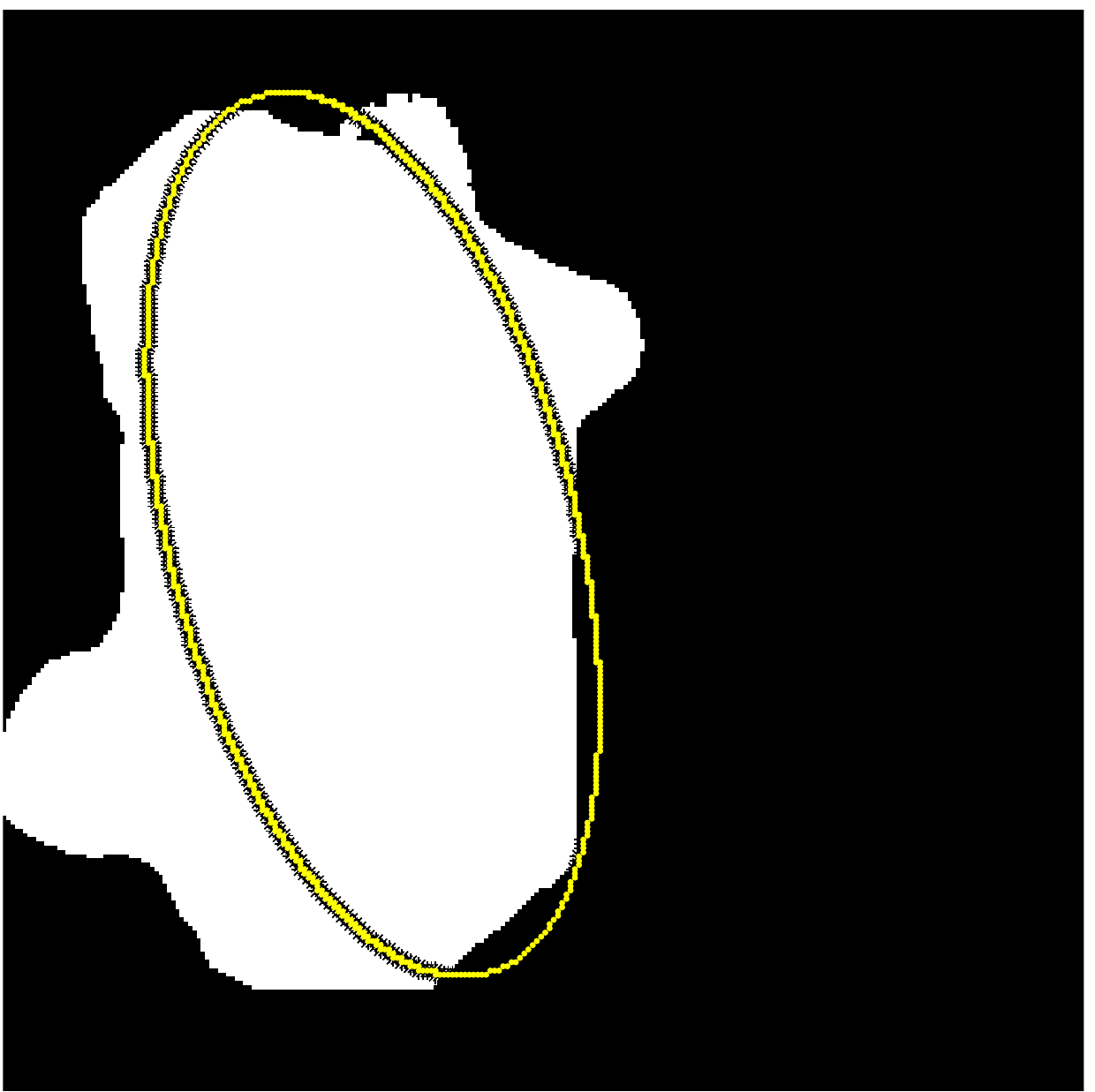}&
\includegraphics[height =2.2cm]{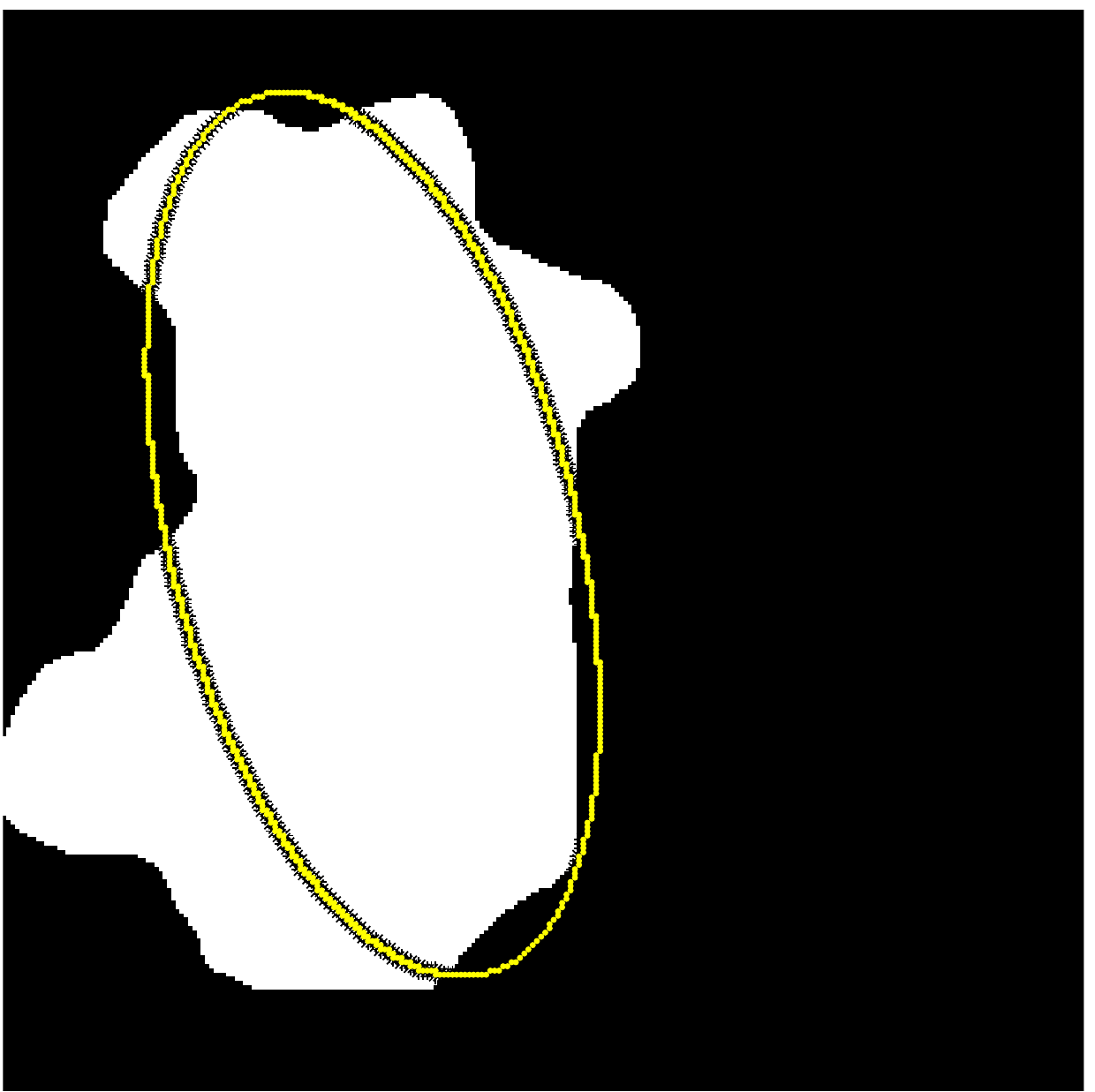} &
\includegraphics[height =2.2cm]{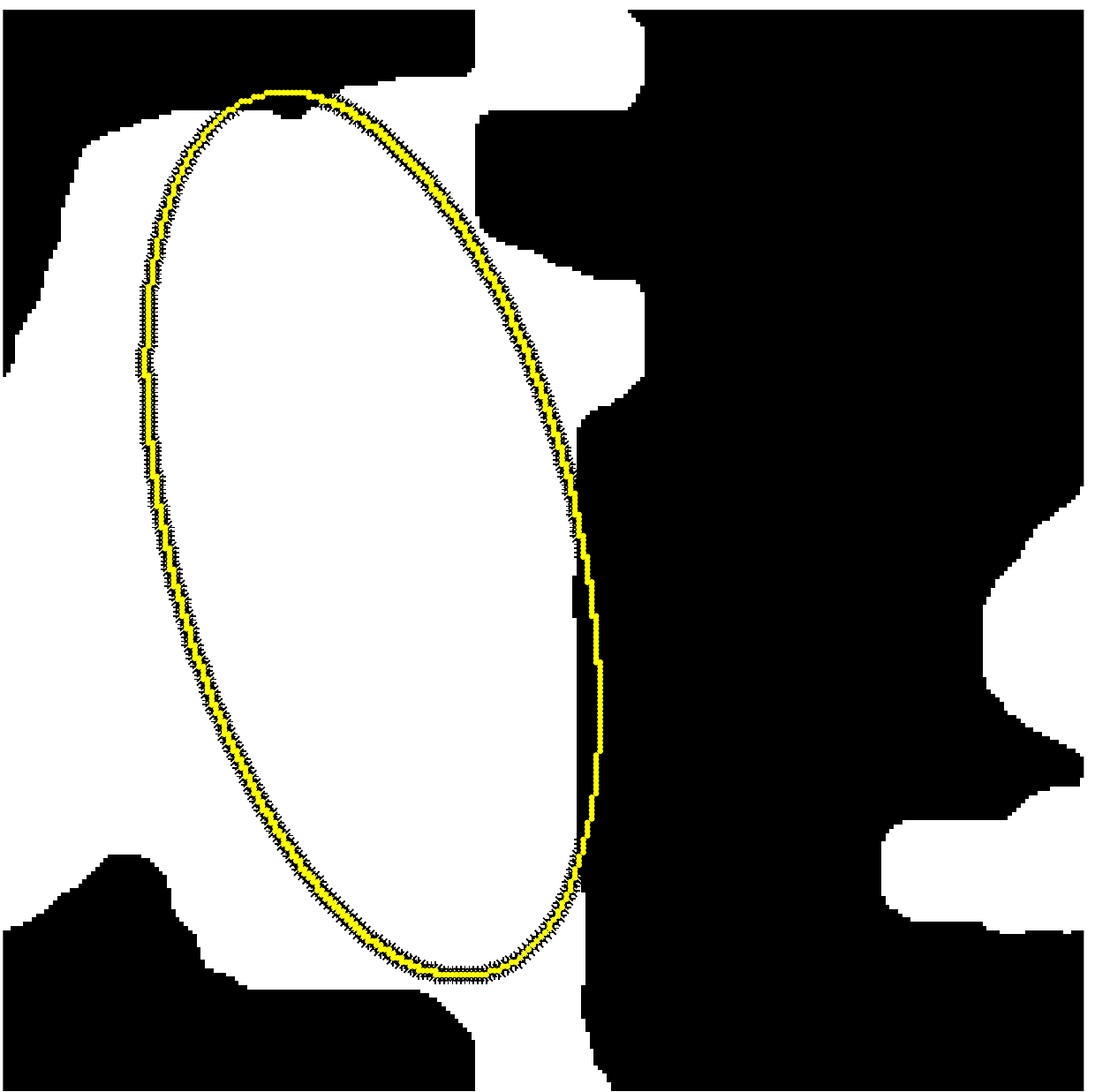}\vspace{-6mm}\\
&&&\ccol\hfill\small error 22.5\%\;&\ccol\hfill\small error 46.5\%\;&\ccol\hfill\small error 26.0\%\;&\ccol\hfill\small error 11.7\%\;&\ccol\hfill \small error 10.9\%\;&\ccol\hfill\small error 22.7\%\;\\
&(a) Data $\underline{\underline{f}}$ & (b) $\widehat{\underline{\underline{h}}}_{L}$& (c) \cite{Arbelaez_P_2011_j-ieee-tpami_con_dhis} & (d)\cite{Yuan_J_2015_j-ieee-tip_fac_bts} & (e)  $\widehat{\underline{\underline{\Omega}}}^{\textrm{S}}$ &(f)  $\widehat{\underline{\underline{\Omega}}}^{\textrm{TV}}$ & (i)  $\widehat{\underline{\underline{\Omega}}}^{\textrm{TVW}}$ & (j)  $\widehat{\underline{\underline{\Omega}}}^{\textrm{RMS}}$
\end{tabular}
\caption{Results obtained with the different proposed solutions compared to a basic smoothing of $\widehat{\underline{\underline{h}}}$ and the state-of-the-art texture segmentation approaches proposed in \cite{Arbelaez_P_2011_j-ieee-tpami_con_dhis} and \cite{Yuan_J_2015_j-ieee-tip_fac_bts}. Top row: $(h_1,h_2) = (0.5,0.7)$; bottom row:  $(h_1,h_2) = (0.6,0.7)$.
\label{fig:deltah5-7}}
\end{figure*}

\setlength{\tabcolsep}{0.2pt}
\begin{figure*}
\centering
\begin{tabular}{ccccccccc}
\begin{sideways}\parbox{30mm}{\footnotesize$\;\;h_i\in(0.2,0.4,0.7)$}\end{sideways}
&\includegraphics[height =2.2cm]{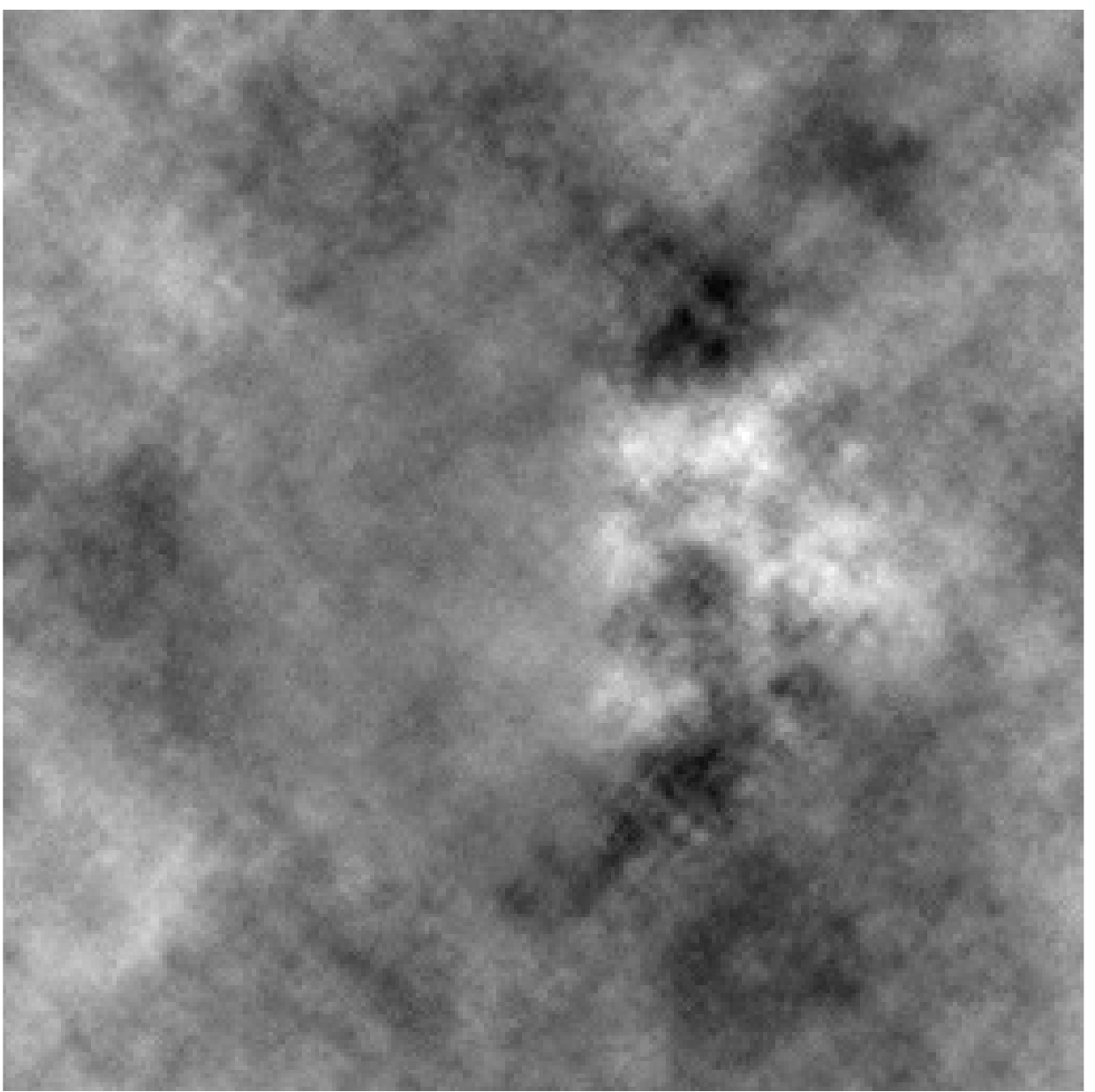} &
\includegraphics[height =2.2cm]{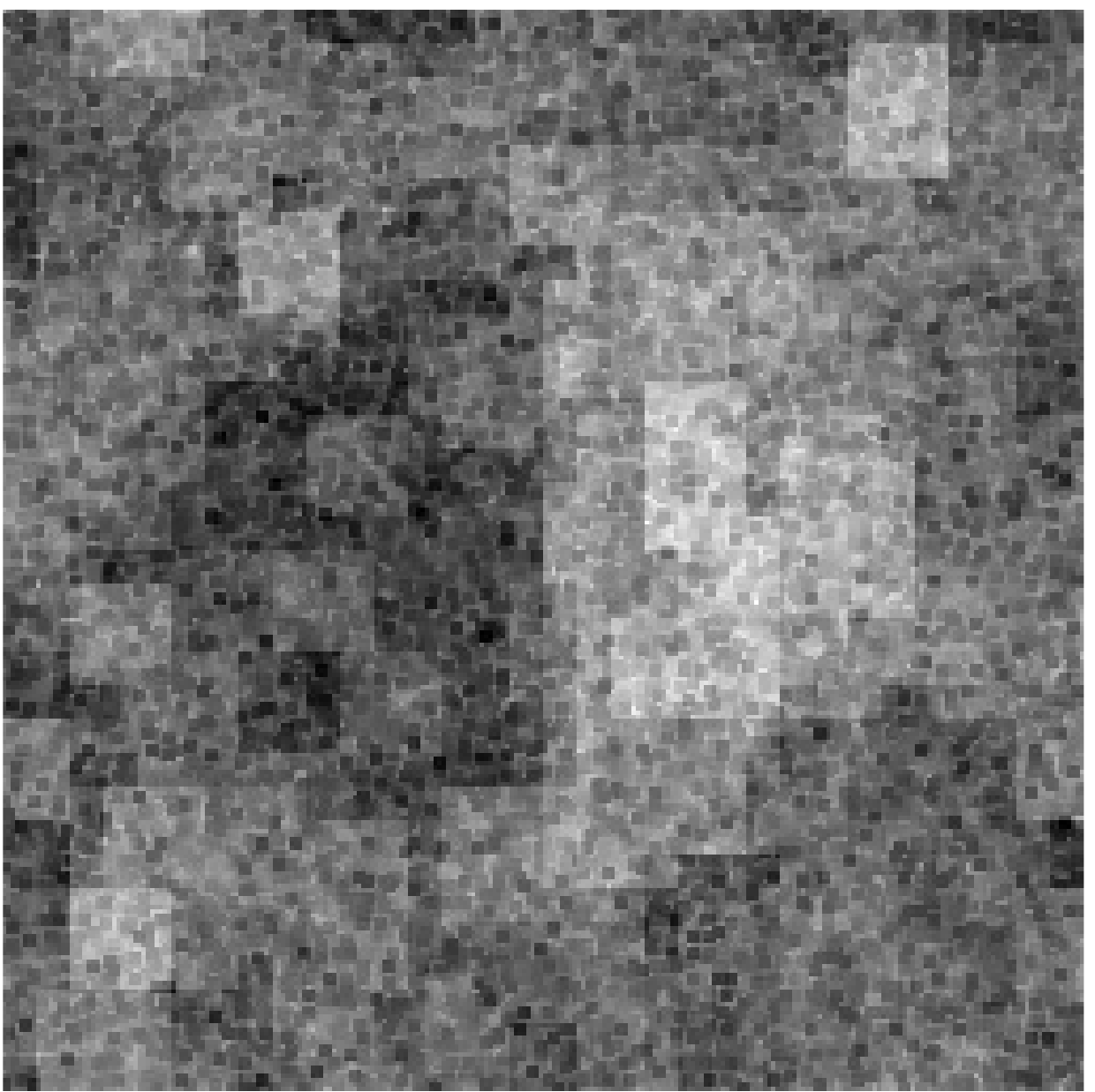}&
\includegraphics[height =2.2cm]{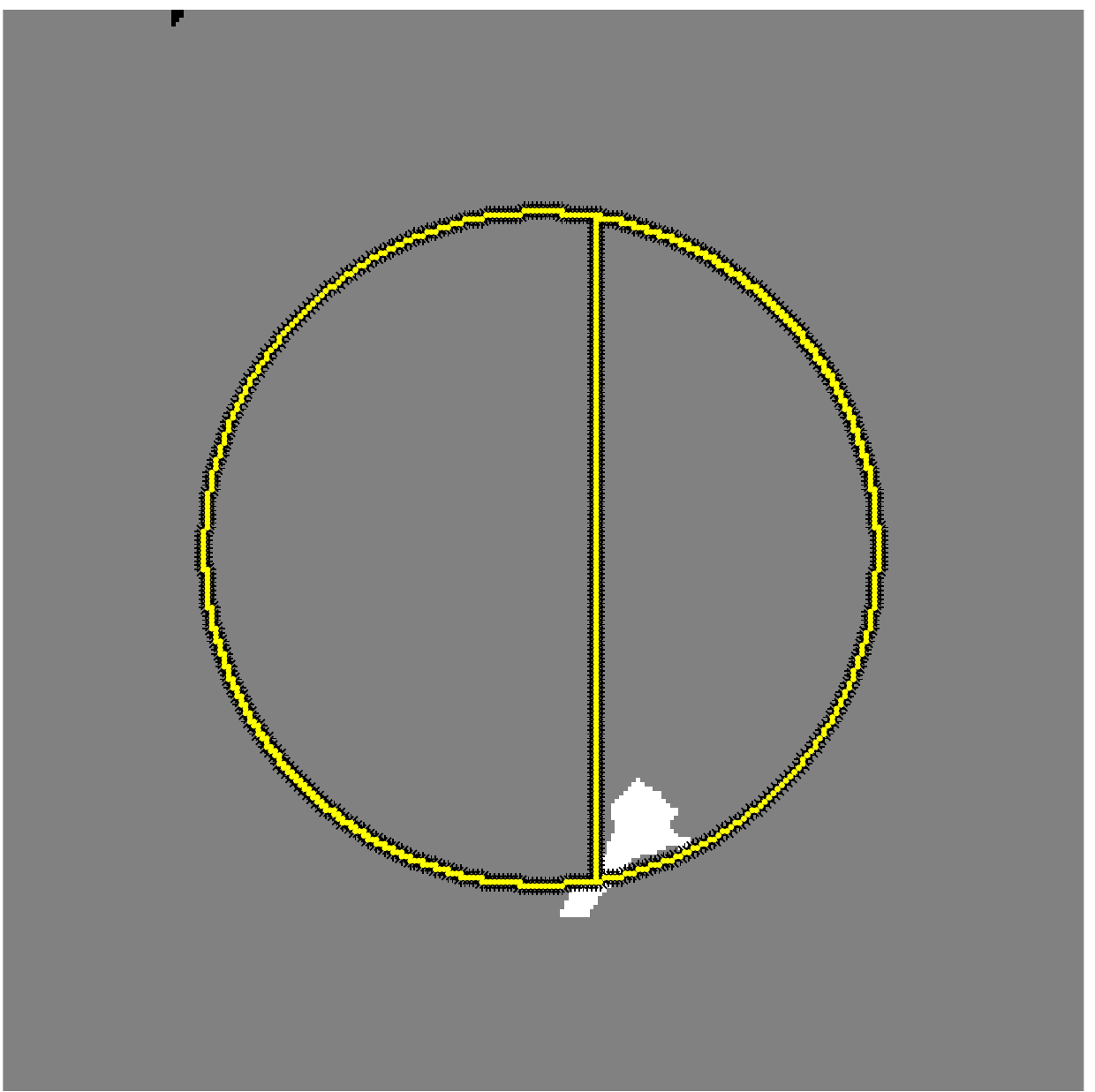} &
\includegraphics[height =2.2cm]{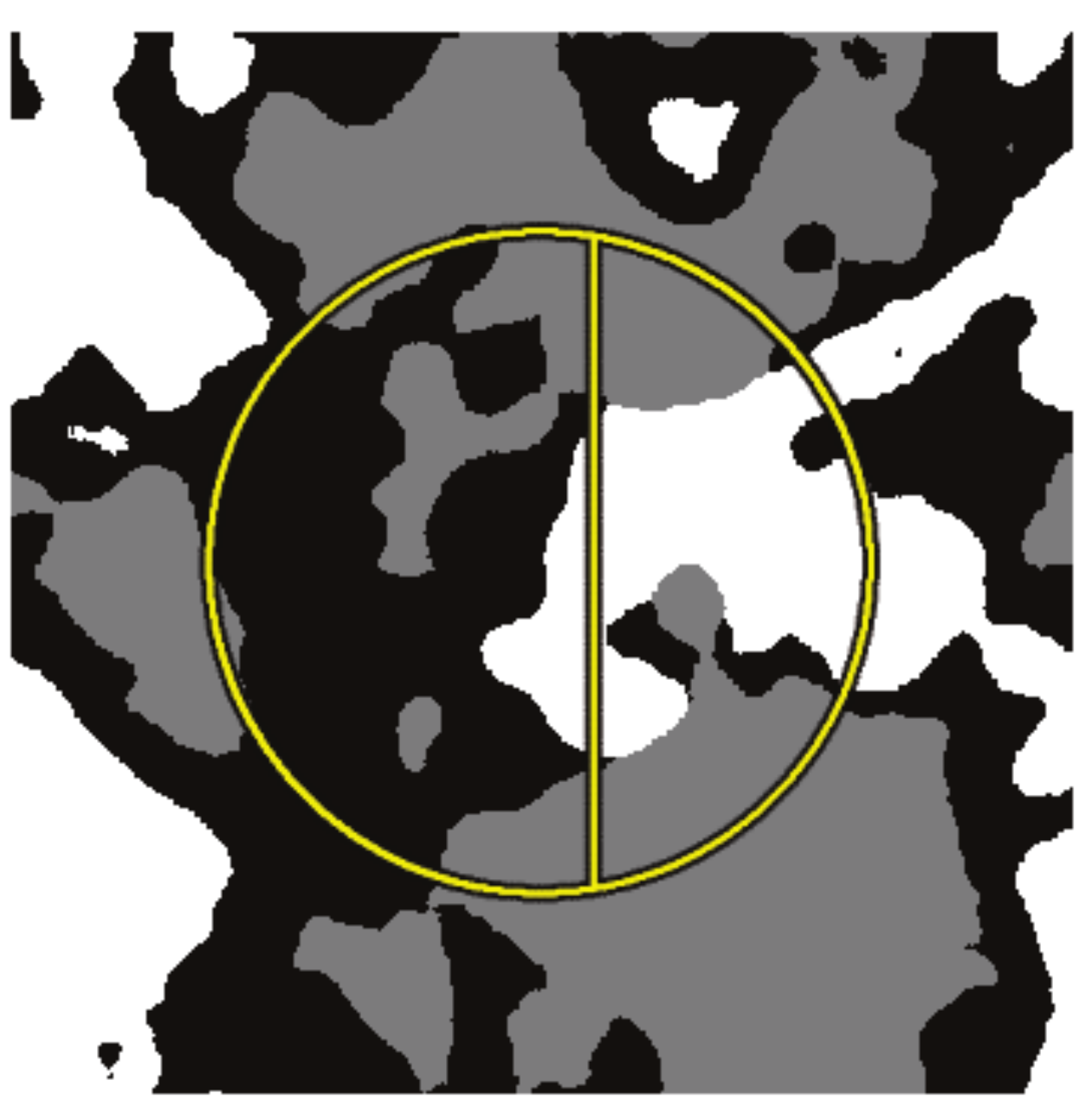}&
\includegraphics[height =2.2cm]{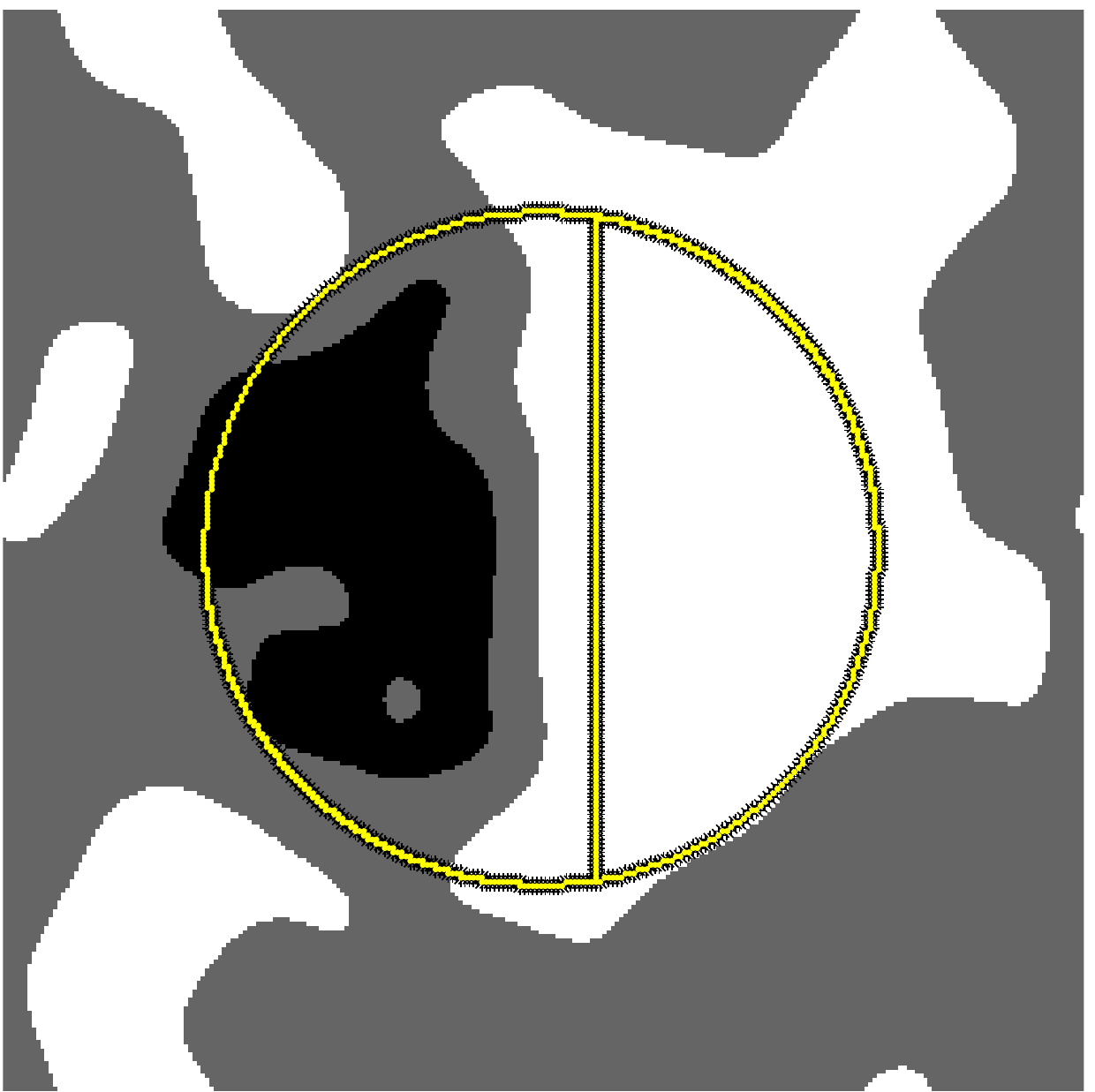} &
\includegraphics[height =2.2cm]{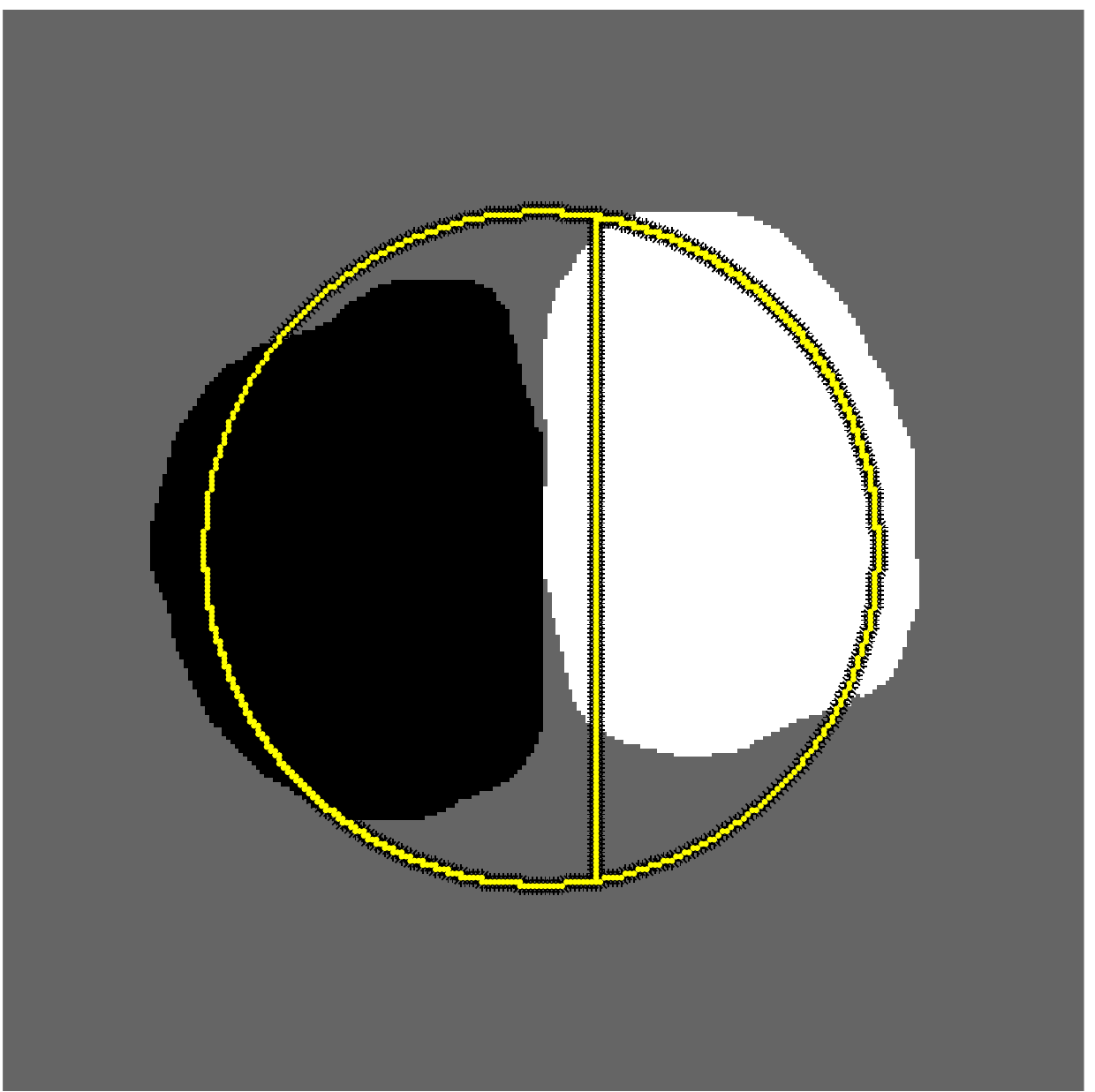} &
\includegraphics[height =2.2cm]{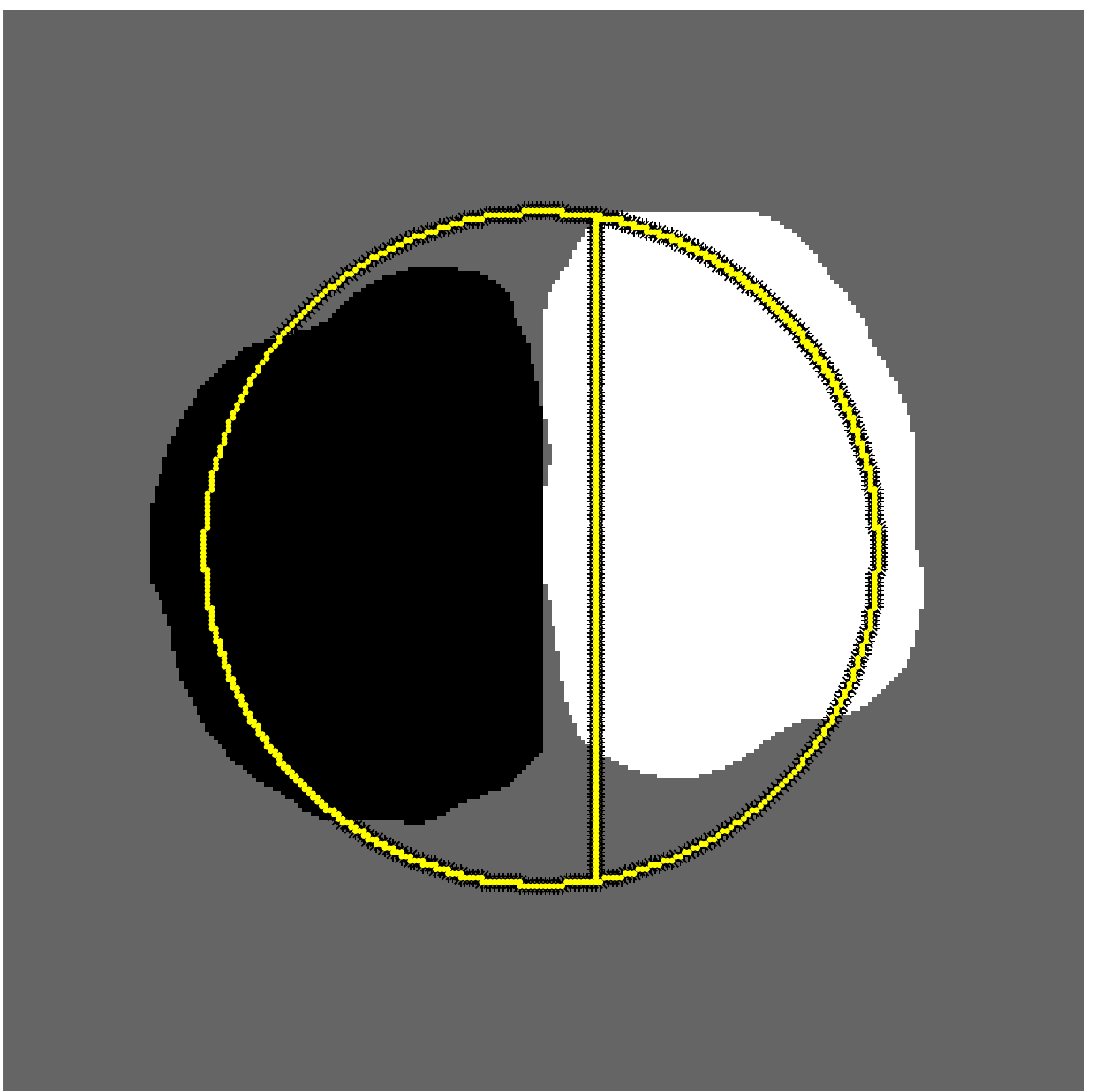} &
\includegraphics[height =2.2cm]{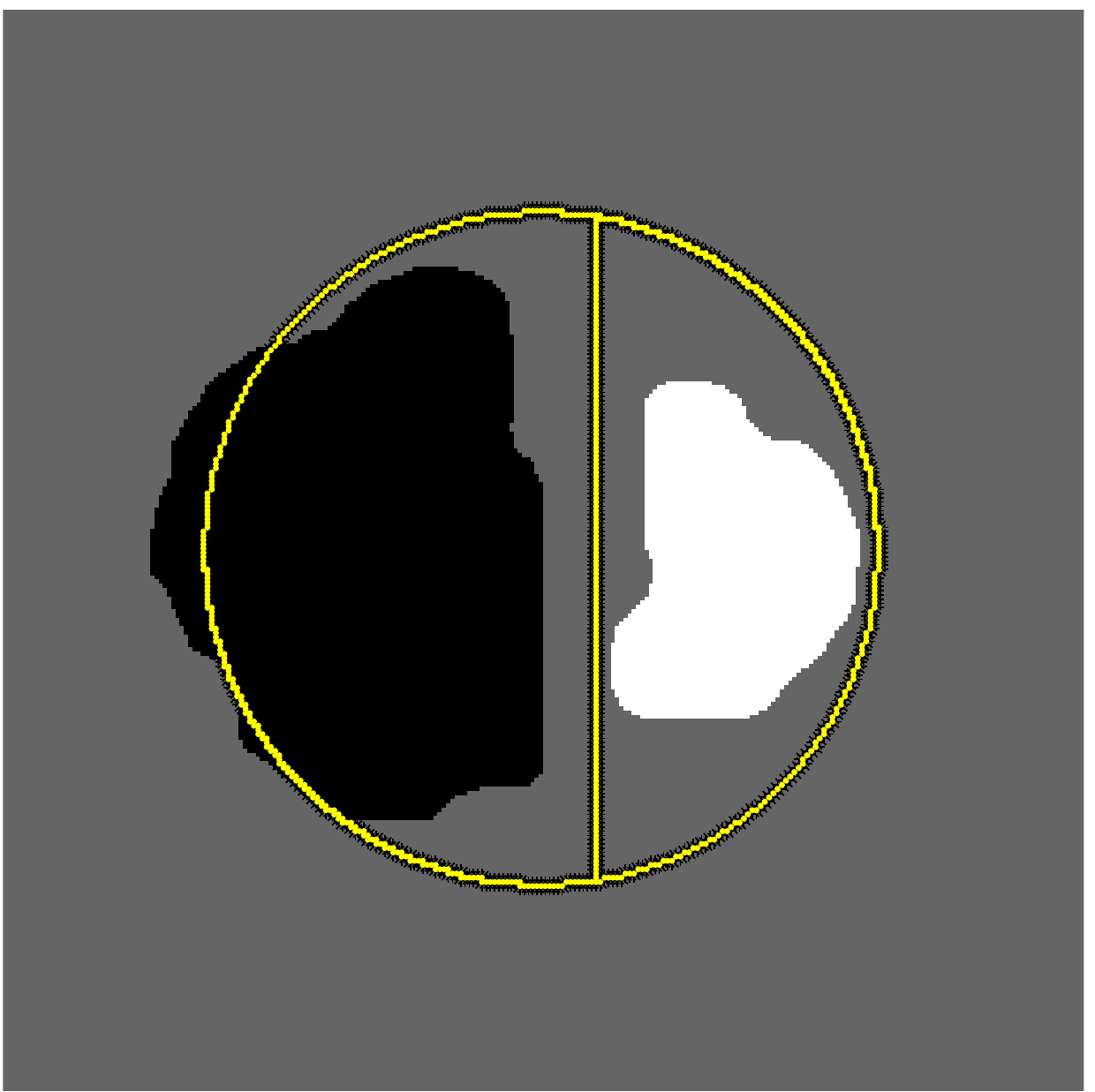}
\vspace{-6mm}\\
&&&\ccol \hfill \small error 20.1\%\;&\ccol \hfill\small error 56.05\%\;&\ccol \hfill\small error 22.3\%\;&\ccol \hfill\small error 7.54\%\;&\ccol \hfill \small error 7.49\%\;&\ccol \hfill\small error 9.19\%\;\\
&(a) Data $\underline{\underline{f}}$ & (b) $\widehat{\underline{\underline{h}}}_{L}$& (c) \cite{Arbelaez_P_2011_j-ieee-tpami_con_dhis} & (d)\cite{Yuan_J_2015_j-ieee-tip_fac_bts} & (e)  $\widehat{\underline{\underline{\Omega}}}^{\textrm{S}}$ &(f)  $\widehat{\underline{\underline{\Omega}}}^{\textrm{TV}}$ & (i)  $\widehat{\underline{\underline{\Omega}}}^{\textrm{TVW}}$ & (j)  $\widehat{\underline{\underline{\Omega}}}^{\textrm{RMS}}$
\end{tabular}
\caption{Results of the labeling obtained with the different proposed solutions when $Q=3$.  \label{fig:results_Q3} }
\end{figure*}

\begin{figure*}[t]
\centering
\begin{tabular}{cc}
\includegraphics[height =5cm]{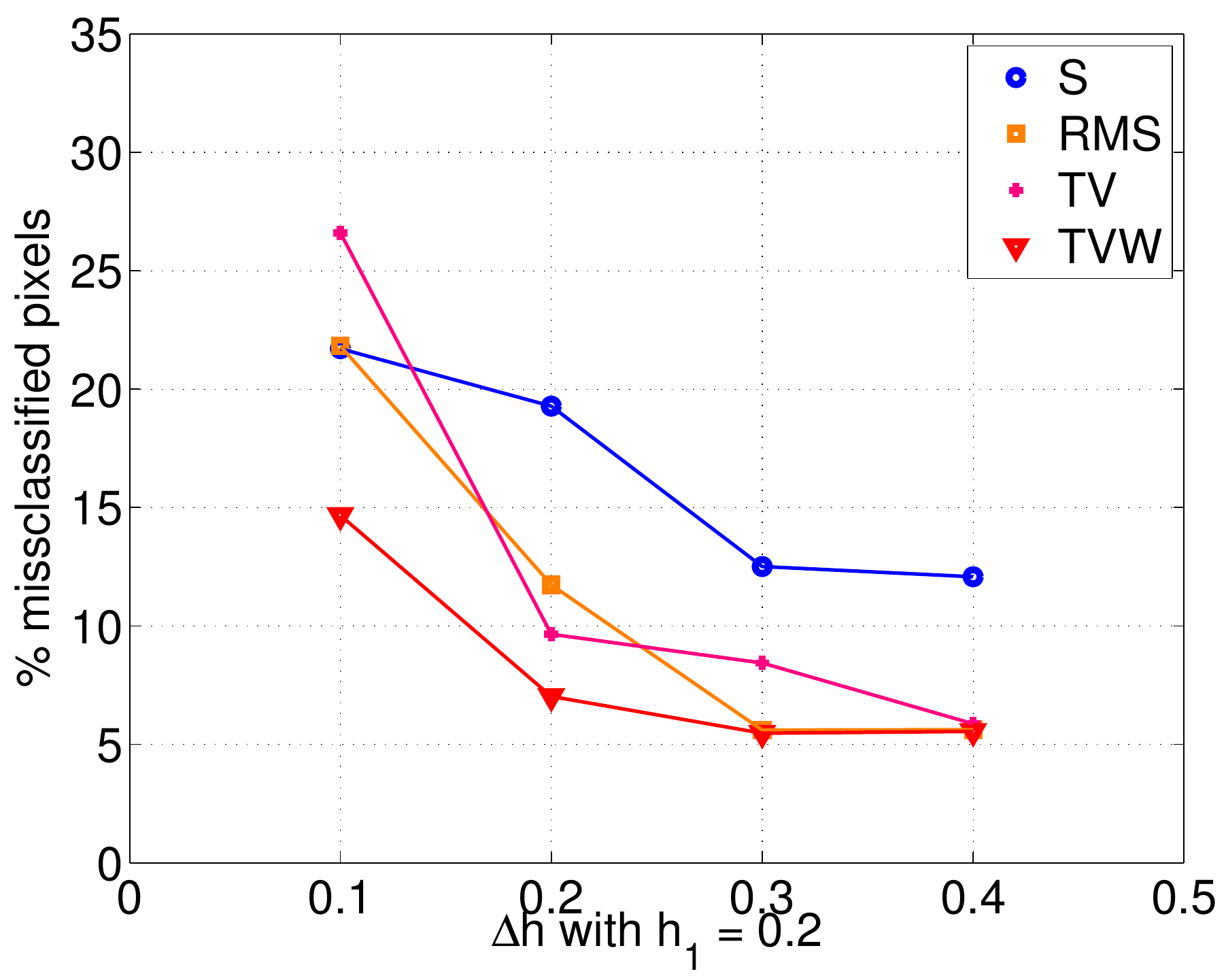} &
\includegraphics[height =5cm]{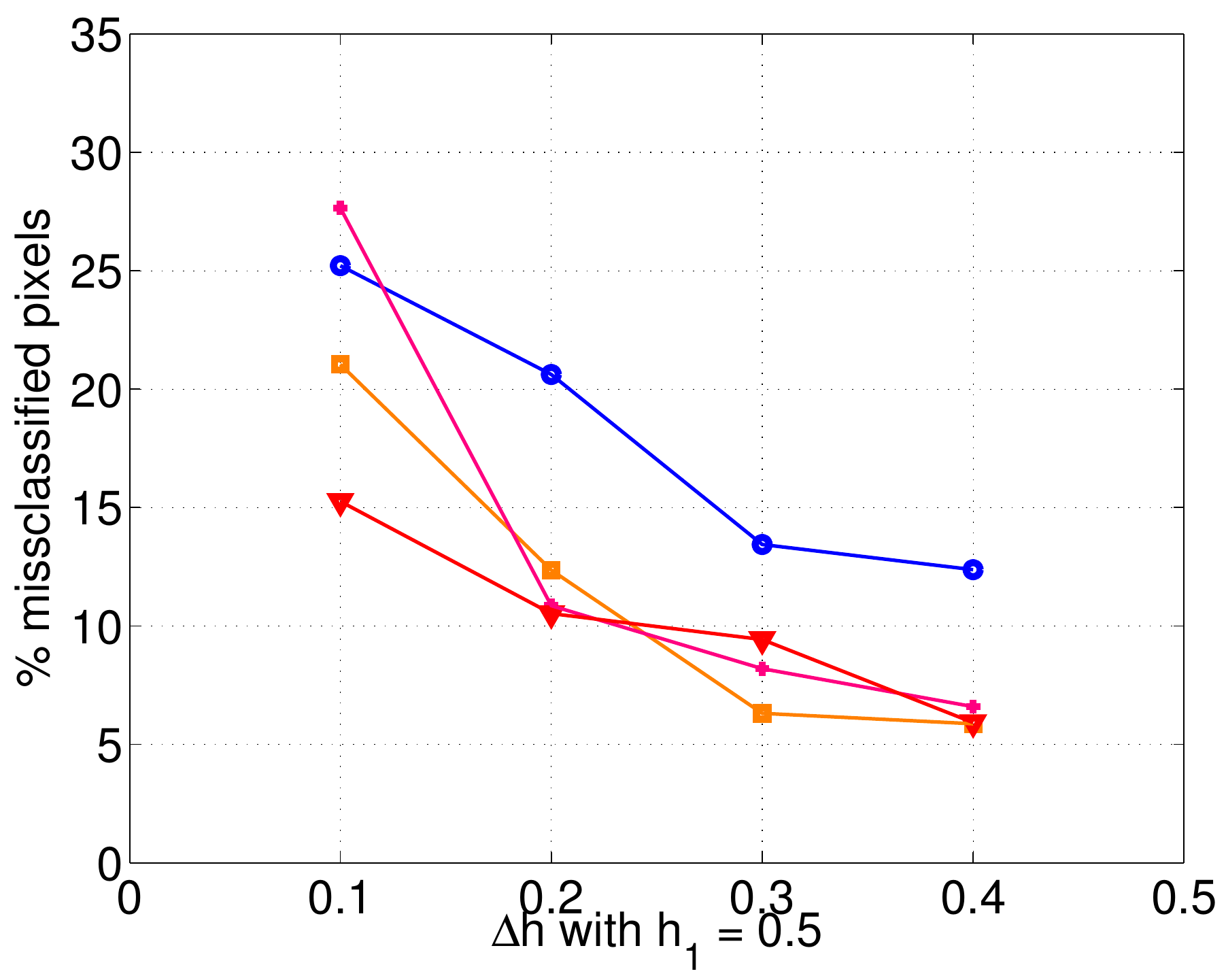}
\end{tabular}
\caption{Results obtained with the different proposed solutions (in red) compared to a basic smoothing (in blue) as a function of $\Delta h = h_2 -h_1$.
\label{fig:delta}
}
\end{figure*}

\subsection{Quantitative performance assessment}
\label{sec:perfsim}

The proposed local regularity based labeling procedures are first compared using 10 independent realizations of multifractional Brownian fields with $Q=2$ areas of constant regularities $h_1$ and $h_2$, respectively, given by the  ellipse model shown in Fig.~\ref{fig:illprinciple}~(b) ($h_2$ corresponding to the inside of the ellipse).
Performance are evaluated for a large range of values of the regularization parameter $\lambda$ (respectively, {standard deviation $\sigma$ for the local Gaussian smoothing based solution}). To assess performance, misclassified pixel rate is evaluated as follows:
The area with the largest median of estimated local regularity values is associated with the area of the original mask with largest regularity, the area with the second largest median of estimated local regularity values with the area of the original mask with second largest regularity value, and so on.
Achieved results are reported in Fig.~\ref{fig:results_lambda} (a) (for $h_1=0.5$ and $h_2=0.7$) and (b) ($h_1=0.6$ and $h_2=0.7$), respectively.

The three TV-based strategies ($\widehat{\underline{\underline{\Omega}}}^{\textrm{TV}}$, $\widehat{\underline{\underline{\Omega}}}^{\textrm{TVW}}$ and $\widehat{\underline{\underline{\Omega}}}^{\textrm{RMS}}$) clearly outperform the local smoothing based solution $\widehat{\underline{\underline{\Omega}}}^{\textrm{S}}$ over a large range of  $\lambda$.
The local smoothing procedure yields at best pixel misclassification rates of $12\%$ (Fig.~\ref{fig:results_lambda} (a)) and $28\%$ (Fig.~\ref{fig:results_lambda} (b)). 
In contrast, the best results obtained with the TV-based strategies drop down to less than $7\%$ (Fig.~\ref{fig:results_lambda} (a)) and $12\%$ (Fig.~\ref{fig:results_lambda} (b)) of misclassified pixels.

Among the TV-based algorithms, $\widehat{\underline{\underline{\Omega}}}^{\textrm{TVW}}$, relying on the joint estimation of regularity and weights, is the least sensitive to the precise selection of the regularization parameter $\lambda$ and consistently yields the best performance.
Notably, it outperforms all other procedures when the difference in regularity $|h_2-h_1|$ is small (Fig.~\ref{fig:results_lambda} (b)).
{%
For segmentation, the performance of the TV procedure ($\widehat{\underline{\underline{\Omega}}}^{\textrm{TV}}$) is similar to that of $\widehat{\underline{\underline{\Omega}}}^{\textrm{TVW}}$, yet $\widehat{\underline{\underline{\Omega}}}^{\textrm{TV}}$ is more sensitive to the precise tuning of $\lambda$ and yields large errors when $\lambda$ is chosen too small or too large.
For these cases, $\widehat{\underline{\underline{\Omega}}}^{\textrm{TV}}$ detects only one area, while $\widehat{\underline{\underline{\Omega}}}^{\textrm{TVW}}$ still segments the texture} {into} {two areas}. Solution $\widehat{\underline{\underline{\Omega}}}^{\textrm{RMS}}$ also is more robust to the tuning of $\lambda$, compared to $\widehat{\underline{\underline{\Omega}}}^{\textrm{TV}}$, yet shows slightly decreased misclassification rates.
Furthermore, $\widehat{\underline{\underline{\Omega}}}^{\textrm{RMS}}$ has the practical advantage of being the only solution that does not require the practically cumbersome step of thresholding histograms (hence avoiding the empirical tuning of binning and smoothing parameters, for instance).%

To illustrate, results obtained by each of the proposed procedures for the value of $\lambda$ (or $\sigma$) leading to a minimal classification error (marked with symbols in Fig.~\ref{fig:results_lambda}) are reported in Fig.~\ref{fig:deltah5-7} when considering one randomly selected realization of multifractional Brownian fields  with $Q=2$ areas of constant regularity defined by $(h_1,h_2)=(0.5,0.7)$ (top) and $(h_1,h_2)=(0.6,0.7)$ (bottom), respectively. Fig.~\ref{fig:deltah5-7} (a) shows the analyzed textures, illustrating that the two texture areas can not be distinguished visually. The local smoothing based labeling results $\widehat{\underline{\underline{\Omega}}}^{\textrm{S}}$ are clearly the poorest, both for  $(h_1,h_2)=(0.5,0.7)$ and $(h_1,h_2)=(0.6,0.7)$.
In contrast, all three TV-based solutions $\widehat{\underline{\underline{\Omega}}}^{\textrm{TV}}$, $\widehat{\underline{\underline{\Omega}}}^{\textrm{TVW}}$ and $\widehat{\underline{\underline{\Omega}}}^{\textrm{RMS}}$ yield satisfactory performance.
Solution $\widehat{\underline{\underline{\Omega}}}^{\textrm{TVW}}$ achieves the lowest misclassification error and is also visually the most convincing in terms of segmented regions, at the price though of the largest computational cost.
Solution $\widehat{\underline{\underline{\Omega}}}^{\textrm{TV}}$ shows only slightly larger misclassification rates sand may be preferred in certain applications for its significantly smaller computational cost.
Solution $\widehat{\underline{\underline{\Omega}}}^{\textrm{RMS}}$ shows larger misclassification rate
when the difference in regularity decreases.

Moreover, the above analysis is complemented by the study of a situation with of $Q=3$ areas (constant local regularities $(h_1,h_2,h_3)=(0.2,0.4,0.7)$), with a more complex geometry, including notably corners.
The regularity mask, texture and labeling results are illustrated in Fig.~\ref{fig:results_Q3}.
The local smoothing solution $\widehat{\underline{\underline{\Omega}}}^{\textrm{S}}$ fails to distinguish the three areas, yielding several disconnected domains, while all proposed TV based approaches correctly and satisfactorily detect the three distinct areas.
Solutions $\widehat{\underline{\underline{\Omega}}}^{\textrm{TV}}$ and $\widehat{\underline{\underline{\Omega}}}^{\textrm{TVW}}$ show similar results and again have slightly smaller classification error rates than $\widehat{\underline{\underline{\Omega}}}^{\textrm{RMS}}$.
None of the proposed procedures recovers the two sharp corners of the original regularity mask.
In particular, $\widehat{\underline{\underline{\Omega}}}^{\textrm{TV}}$ and $\widehat{\underline{\underline{\Omega}}}^{\textrm{TVW}}$ yield segmented areas with pronouncedly smooth borders.

In conclusion, the lowest misclassification rates are achieved by $\widehat{\underline{\underline{\Omega}}}^{\textrm{TVW}}$, followed by $\widehat{\underline{\underline{\Omega}}}^{\textrm{TV}}$ and $\widehat{\underline{\underline{\Omega}}}^{\textrm{RMS}}$, and worst results are obtained with $\widehat{\underline{\underline{\Omega}}}^{\textrm{S}}$.
The quality of the solutions obtained with the different strategies are related to their complexity, reflected by computation times.
While $\widehat{\underline{\underline{\Omega}}}^{\textrm{S}}$ is obtained in less than 1 second,
{the fastest TV} based solution is $\widehat{\underline{\underline{\Omega}}}^{\textrm{TV}}$, with a computational time  of about 10 seconds in our experiment, followed by $\widehat{\underline{\underline{\Omega}}}^{\textrm{RMS}}$ with a 1-minute computational time.
The solution achieving the best performance, $\widehat{\underline{\underline{\Omega}}}^{\textrm{TVW}}$, further requires around 3 minutes of computational time.

In order to evaluate more accurately the behaviour of the proposed method as a function of $\Delta h = h_2 - h_1$, 
additional experiments are conducted in which $h_2 = h_1+ \Delta h$ for $\Delta h = \{0.1,0.2,0.3,0.4\}$. Average (over 10 realizations) pixel misclassification rates for two different situations, $h_1=0.2$ and $h_1=0.5$, are reported in Fig~\ref{fig:delta} (the reported results are obtained with values of $\lambda$ that lead to best performance for the configurations).
The results based on TV are displayed in hot colors (orange/pink/red) while the results obtained with the basic smoothing is displayed in blue. 
As expected, the larger $\Delta h$, the better the performance for all methods.
Moreover, the results lead to the conclusion that the TV based approaches have superior performance and that the method that estimates simultaneously the weights and the H\"older exponent, i.e. $\widehat{\underline{\underline{\Omega}}}^{\textrm{TVW}}$, yields overall smallest misclassification rates.

\subsection{Comparisons with state-of-the-art segmentation procedures and application to real-world textures.}

\noindent \textcolor{black}{\noindent{\bf Comparisons with state-of-the-art segmentation procedures.} \quad
Comparisons against two texture segmentation procedures, chosen because considered state-of-the-art in the dedicated literature, are now discussed.
The first approach \cite{Arbelaez_P_2011_j-ieee-tpami_con_dhis} relies on a multiscale contour detection procedure using brightness, color and texture (using textons) information followed by the computation of an oriented watershed transform.}
The second approach  \cite{Yuan_J_2015_j-ieee-tip_fac_bts} relies on Gabor coefficients as features followed by a feature selection procedure relying on a matrix factorization step.
Results are reported in Fig.~\ref{fig:results_lambda}, \ref{fig:deltah5-7}, and \ref{fig:results_Q3} and unambiguously show that such approaches lead to much poorer segmentation performance 
as compared to the proposed TV-based procedures and appear ineffective for the segmentation of scale-free textures.

\begin{figure*}
\centering
\begin{tabular}{ccccc}
\includegraphics[height =3.05cm]{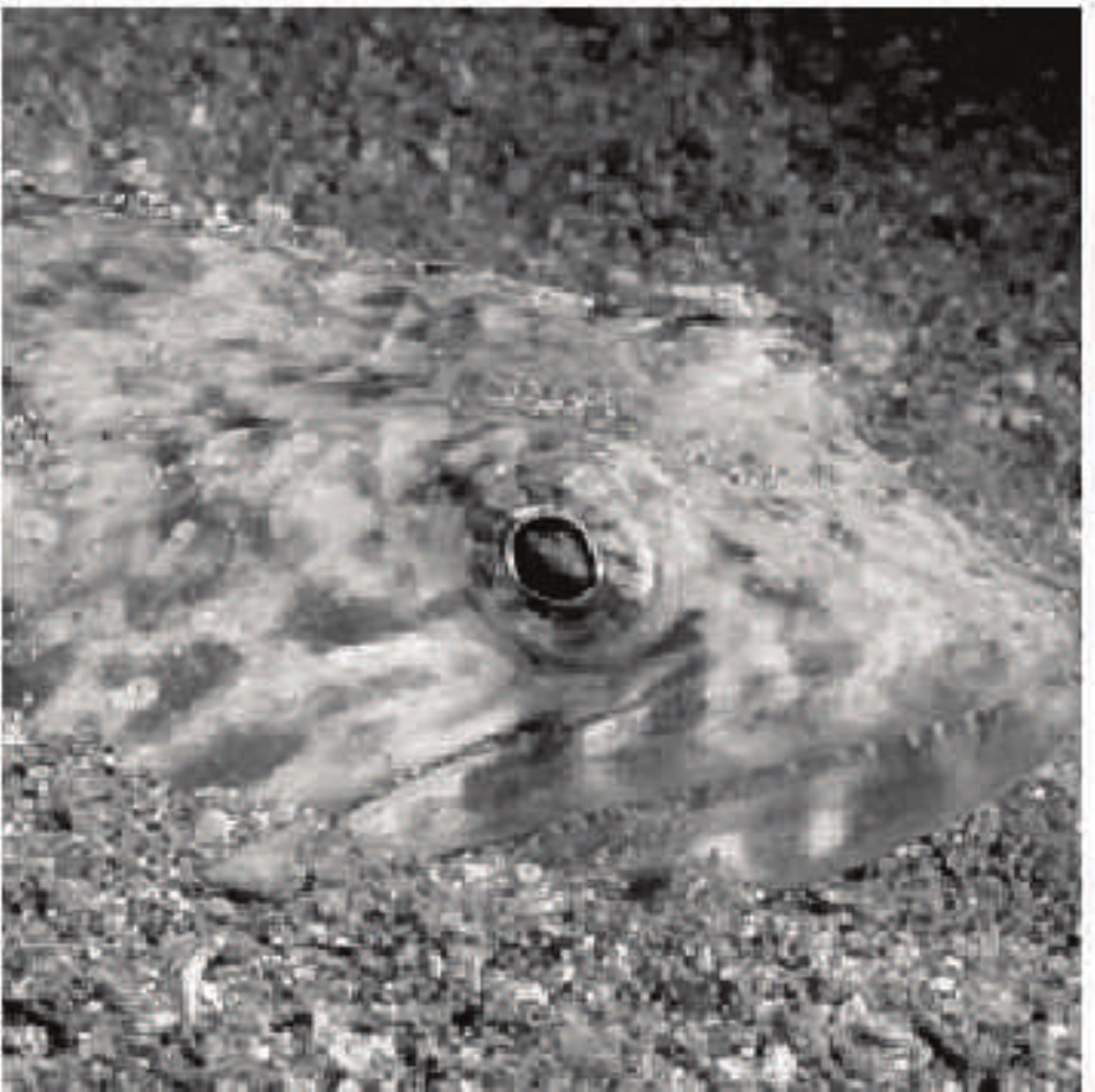} &
\includegraphics[height =3.05cm]{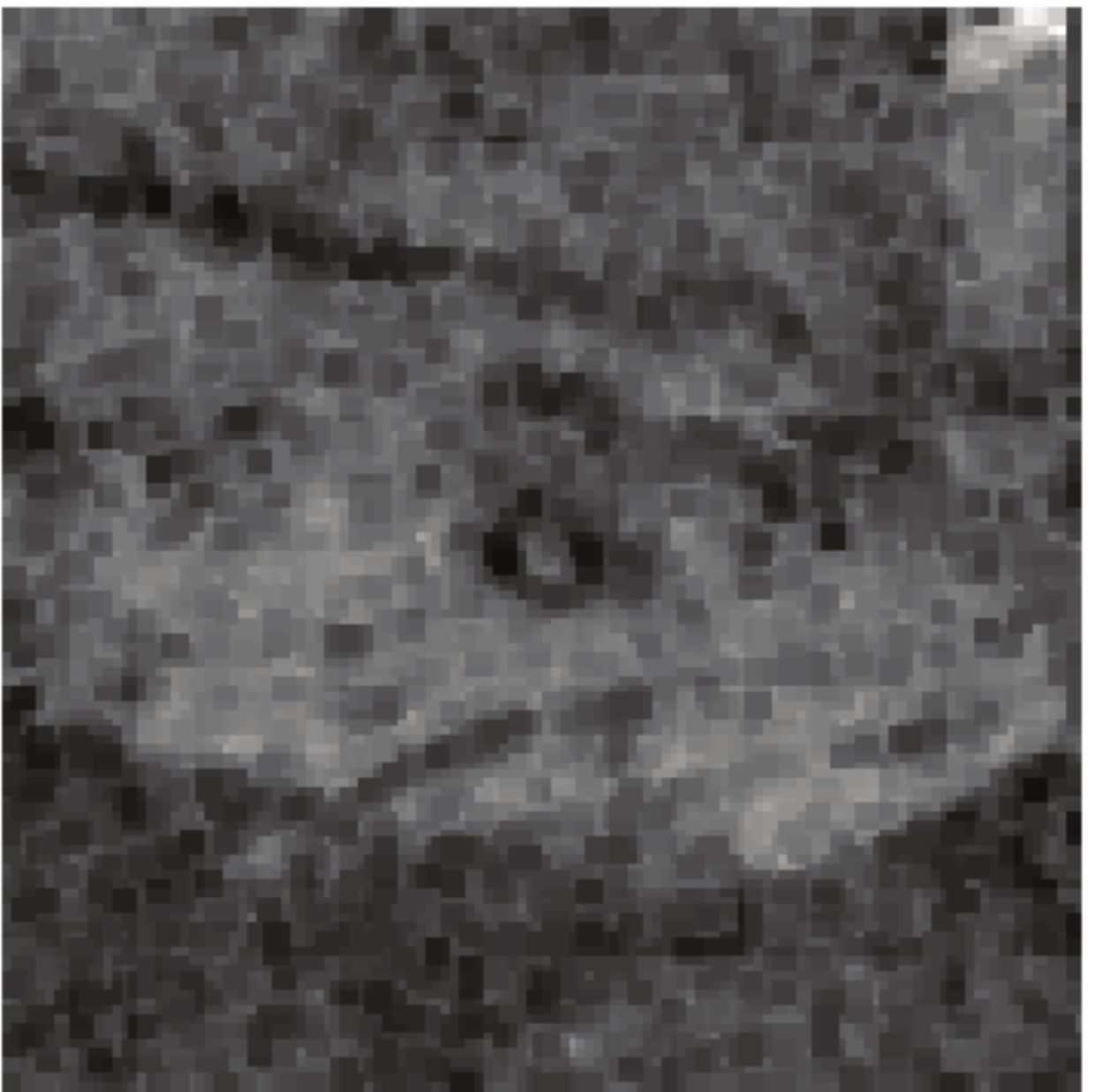}  & \includegraphics[height =3.05cm]{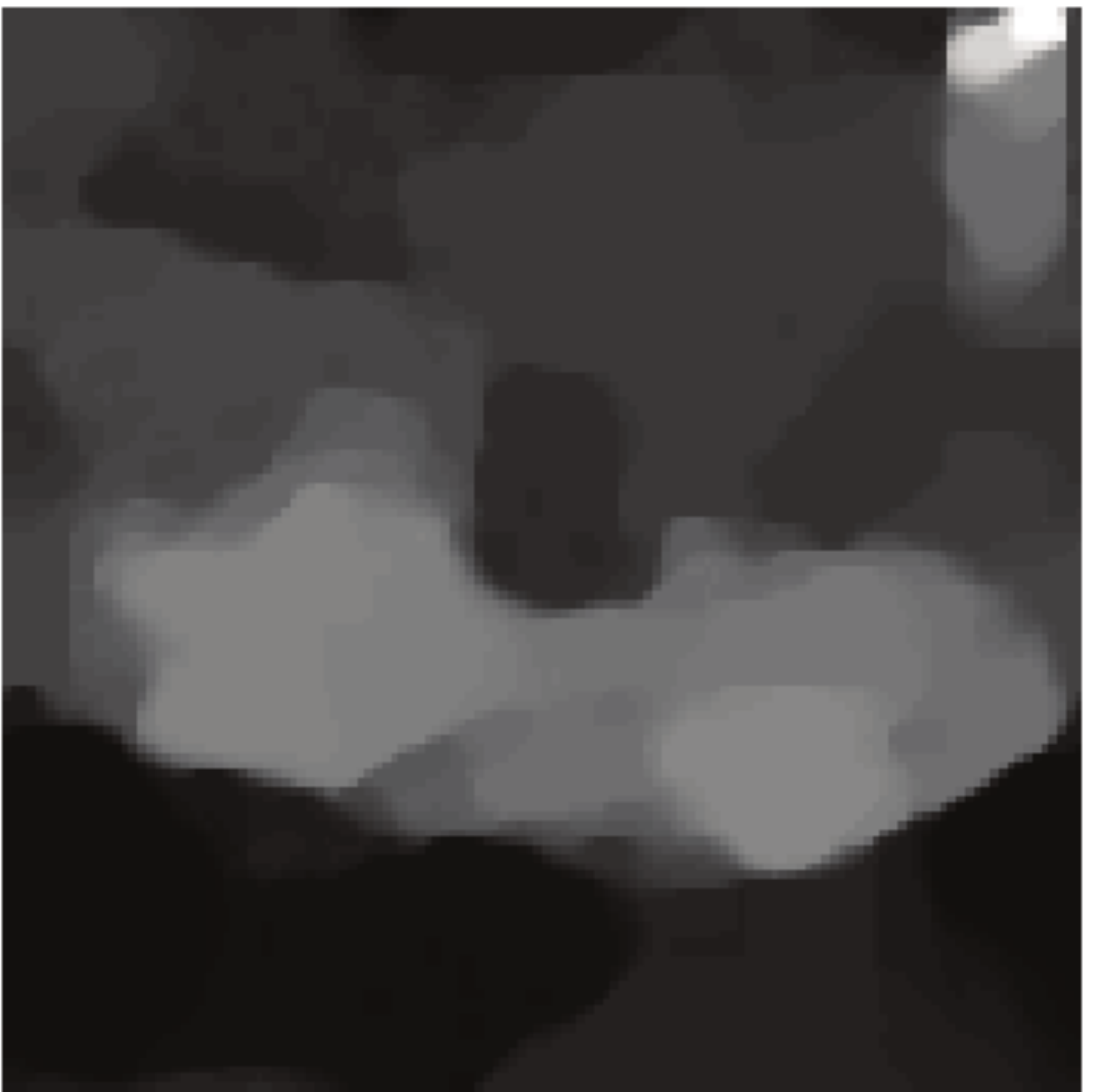}  &
\includegraphics[height =3.05cm]{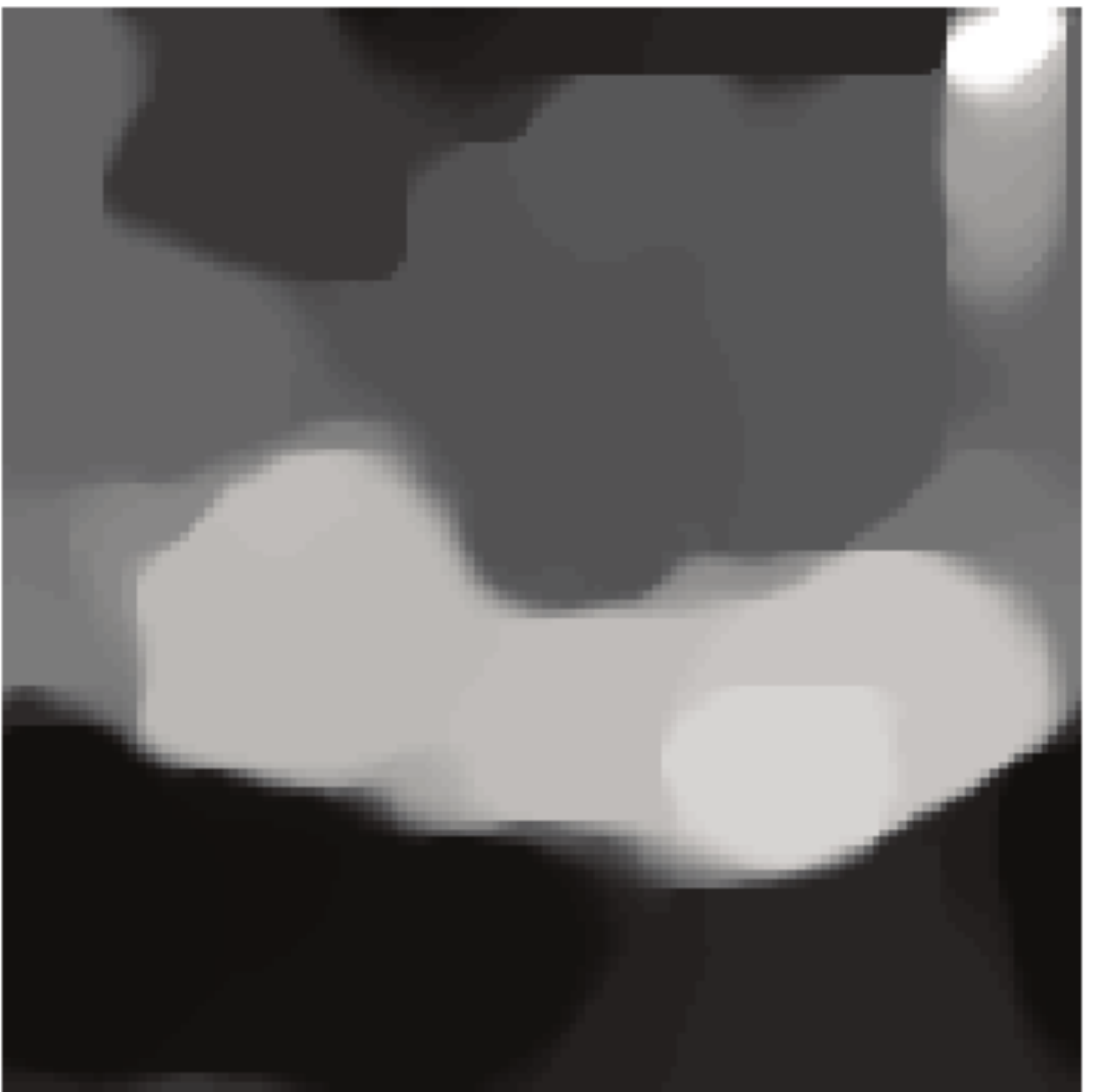}  &\\
 Data $\underline{\underline{f}}$ &  $\widehat{\underline{\underline{h}}}_{L}$ & $\widehat{\underline{\underline{h}}}_L^{\textrm{TV}}$&  $\widehat{\underline{\underline{h}}}_L^{\textrm{TVW}}$ & \\
\includegraphics[height =3.05cm]{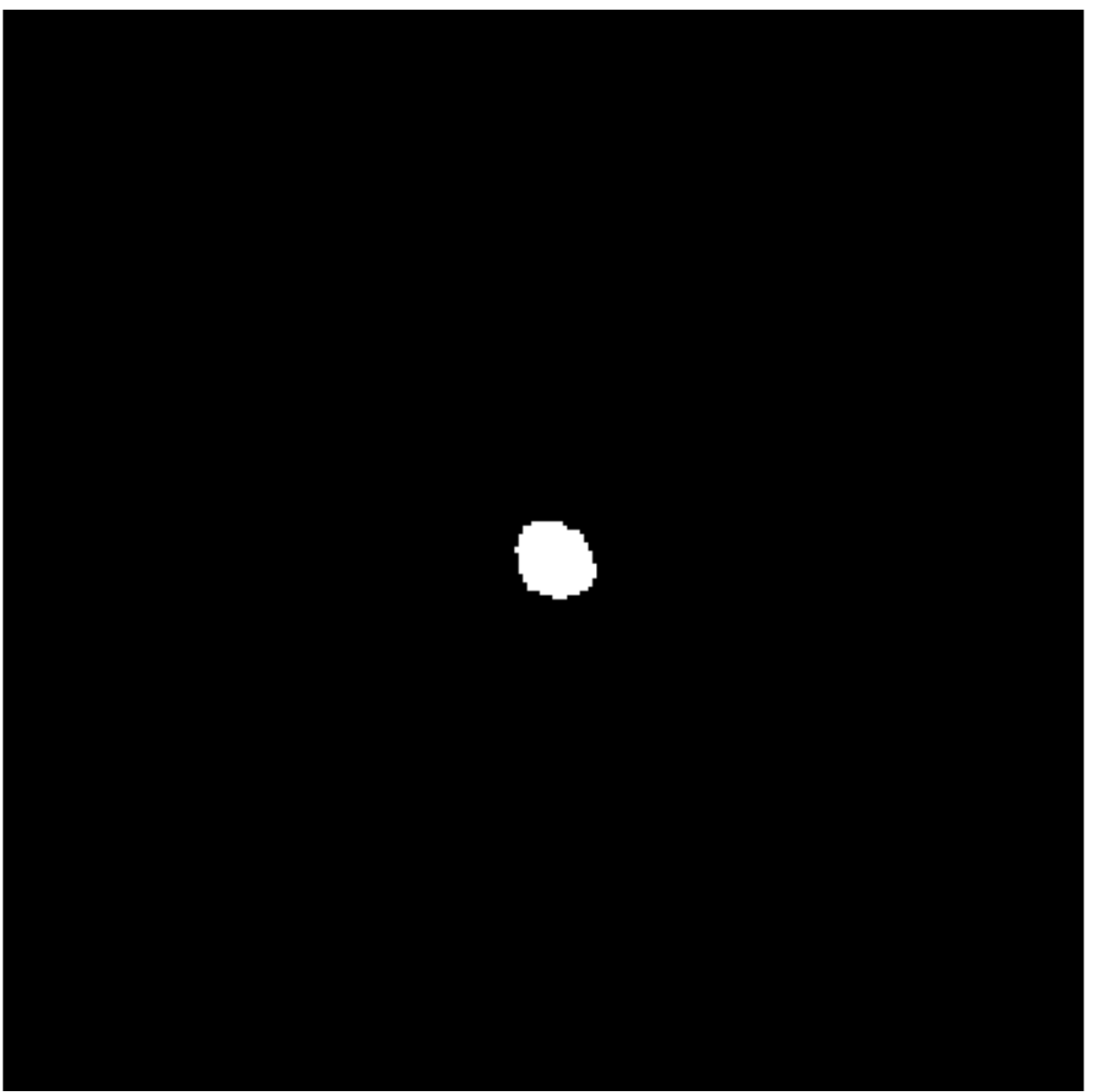} &
\includegraphics[height =3.05cm]{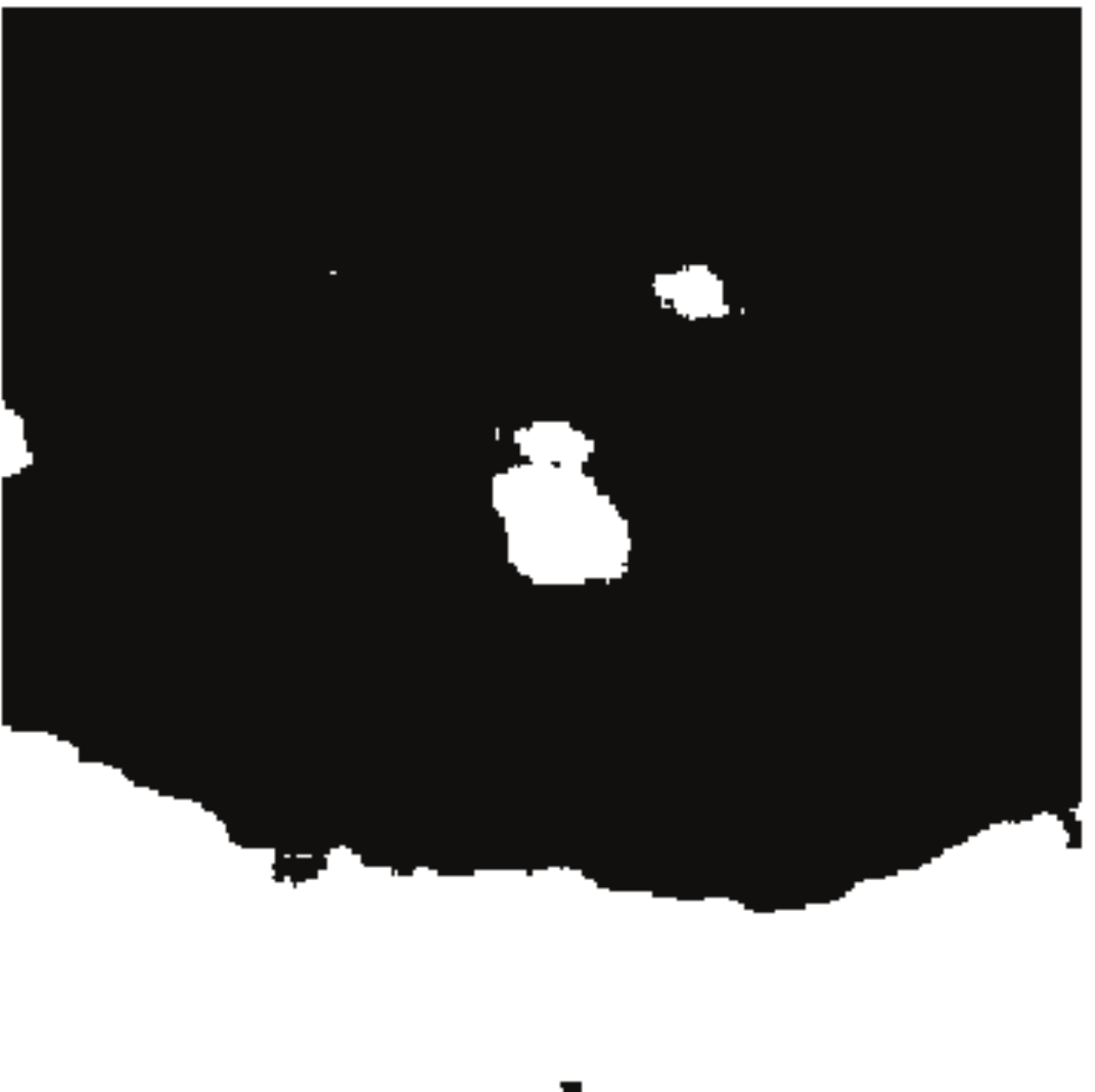} &
\includegraphics[height =3.05cm]{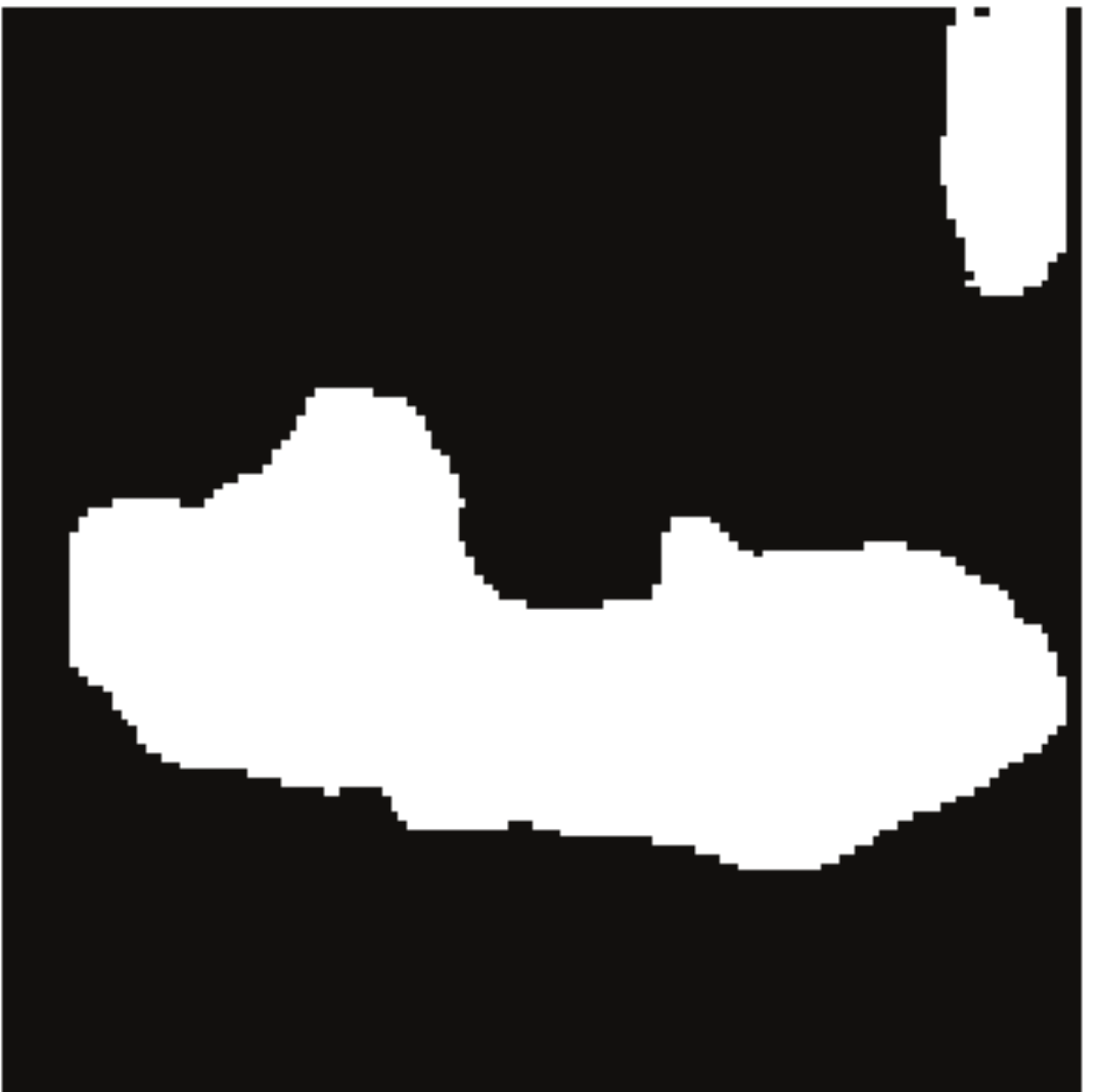} &
\includegraphics[height =3.05cm]{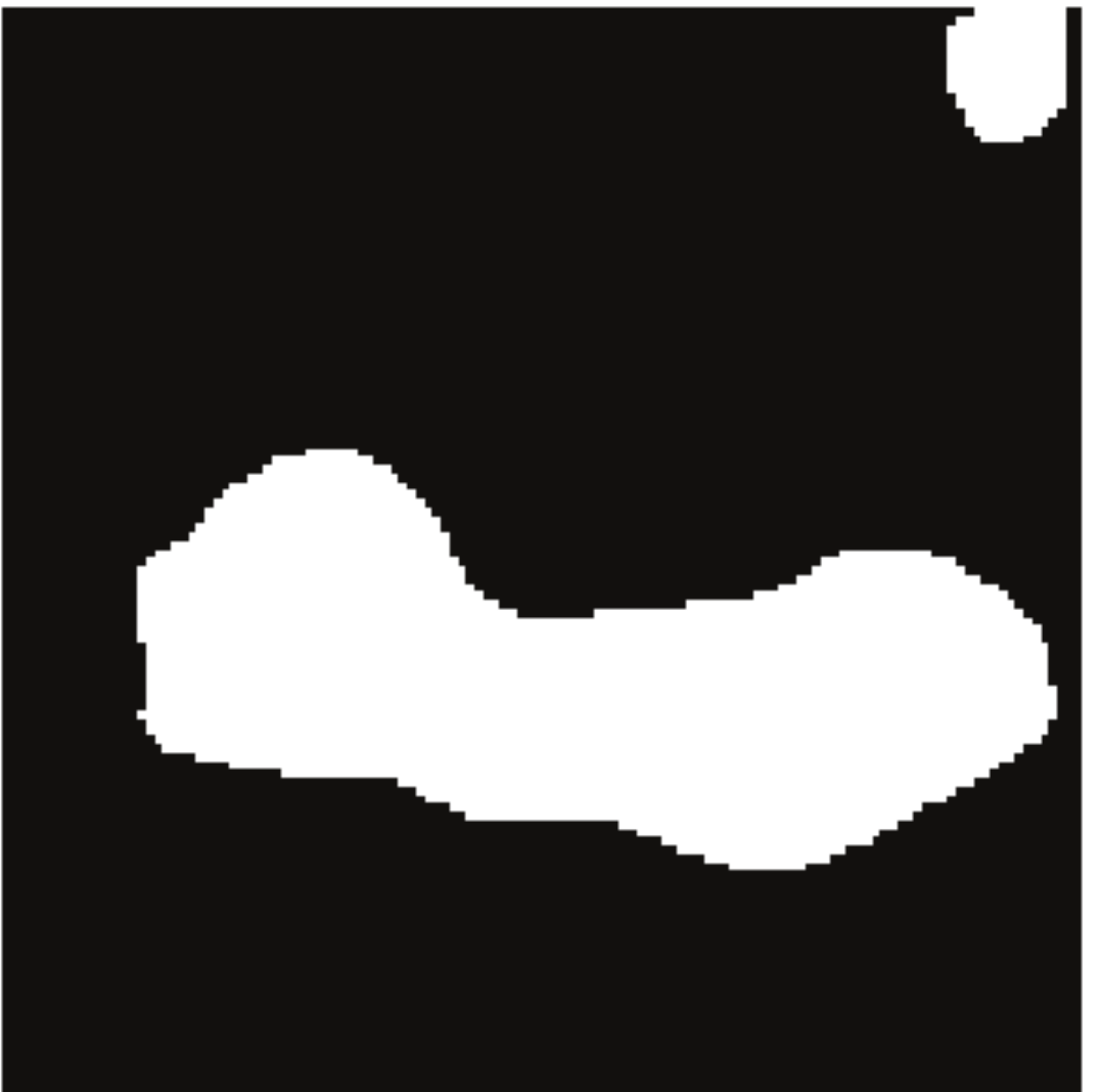}  &
\includegraphics[height =3.05cm]{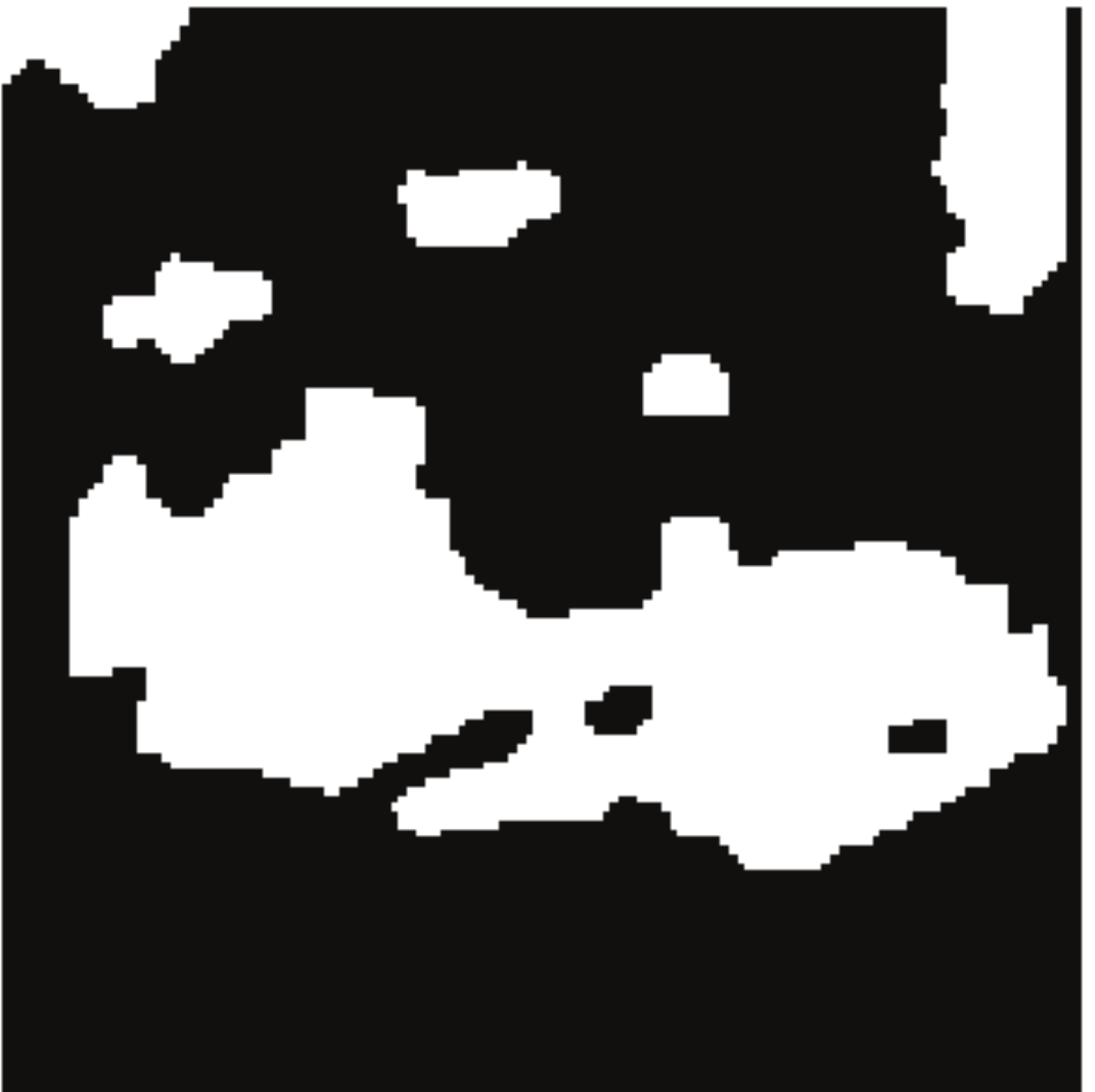}  \\
\cite{Arbelaez_P_2011_j-ieee-tpami_con_dhis}  with $Q=2$&\cite{Yuan_J_2015_j-ieee-tip_fac_bts}  with $Q=2$& $\widehat{\underline{\underline{\Omega}}}^{\textrm{TV}}$ with $Q=2$ &  $\widehat{\underline{\underline{\Omega}}}^{\textrm{TVW}}$  with $Q=2$&   $\widehat{\underline{\underline{\Omega}}}^{\textrm{RMS}}$ with $Q=2$\\
\includegraphics[height =3.05cm]{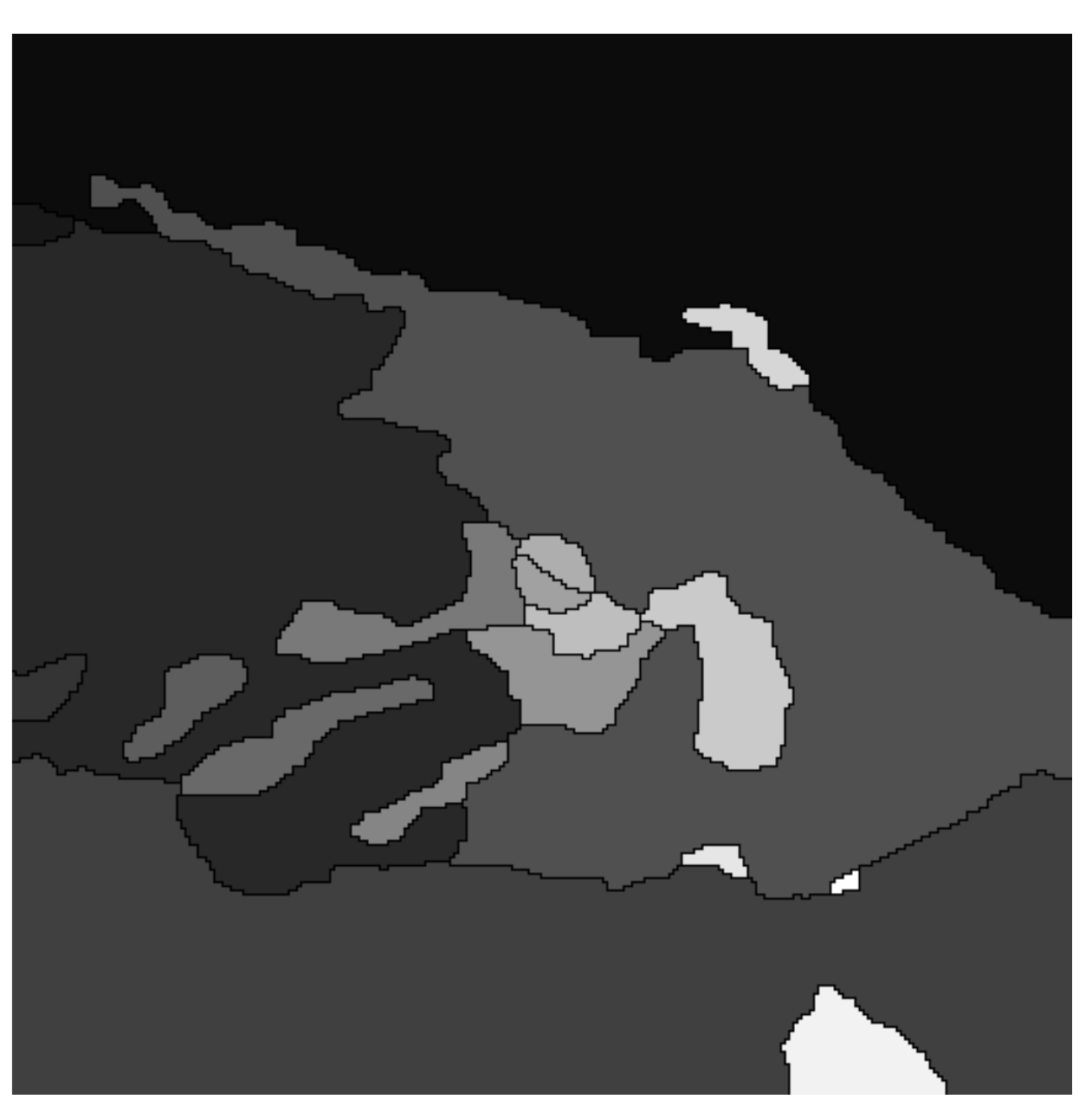} & \includegraphics[height =3.05cm]{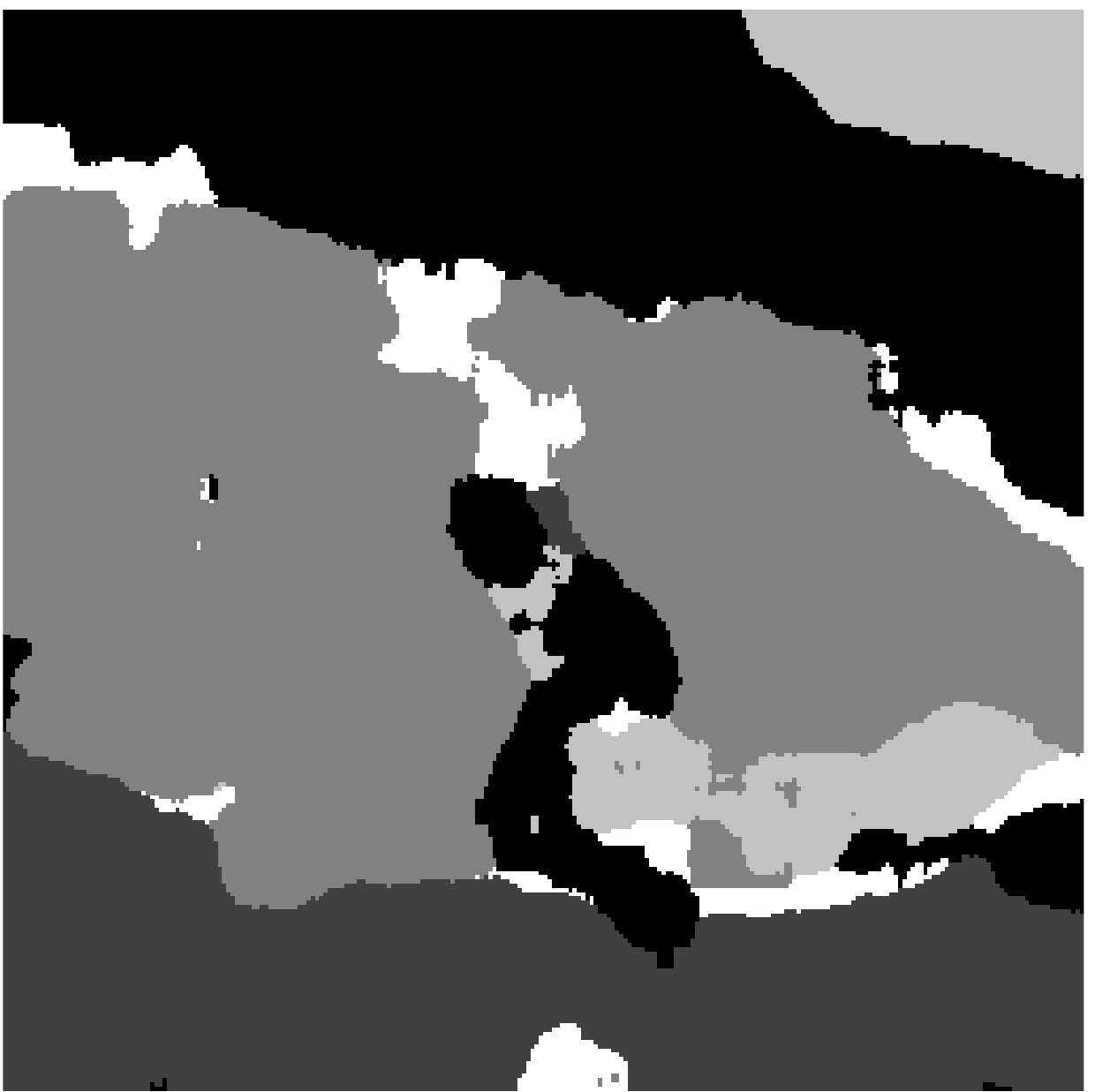} &\includegraphics[height =3.05cm]{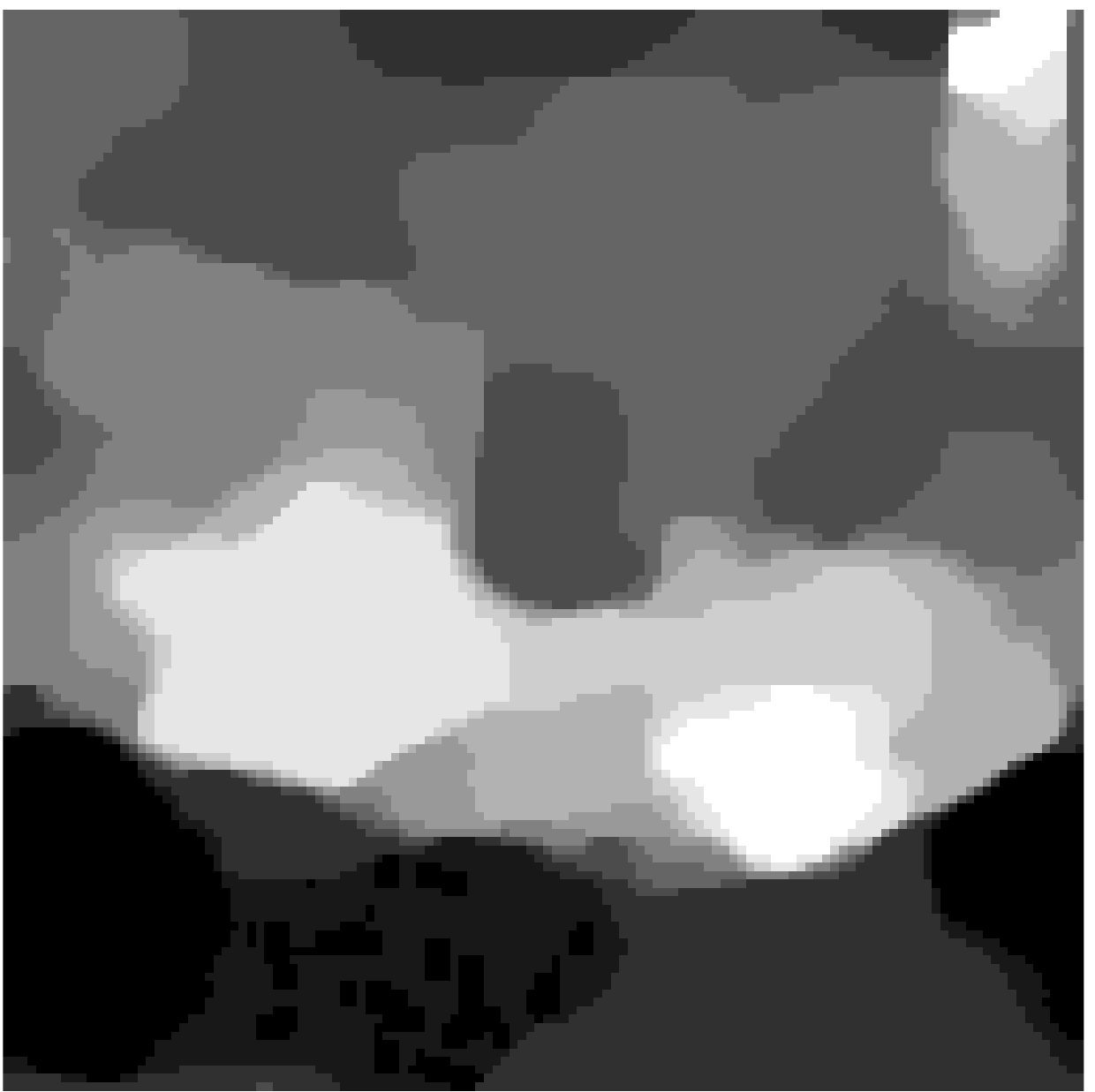}  &
\includegraphics[height =3.05cm]{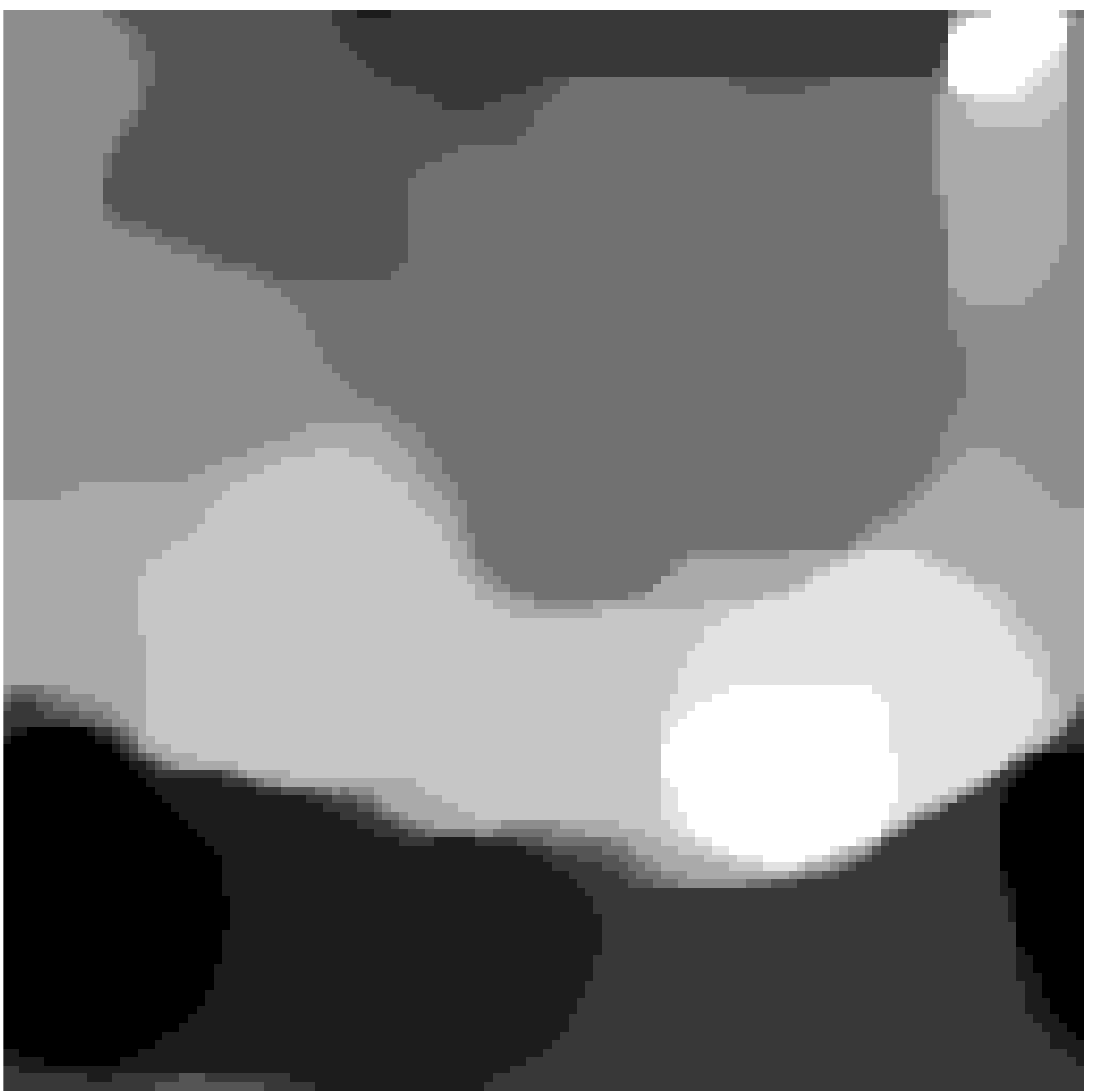} & \includegraphics[height =3.05cm]{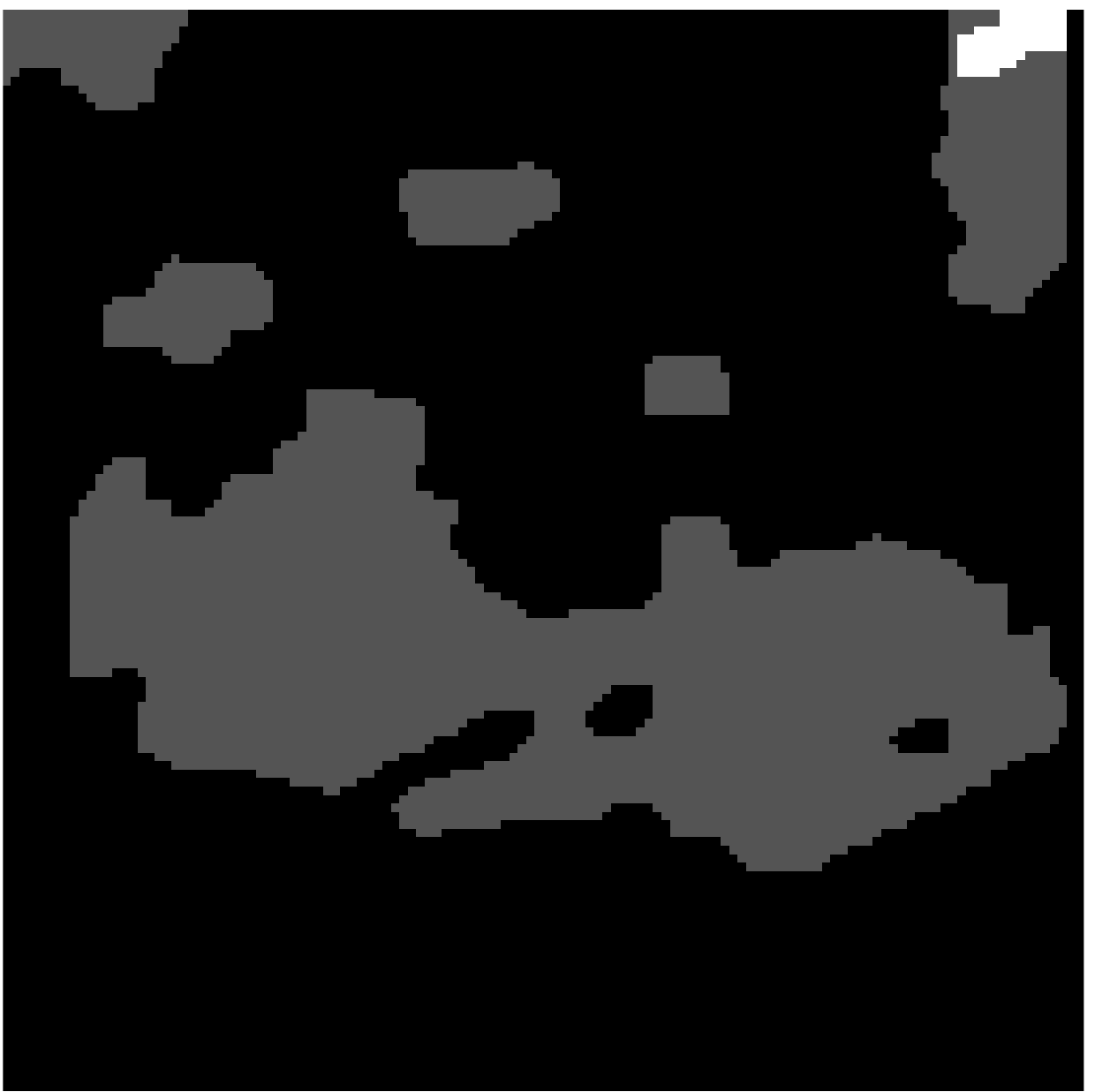} \\
\cite{Arbelaez_P_2011_j-ieee-tpami_con_dhis} with $Q=9$& \cite{Yuan_J_2015_j-ieee-tip_fac_bts} with $Q=5$& $\widehat{\underline{\underline{\Omega}}}^{\textrm{TV}}$ with $Q=10$ &  $\widehat{\underline{\underline{\Omega}}}^{\textrm{TVW}}$  with $Q=9$&   $\widehat{\underline{\underline{\Omega}}}^{\textrm{RMS}}$ with $Q=3$
\end{tabular}
\caption{Experiments on real data: image extracted from the Berkeley Segmentation Database.  \label{fig:results_real1} }
\end{figure*}

\begin{figure*}
\centering
\begin{tabular}{ccccc}
\includegraphics[height =3.05cm]{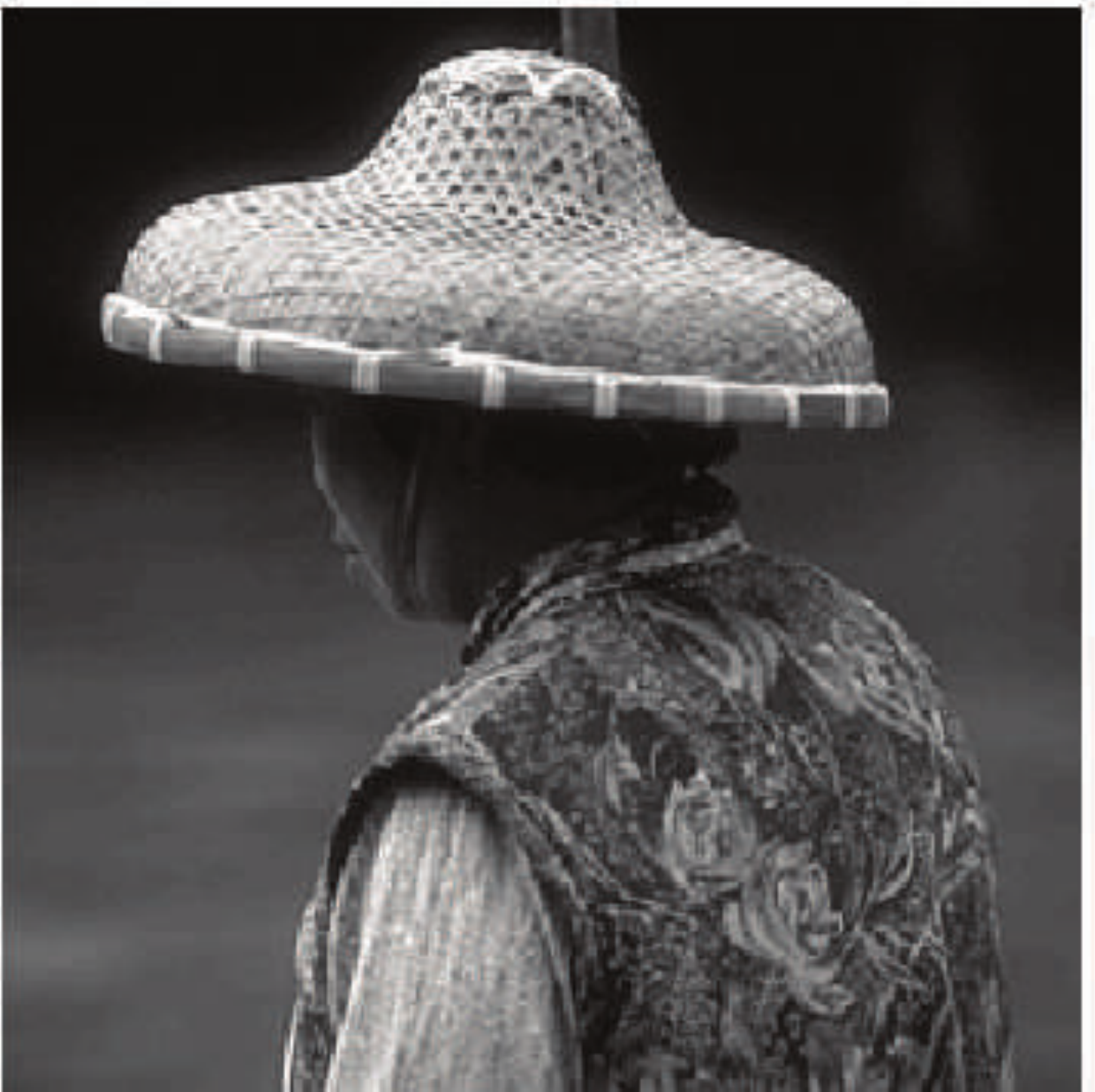} &
\includegraphics[height =3.05cm]{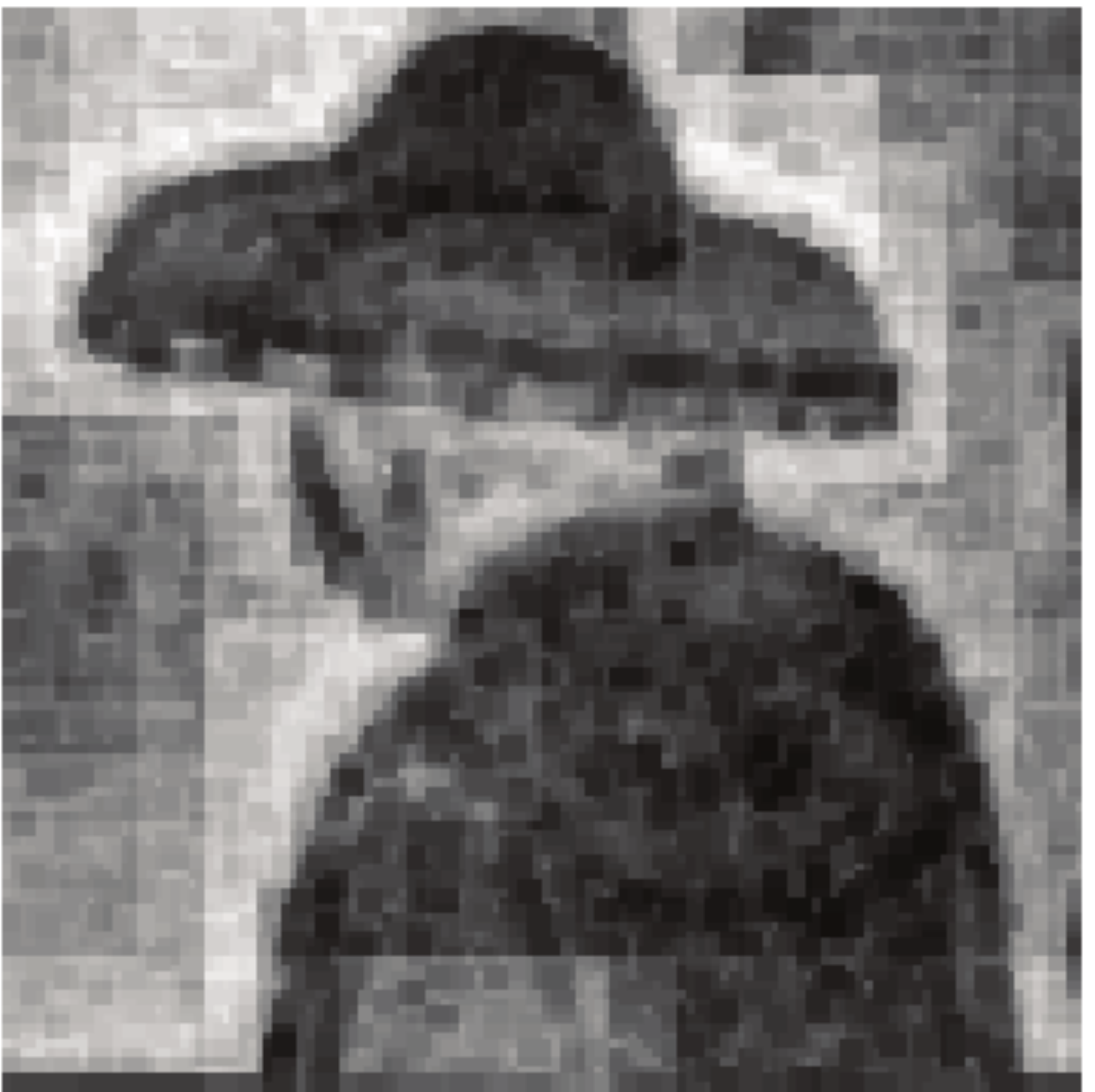}  &
\includegraphics[height =3.05cm]{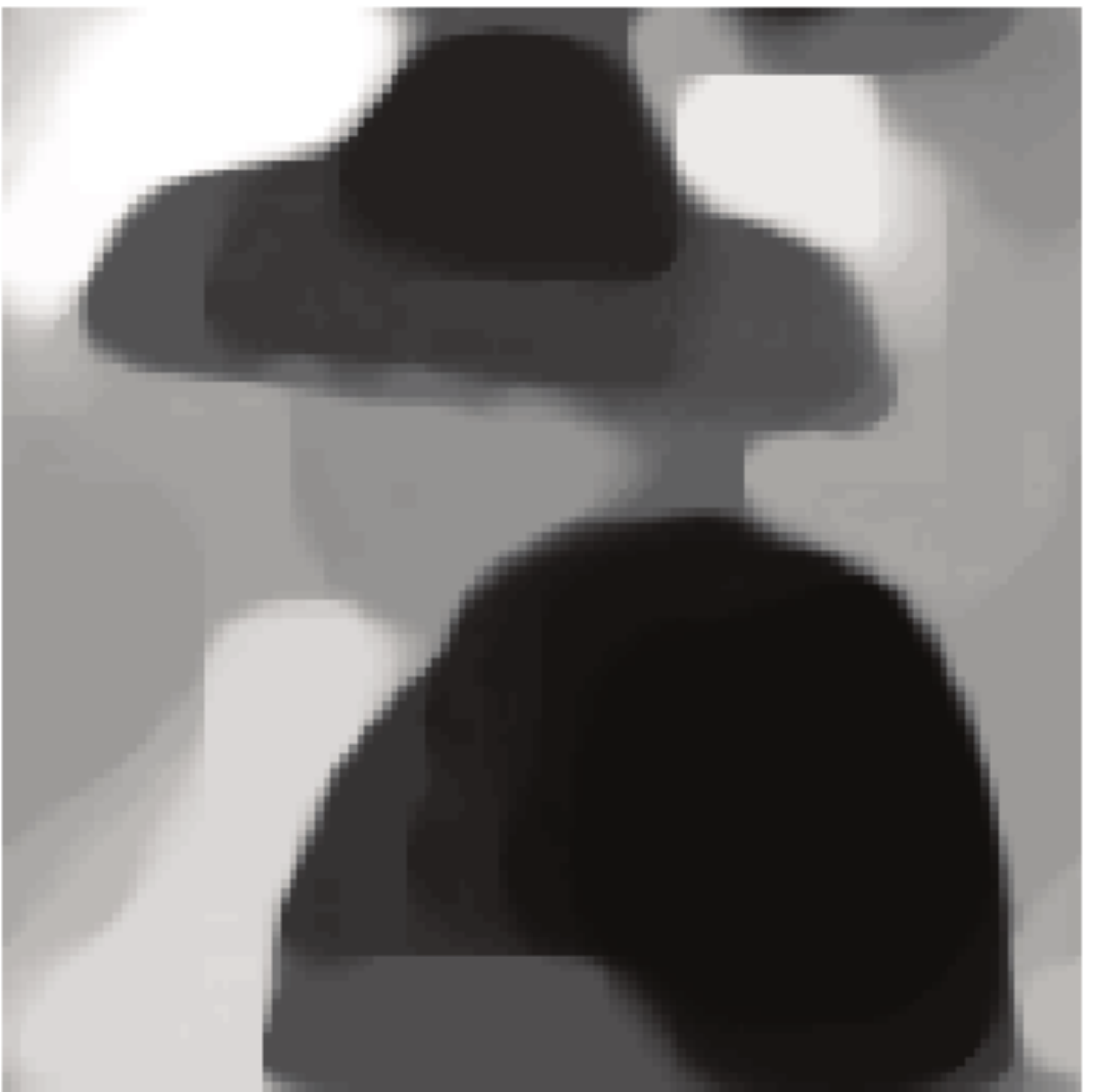}  &
\includegraphics[height =3.05cm]{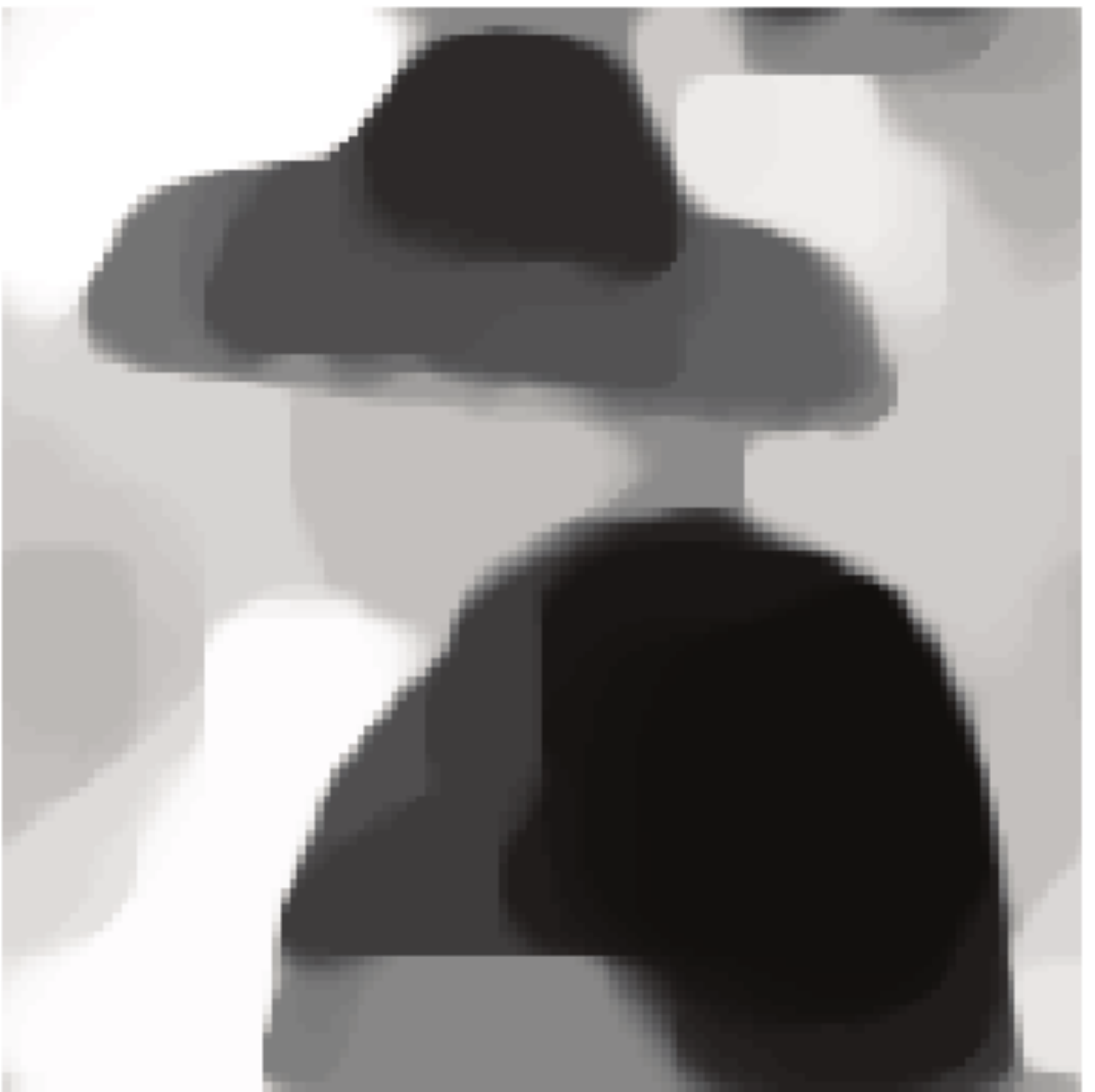} &\\
 Data $\underline{\underline{f}}$ &  $\widehat{\underline{\underline{h}}}_{L}$ &  $\widehat{\underline{\underline{h}}}_L^{\textrm{TV}}$&  $\widehat{\underline{\underline{h}}}_L^{\textrm{TVW}}$& \\
\includegraphics[height =3.05cm]{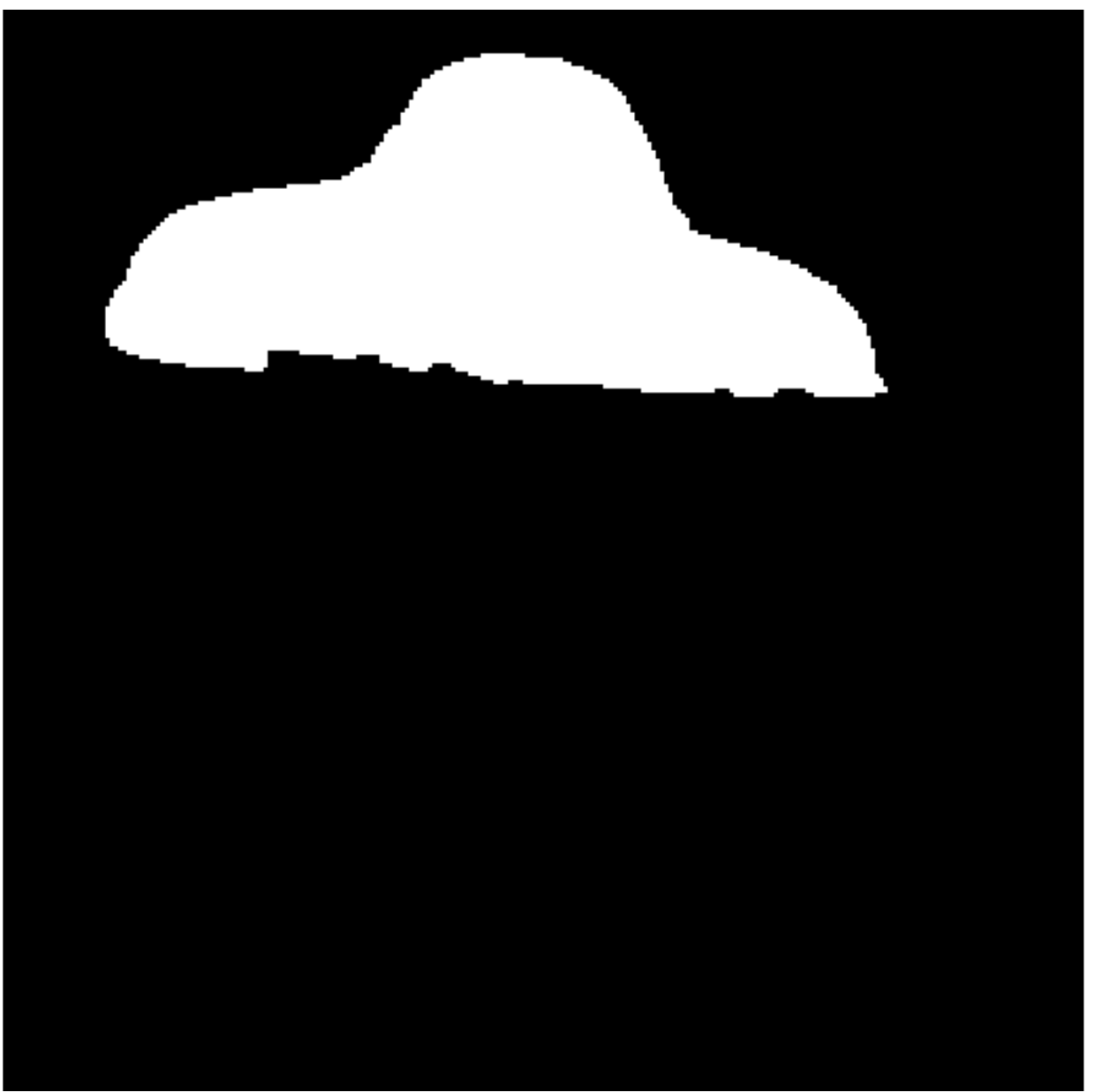} &
\includegraphics[height =3.05cm]{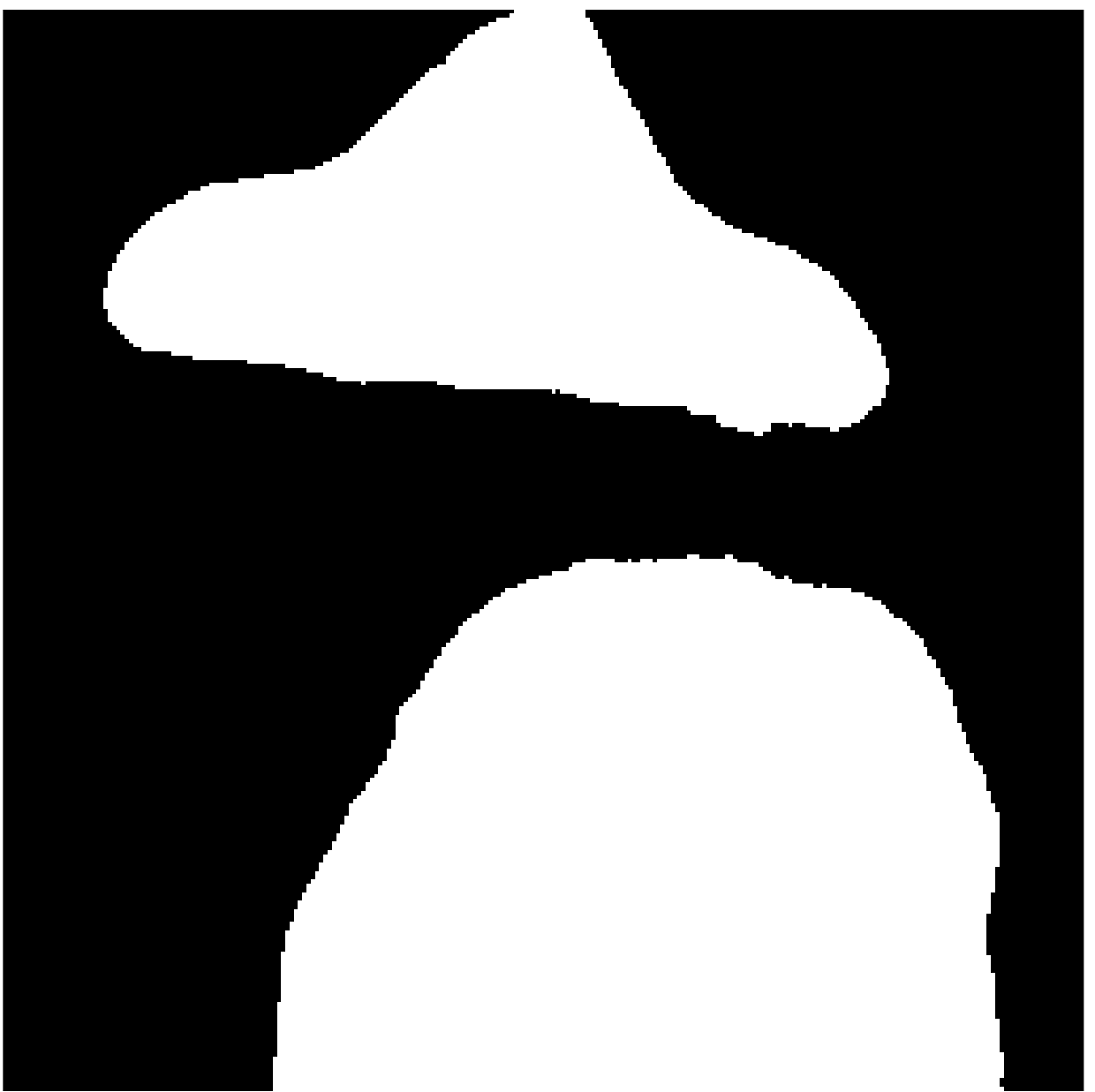} &
\includegraphics[height =3.05cm]{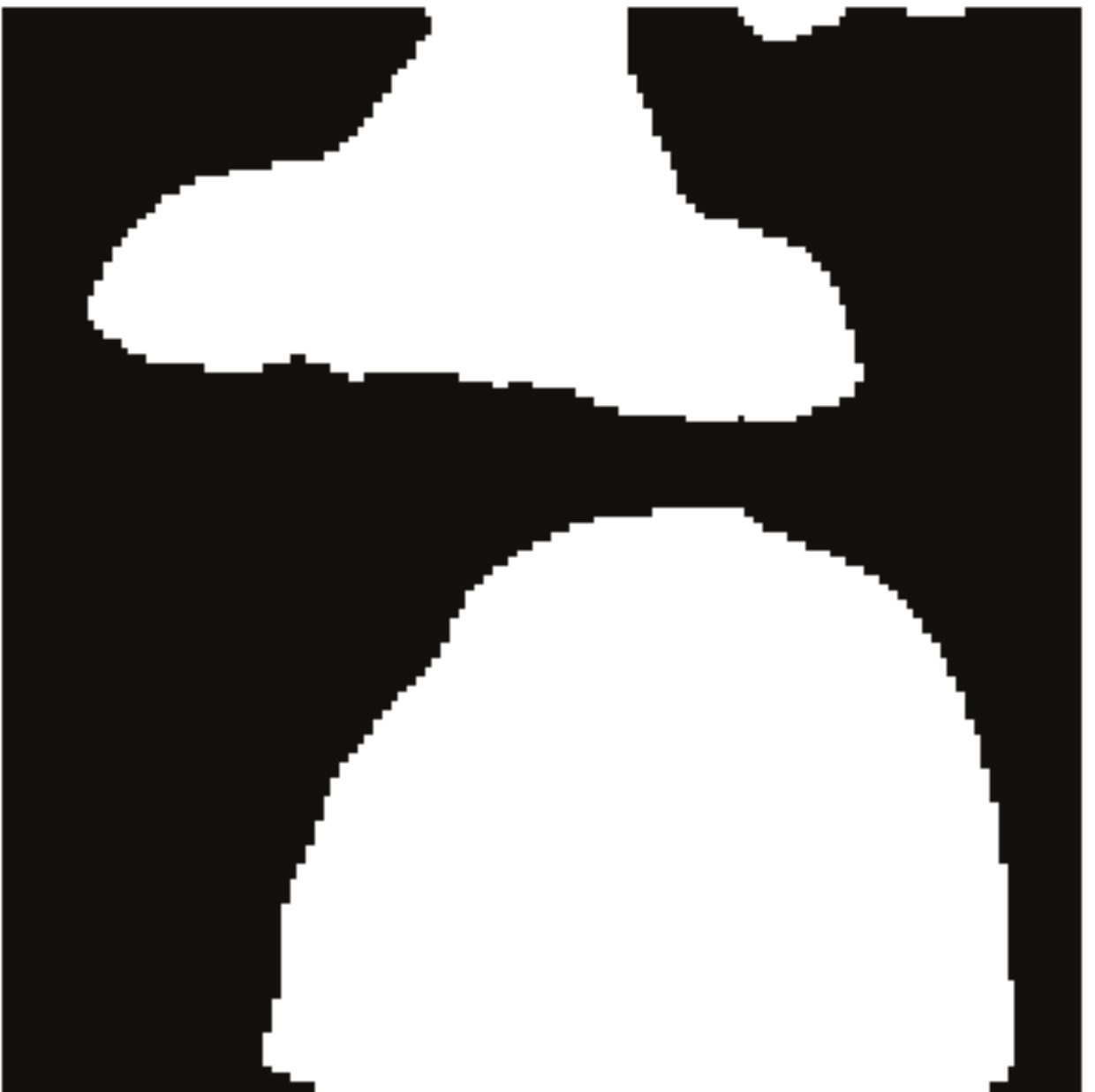} &
\includegraphics[height =3.05cm]{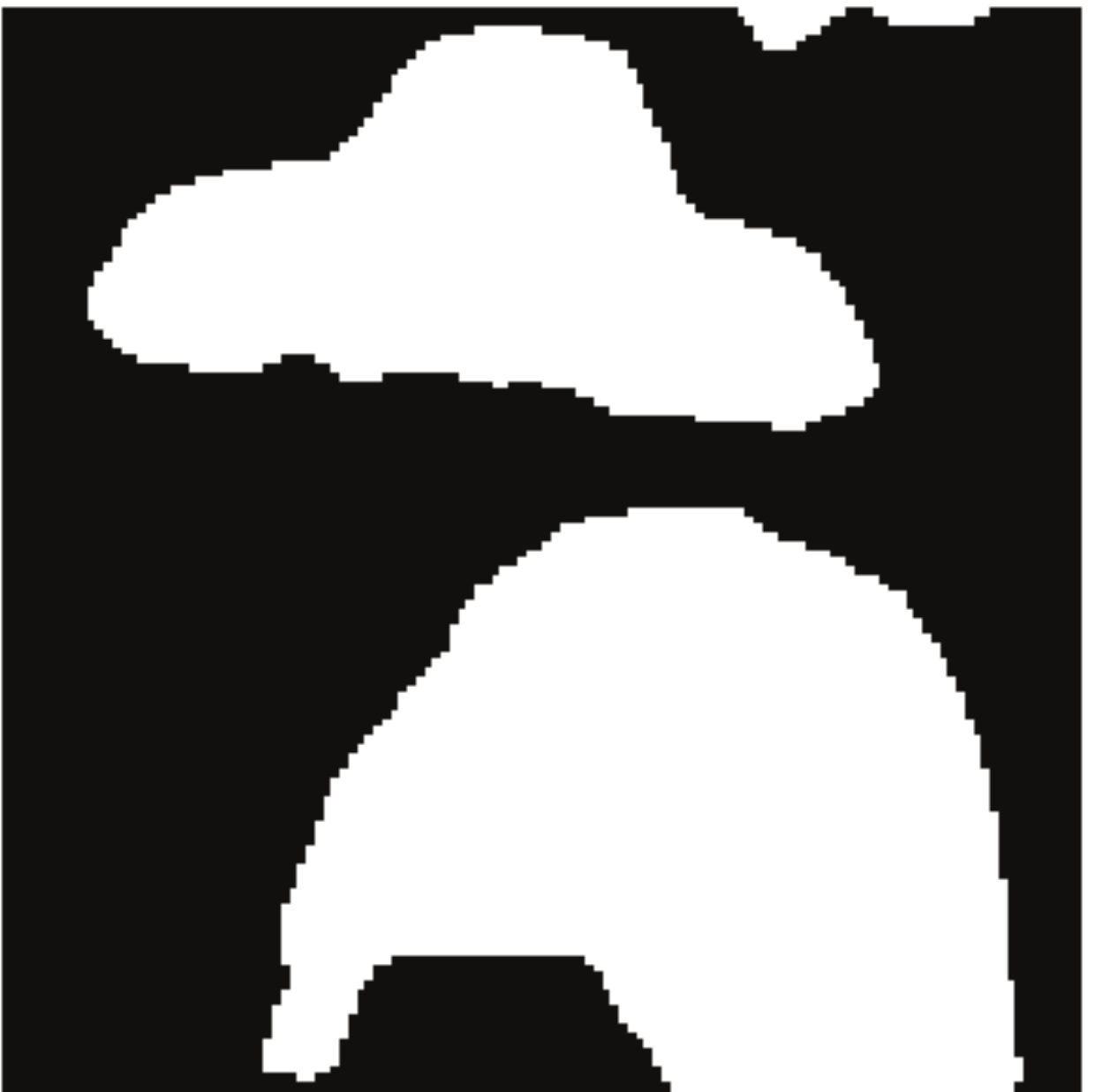}  &
\includegraphics[height =3.05cm]{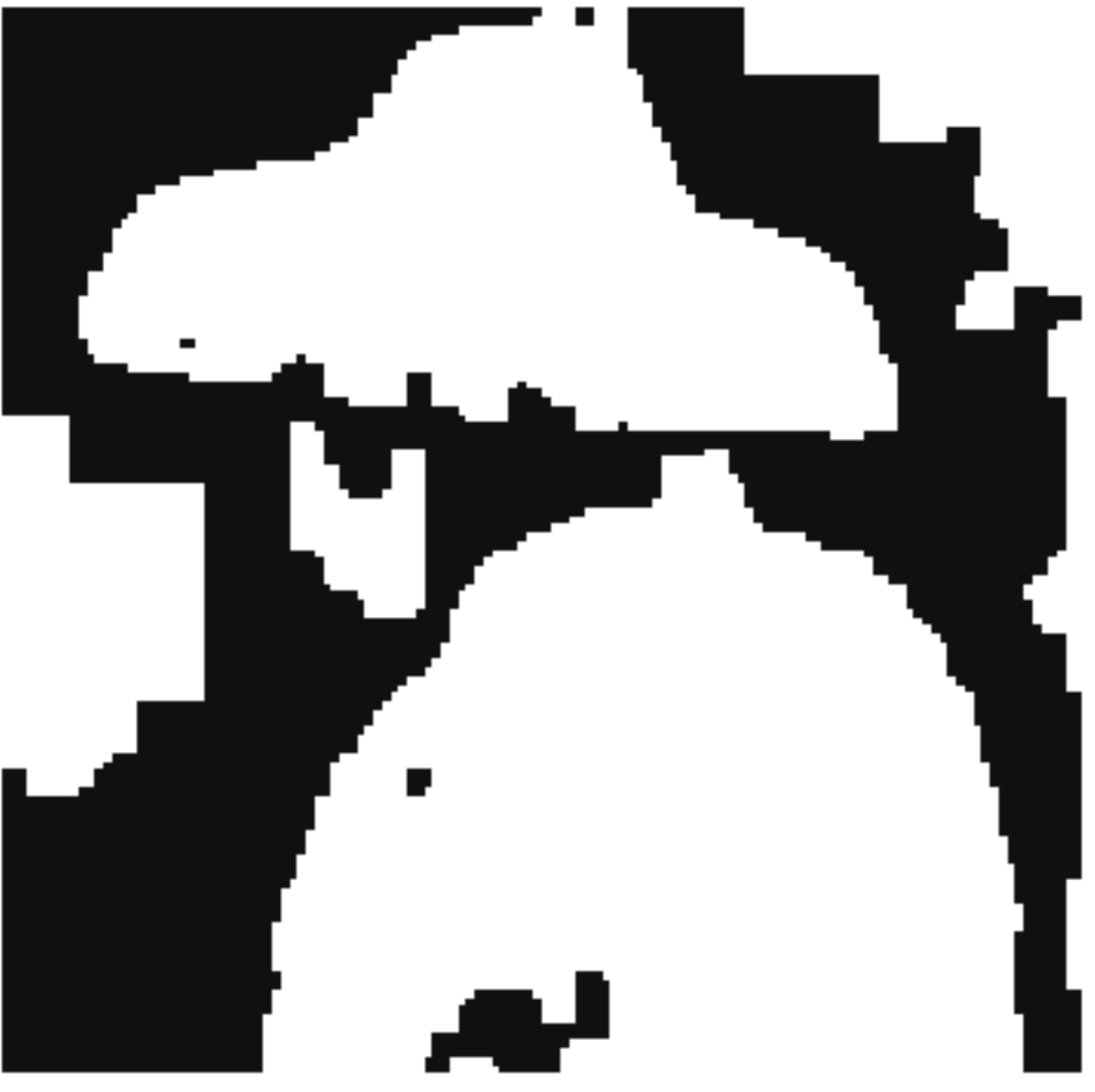}  \\
 \cite{Arbelaez_P_2011_j-ieee-tpami_con_dhis} with $Q=2$&\cite{Yuan_J_2015_j-ieee-tip_fac_bts} with $Q=2$&  $\widehat{\underline{\underline{\Omega}}}^{\textrm{TV}}$ with $Q=2$ & (h)  $\widehat{\underline{\underline{\Omega}}}^{\textrm{TVW}}$ with $Q=2$ &  $\widehat{\underline{\underline{\Omega}}}^{\textrm{RMS}}$ with $Q=2$\\
\includegraphics[height =3.05cm]{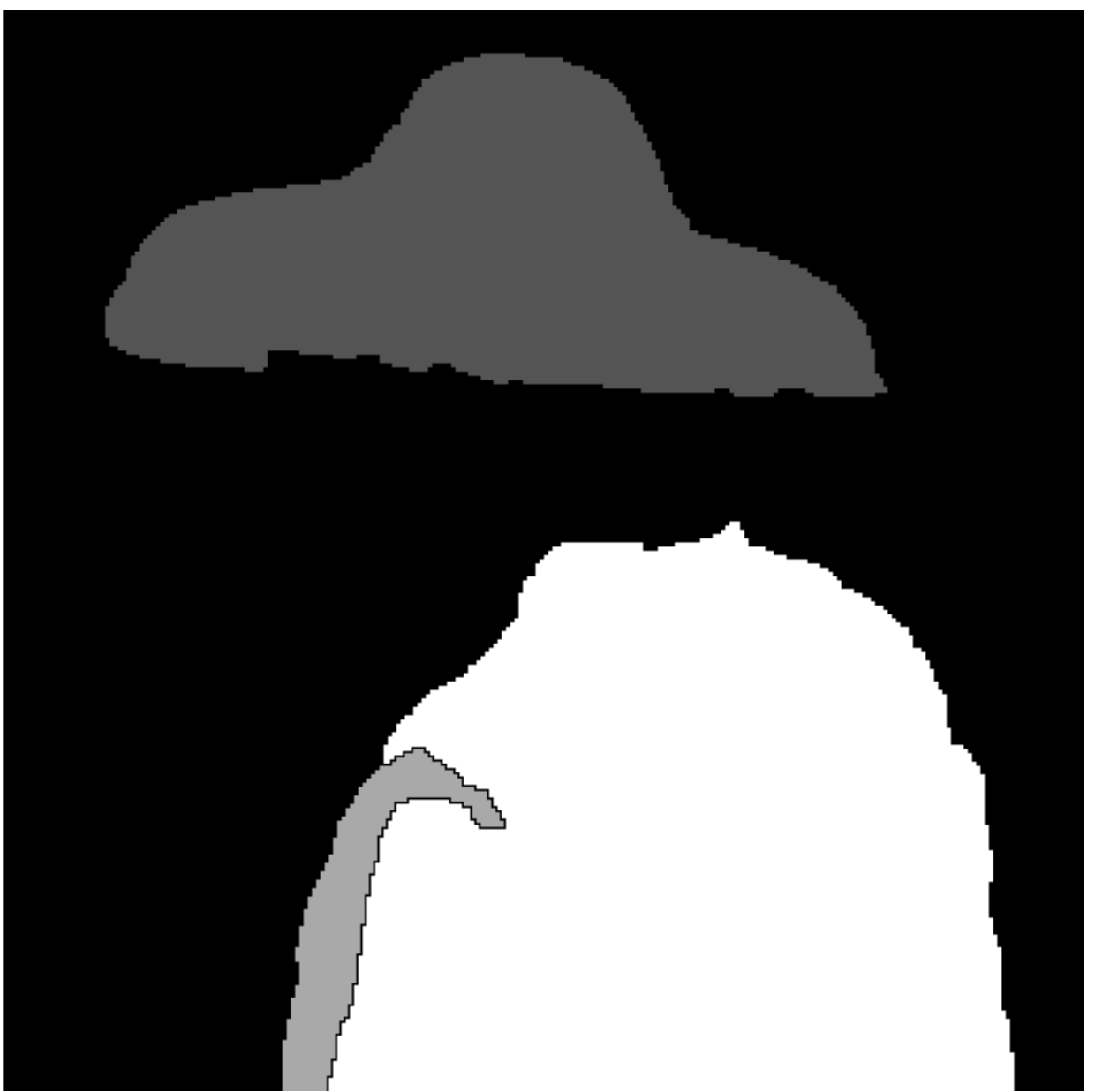} & \includegraphics[height =3.05cm]{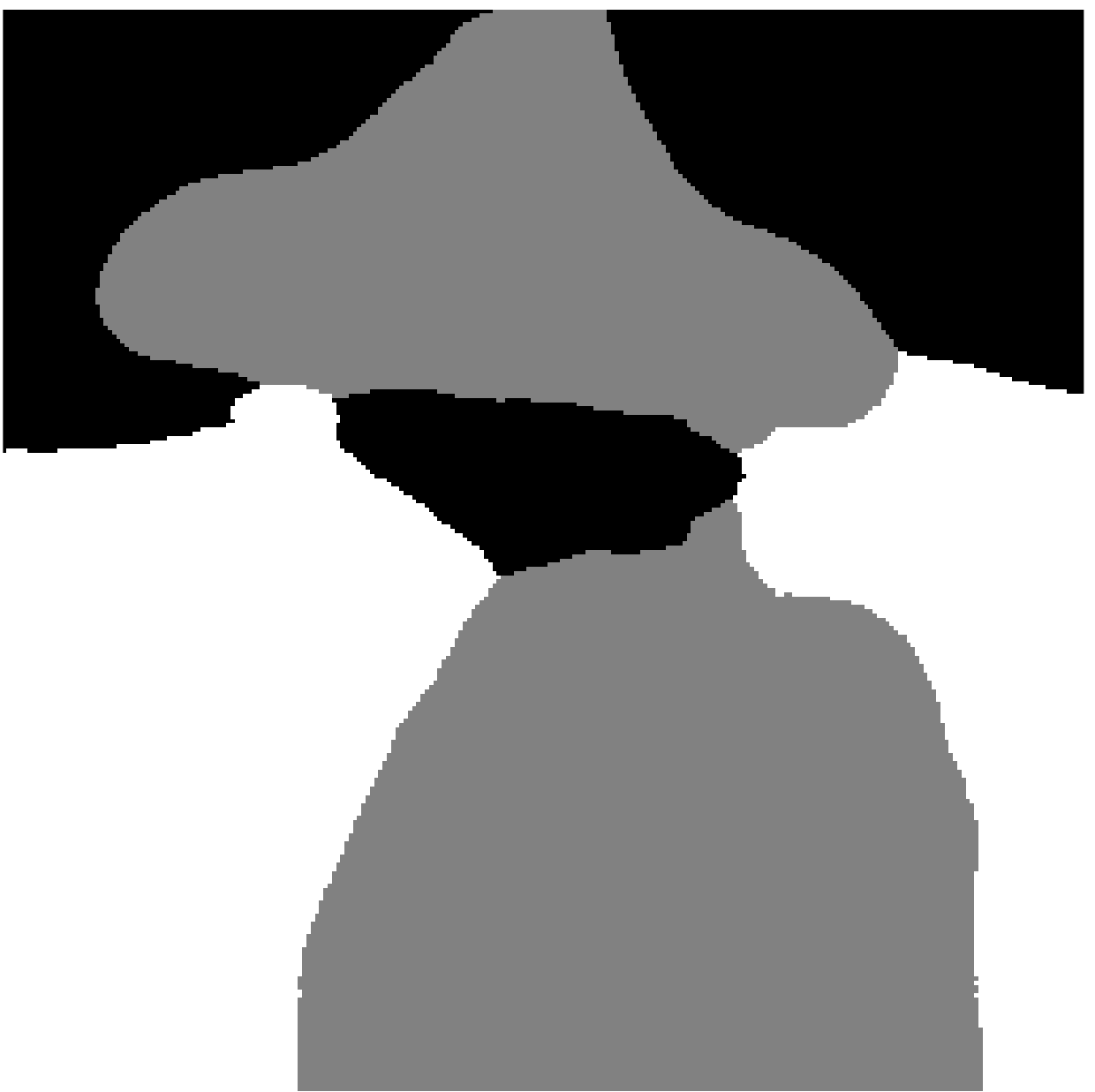}& \includegraphics[height =3.05cm]{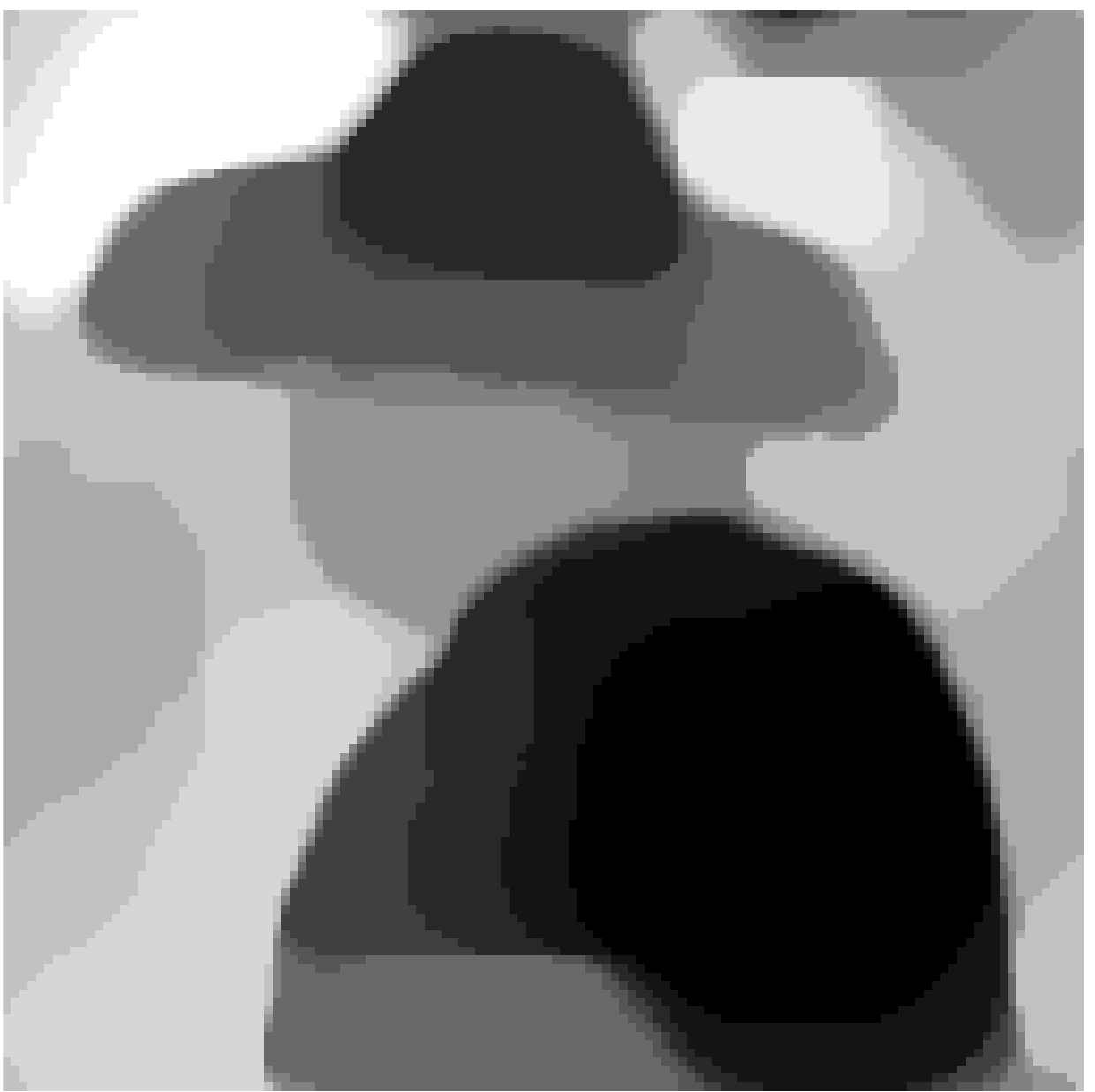}  &
\includegraphics[height =3.05cm]{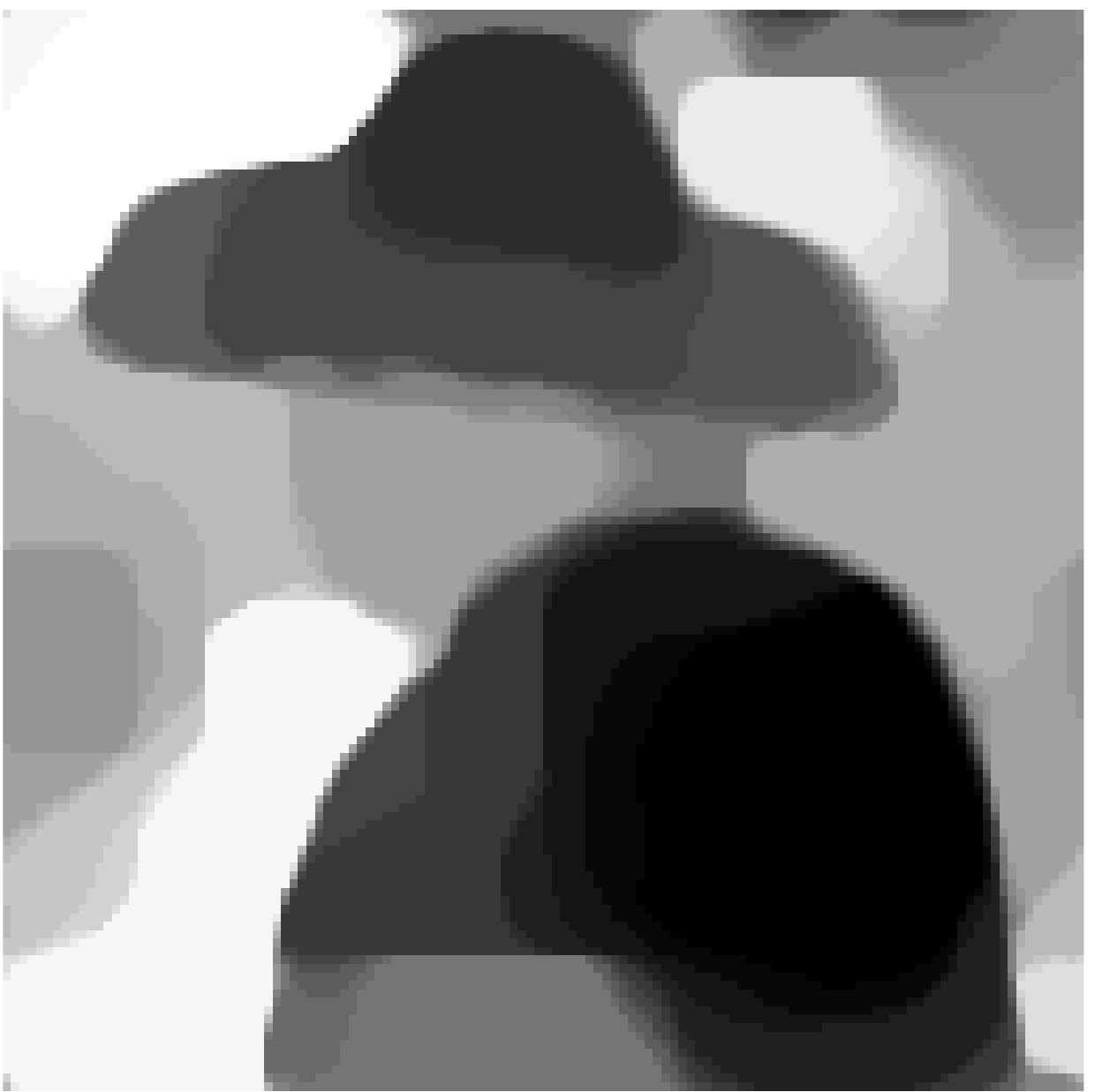} & \includegraphics[height =3.05cm]{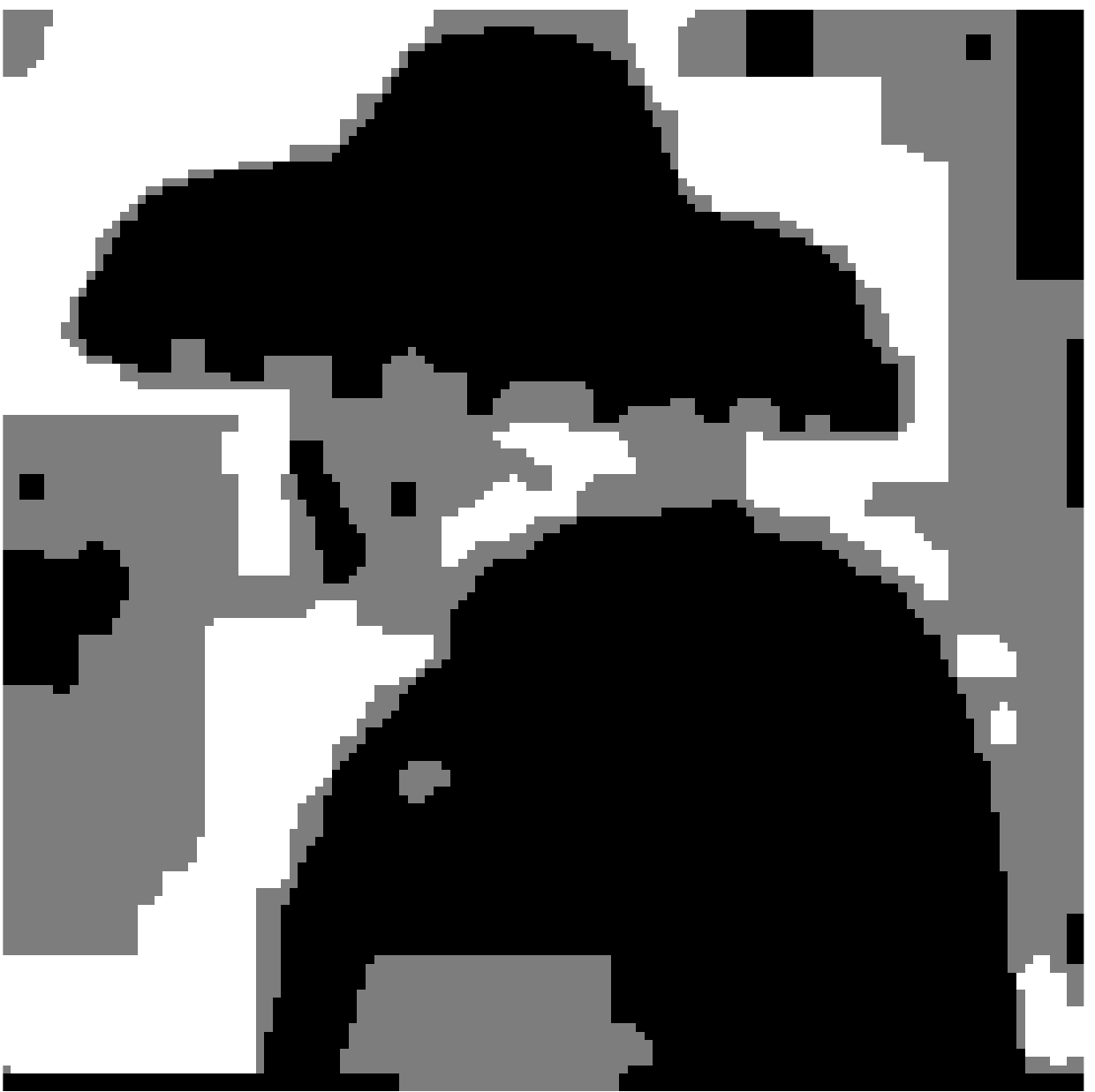}\\
\cite{Arbelaez_P_2011_j-ieee-tpami_con_dhis} with $Q=4$& \cite{Yuan_J_2015_j-ieee-tip_fac_bts} with $Q=3$& $\widehat{\underline{\underline{\Omega}}}^{\textrm{TV}}$ with $Q=12$ &  $\widehat{\underline{\underline{\Omega}}}^{\textrm{TVW}}$  with $Q=22$ &   $\widehat{\underline{\underline{\Omega}}}^{\textrm{RMS}}$ with $Q=3$
\end{tabular}
\caption{Experiments on real data: image extracted from the Berkeley Segmentation Database.  \label{fig:results_real2} }
\end{figure*}

\noindent{\bf Real-world textures.} \quad
\textcolor{black}{To assess the level of generality of the TV-based segmentation procedures,
their performance are quantified and compared on sample textures chosen randomly from a large database considered as reference in the dedicated literature, the Berkeley Segmentation Dataset\footnote{https://www.eecs.berkeley.edu/Research/Projects/CS/vision/bsds}.
No ground truth segmentation is available for that database.
Samples are shown in Fig.~\ref{fig:results_real1} and  Fig.~\ref{fig:results_real2} together with
local regularity TV based estimation and segmentation procedure outcomes.
For comparison,  results achieved with the two state-of-the-art approaches in \cite{Arbelaez_P_2011_j-ieee-tpami_con_dhis} and \cite{Yuan_J_2015_j-ieee-tip_fac_bts} are also displayed.
For both examples, we provide two types of results: a 2-label segmentation result ($Q=2$, Fig.~\ref{fig:results_real1} and  \ref{fig:results_real2}, top rows) and the $Q$-labels segmentation results where, for the state-of-the-art methods, $Q$ had been 
tuned empirically in order to yield the visually most convincing solution, and where the solutions $\widehat{\underline{\underline{\Omega}}}^{\textrm{TV}}$ and $\widehat{\underline{\underline{\Omega}}}^{\textrm{TVW}}$ are deduced from $\widehat{\underline{\underline{h}}}_L^{\textrm{TV}}$ and  $\widehat{\underline{\underline{h}}}_L^{\textrm{TVW}}$ (Fig.~\ref{fig:results_real1} and  \ref{fig:results_real2}, bottom rows).
For both examples, we observe that the outcome of a 2-label segmentation is overall more satisfactory for the proposed TV solution than for the state-of-the-art approaches \cite{Arbelaez_P_2011_j-ieee-tpami_con_dhis} and \cite{Yuan_J_2015_j-ieee-tip_fac_bts}, which tend to fail to extract meaningful shapes.
When the number of labels $Q$ is increased, the behaviour of the methods \cite{Arbelaez_P_2011_j-ieee-tpami_con_dhis} and \cite{Yuan_J_2015_j-ieee-tip_fac_bts} improves, yet the proposed TV methods, and in particular  $\widehat{\underline{\underline{h}}}_L^{\textrm{TV}}$ and $\widehat{\underline{\underline{h}}}_L^{\textrm{TVW}}$, also lead to much more informative segmentation results.
Finally, the results achieved with $\widehat{\underline{\underline{\Omega}}}^{\textrm{TVW}}$ are observed to be visually more satisfactory than those of the two other TV-based approaches, notably for the second example.
This is a very satisfactory outcome regarding the general level of applicability of the proposed TV-based segmentation approaches, as there is no reason a priori to believe that the samples in the Berkeley Segmentation Dataset are perfectly scale-free textures.
Further randomly selected examples from that database are available upon request.}

\setlength{\tabcolsep}{0.2pt}

\setlength{\tabcolsep}{0.2pt}

\section{Conclusions and perspectives}
\label{sec:conc}

We proposed, to the best of our knowledge, the first fully operational {nonparametric} texture segmentation procedures that rely on the concept of local regularity.
The segmentation procedures was designed for the class of piecewise constant local regularity images.
The originality of the proposed approach resided in the combination of wavelet leaders based local regularity estimation and proximal solutions for minimizing the convex criterion underlying the segmentation problem.
Three original and distinct proximal solutions were proposed, all relying on a total variation penalization and proximal based resolution: TV denoising of local regularity estimates followed by thresholding for label determination; TV penalized joint estimation of local regularity and estimation weights followed by thresholding for label determination; direct labeling of local regularity estimates using a TV based partitioning strategy.
The performance of the procedures was validated using stochastic Gaussian model processes with prescribed region-wise constant local regularity and illustrated using realistic model images with real-world textures.
All proposed labeling procedures yielded satisfactory results.
They significantly improved over labeling based directly on (smoothed) local regularity estimates, at the price though of increased computational costs.
Procedure $\widehat{\underline{\underline{h}}}^{\textrm{RMS}}$ further avoided to devise a detailed procedure for histogram thresholding, yet yielding slightly poorer results compared to $\widehat{\underline{\underline{h}}}^{\textrm{TVW}}$.

When constant regularity areas were labeled, regularity can be re-estimated a posteriori by averaging within each area the $ X(a,\kx) $ prior to performing linear regressions \cite{Wendt_H_2007_j-ieee-spm_bootstrap_ema}.

The proposed procedures are currently being used for the analysis of biomedical textures, with encouraging preliminary results.
Comparisons with alternative texture characterization features, such as local entropy rates, are under investigation.

Future work will include investigating in how far the proposed approach can be adapted to handle different regularity models, such as images with piecewise smooth local regularity.
Furthermore, the analysis of textures from real-world applications would benefit from substituting
{piecewise constant H\"older exponents with piecewise constant \emph{multifractal spectra}, providing richer and more realistic models}, at the price, yet, of more severe estimation and segmentation issues.

\end{document}